%% file: root.tex
\documentclass[Afour,sageh,times]{sagej}

\usepackage{moreverb,url}
\usepackage[table]{xcolor}
\usepackage[colorlinks,bookmarksopen,bookmarksnumbered,citecolor=red,urlcolor=red]{hyperref}

\usepackage{pifont}
\newcommand{\cmark}{\ding{51}}%
\newcommand{\lmark}{\ding{121}}%

\newcommand\BibTeX{{\rmfamily B\kern-.05em \textsc{i\kern-.025em b}\kern-.08em
T\kern-.1667em\lower.7ex\hbox{E}\kern-.125emX}}

\def\volumeyear{2016}

\usepackage{todonotes}
\usepackage[printonlyused]{acronym} 

\usepackage{booktabs}       
\usepackage[toc,page]{appendix}
\usepackage{threeparttable, tablefootnote}
\usepackage{tcolorbox}
\usepackage{subcaption}
\usepackage{multirow, makecell}
\usepackage{epstopdf}
\usepackage{tikz}
\usetikzlibrary{trees,positioning,shapes,shadows,arrows}

\usepackage{amssymb}
\relax
\DeclareFontFamily{U}{eur}{\skewchar\font'177}
\DeclareFontShape{U}{eur}{m}{n}{%
	<-6> eurm5 <6-8> eurm7 <8-> eurm10}{}
\DeclareFontShape{U}{eur}{b}{n}{%
	<-6> eurb5 <6-8> eurb7 <8-> eurb10}{}
\DeclareSymbolFont{ugrf@m}{U}{eur}{m}{n}
\SetSymbolFont{ugrf@m}{bold}{U}{eur}{b}{n}
\DeclareMathSymbol{\uptau}{\mathord}{ugrf@m}{"1C}
\DeclareMathSymbol{\Upphi}{\mathord}{ugrf@m}{"08}
\DeclareMathSymbol{\upnu}{\mathord}{ugrf@m}{"17}

\usepackage{graphics}
\definecolor{Light0}{rgb}{0.98, 0.95, 0.99}
\definecolor{Light1}{rgb}{0.98, 0.95, 0.90}
\definecolor{Light2}{rgb}{0.98, 0.98, 0.93}
\definecolor{Light3}{rgb}{0.98, 0.98, 1}

\input{acronyms.tex}
\input{macrosNew.tex}
\graphicspath{{img/}}

\usepackage{totcount}
\newtotcounter{citenum}
\def\oldcite{}
\let\oldcite=\bibcite
\def\bibcite{\stepcounter{citenum}\oldcite}

\begin{document}
\twocolumn

\runninghead{Saveriano, Abu-Dakka, Kramberger, and Peternel}

\title{Dynamic Movement Primitives in Robotics: A Tutorial Survey}

\author{Matteo Saveriano\affilnum{1}, Fares J. Abu-Dakka\affilnum{2}, Alja\v{z} Kramberger\affilnum{3}, and Luka Peternel\affilnum{4}}

\affiliation{\affilnum{1}Department of Computer Science and Digital Science Center, University of Innsbruck, Innsbruck, Austria\\
\affilnum{2}Intelligent Robotics Group, Department of Electrical Engineering and Automation (EEA), Aalto University, Espoo, Finland. \\
\affilnum{3}SDU Robotics, the Maersk McKinney Moller Institute, University of Southern Denmark, Odense, Denmark.\\
\affilnum{4}Delft Haptics Lab, Department of Cognitive Robotics, Delft University of Technology, Delft, The Netherlands}

\corrauth{Fares J. Abu-Dakka, Intelligent Robotics Group, Department of Electrical Engineering and Automation (EEA), Aalto University, Maarintie 8,
02150 Espoo, Finland.}

\email{fares.abu-dakka@aalto.fi}

\begin{abstract}
Biological systems, including human beings, have the innate ability to perform complex tasks in versatile and agile manner. Researchers in sensorimotor control have tried to understand and formally define this innate property. The idea, supported by several experimental findings, that biological systems are able to combine and adapt basic units of motion into complex tasks finally lead to the formulation of the motor primitives theory. In this respect, \acfp*{dmp} represent an elegant mathematical formulation of the motor primitives as stable dynamical systems, and are well suited to generate motor commands for artificial systems like robots. In the last decades, \acp{dmp} have inspired researchers in different robotic fields including imitation and reinforcement learning, optimal control, physical interaction, and human--robot co-working, resulting a considerable amount of published papers. The goal of this tutorial survey is two-fold. On one side, we present the existing \acp{dmp} formulations in rigorous mathematical terms, and discuss advantages and limitations of each approach as well as practical implementation details. In the tutorial vein, we also search for existing implementations of presented approaches and release several others. On the other side, we provide a systematic and comprehensive review of existing literature and categorize state of the art work on \ac{dmp}. The paper concludes with a discussion on the limitations of \acp{dmp} and an outline of possible research directions.
\end{abstract}


\maketitle

\input{Introduction.tex}

\input{DMP_Formulation.tex}

\input{DMP_Extentions.tex}

\input{DMP_System_Integration.tex}

\input{DMP_Applications.tex}

\input{DMP_Discussion}

\section{Concluding remarks}\label{sec:conclusion}
Since their introduction in early 2000's, \acp{dmp} have established as one of the most used and popular approaches for motor commands generator system in robotics. Several authors have exploited and extended the classical formulation to overcome some limitations and fulfill different requirements. Their research resulted in a large amount of papers published over the last two decades.

One of the aims of this paper is to categorize and review the vast literature on \acp{dmp}. We took a systematic review approach and automatically searched for \ac{dmp} related papers in a popular database. A manual inspection of the resulting papers, guided by clear and unbiased criteria, led to the papers included on this tutorial survey.

Another aim of our work is to provide a tutorial on \acp{dmp} that presents the classical formulation and the key extensions in rigorous mathematical terms. We made an effort to unify the notation among different approaches in order to make them easier to understand. Moreover, we provide useful guidelines that guide the reader to select the right approach for a given application. In the tutorial vein, we have also searched for open-source implementation of the described approaches and released to the community several implementations of \ac{dmp}-based approaches.

Advantages of \acp{dmp} have been discussed as well as their limitations and the open-issues. We have summarized them in \tabref{tab:conclusion} where we also indicate the solved issues and the one that require further investigation. In this respect, as research on \ac{dmp} is still very active, we provide a comprehensive discussion that will help the reader to understand what has been done in the field and where he can put his research focus.

\begin{funding}
This work has been partially supported by:
\begin{itemize}
    \item[-] CHIST-ERA project IPALM (Academy of Finland decision 326304).
    \item[-] The Austrian Research Foundation (Euregio IPN 86-N30, OLIVER).
    \item[-] Innovation Fund Denmark (Research and innovation project MADE FAST).
\end{itemize}
\end{funding}

\bibliographystyle{SageH}
\bibliography{ref.bib}

\end{document}

%% file: acronyms.tex
\newacro{vic}[VIC]{Variable Impedance Control}

\newacro{pi2}[PI\textsuperscript{2}]{Policy Improvement with Path Integrals}
\newacro{dmp}[DMP]{Dynamic Movement Primitive}
\newacro{cmp}[CMP]{Compliant Movement Primitive}
\newacro{seds}[SEDS]{Stable Estimator of Dynamical Systems}
\newacro{ilc}[ILC]{Iterative  Learning  Control}
\newacro{gp}[GP]{Gaussian Process}
\newacro{gpr}[GPR]{Gaussian Process Regression}
\newacro{gmm}[GMM]{Gaussian Mixture Model}
\newacro{gmr}[GMR]{Gaussian Mixture Regression}
\newacro{ppc}[PPC]{Passivity-Preservation Control}
\newacro{tpgmm}[TP-GMM]{Task-Parameterized \ac{gmm}}
\newacro{lfd}[LfD]{Learning from Demonstration}
\newacro{il}[IL]{Imitation Learning}
\newacro{wls}[WLS]{weighted least-squares}
\newacro{spd}[SPD]{Symmetric Positive Definite}
\newacro{emg}[EMG]{Electromyography}
\newacro{vil}[VIL]{Variable Impedance Learning}
\newacro{vilc}[VILC]{Variable Impedance Learning Control}
\newacro{ai}[AI]{Artificial Intelligent}
\newacro{sea}[SEA]{Series Elastic Actuation}
\newacro{dof}[DoF]{Degree of Freedom}
\newacro{vmp}[VMP]{Via-points Movement Primitive}
\newacro{lwr}[LWR]{Locally Weighted Regression}
\newacro{rl}[RL]{Reinforcement Learning}
\newacro{auv}[AUV]{Autonomous Underwater Vehicle}
\newacro{uav}[UAV]{Unmanned Areal Vehicle}
\newacro{cmaes}[CMA-ES]{Covariance Matrix Adaptation-Evolution Strategies}
\newacro{ccdmp}[CC-DMP]{Coordinate Change-\acp{dmp}}
\newacro{gpdmp}[GPDMP]{ Global Parametric Dynamic Movement Primitive}
\newacro{bbo}[BBO]{Black-Box Optimization}
\newacro{dmpbbo}[DMPBBO]{\ac{dmp} \ac{bbo}}
\newacro{ros}[ROS]{Robotic Operating System}
\newacro{cnn}[CNN]{Convolutional Neural Network}
\newacro{nn}[NN]{Neural Network}
\newacro{aedmp}[AEDMP]{AutoEncoded \ac{dmp}}
\newacro{power}[PoWER]{Policy Learning by Weighting Exploration with the Returns}
\newacro{pd}[PD]{Proportional Derivative}
\newacro{momp}[MoMP]{Mixture of Motor Primitives}
\newacro{promp}[ProMP]{Probabilistic Movement Primitives}
\newacro{hrl}[HRL]{Hierarchical \ac{rl}}
\newacro{kmp}[KMP]{Kernelized Movement Primitive}
\newacro{rbf}[RBF]{Radial Basis Function}
\newacro{rbfnn}[RBF-NN]{\ac{rbf}-\ac{nn}}
\newacro{}[]{}
\newacro{}[]{}
\newacro{}[]{}

%% file: macrosNew.tex
\newcommand{\bm}[1]{\boldsymbol{\mathbf{#1}}}

\newcommand{\Kpt}{{\bm{K}_t^\mathcal{P}}}
\newcommand{\Dvt}{{\bm{D}_t^\mathcal{V}}}
\newcommand{\Kot}{{\bm{K}_t^\mathcal{O}}}
\newcommand{\Dwt}{{\bm{D}_t^\mathcal{W}}}

\newcommand{\Kp}{{\bm{K}^\mathcal{P}}}
\newcommand{\Dv}{{\bm{D}^\mathcal{V}}}
\newcommand{\Ko}{{\bm{K}^\mathcal{O}}}
\newcommand{\Dw}{{\bm{D}^\mathcal{W}}}

\newcommand{\fet}{{\bm{f}_t^e}}
\newcommand{\fe}{{\bm{f}^e}}
\newcommand{\tauet}{{\bm{\uptau}_t^e}}
\newcommand{\taue}{{\bm{\uptau}^e}}
\newcommand{\spd}{\bm{\mathcal{S}}_{++}^{m}}
\newcommand{\sym}{\bm{Sym}^{m}}

\newcommand{\dz}{\Dot{z}}
\newcommand{\dy}{\Dot{y}}
\newcommand{\dx}{\Dot{x}}
\newcommand{\q}{\bm{q}}
\newcommand{\w}{\bm{w}}
\newcommand{\dq}{\bm{\Dot{q}}}
\newcommand{\LogQ}{\text{Log}^{q}}
\newcommand{\LogR}{\text{Log}^{R}}
\newcommand{\LogS}{\text{Log}^{+}}
\newcommand{\ExpQ}{\text{Exp}^{q}}
\newcommand{\ExpR}{\text{Exp}^{R}}
\newcommand{\ExpS}{\text{Exp}^{+}}
\newcommand{\X}{\bm{X}}
\newcommand{\R}{\bm{R}}
\newcommand{\dR}{\bm{\Dot{R}}}

\newcommand{\trsp}{{^{\top}}}

\newcommand{\figref}[1]{Fig.~\hyperref[#1]{\ref*{#1}}}
\newcommand{\figsref}[1]{Figures~\hyperref[#1]{\ref*{#1}}}
\newcommand{\Figref}[1]{Figure~\hyperref[#1]{\ref*{#1}}}

\newcommand{\tabref}[1]{Table~\hyperref[#1]{\ref*{#1}}}
\newcommand{\secref}[1]{Section~\hyperref[#1]{\ref*{#1}}}
\newcommand{\algoref}[1]{Algorithm~\hyperref[#1]{\ref*{#1}}}

\newcommand{\wrt} {\textit{w.r.t.}~} %
\newcommand{\eg} {\textit{e.g.,}~} %
\newcommand{\ie} {\textit{i.e.,}~} %
\newcommand{\etc}{\textit{etc}} %

\newlength{\Oldarrayrulewidth}

\definecolor{darkgreen}{rgb}{0.0,0.49,0.19}

\newcommand{\python}{\texttt{Python}}
\newcommand{\cpp}{\texttt{C++}}

\newcommand{\matlab}{\texttt{Matlab}}

%% file: Introduction.tex
\section{Introduction}
\label{sec:introduction}
\begin{center}
\emph{How biological systems, like humans and animals, execute complex movements in a versatile and creative manner?} 
\end{center}
In the past decades, researchers of neurobiology and motor control have made a significant effort trying in to answer this research question and their experimental findings lead to the formulation of the \textit{motor} or \textit{motion primitives theory}. The motion primitives theory explains the execution of complex motion with the ability of biological systems of sequencing and adapting units of actions, the so-called motion primitives~\citep{mussa1999modular, flash2005motor}. 

\acfp{dmp} have their roots in the motor control of biological systems and can be seen as a rigorous mathematical formulation of the motion primitives as stable nonlinear dynamical systems~\citep{Schaal2006,schaal2006dynamic}. In this respect, \acp{dmp} represent one of the first attempts to answer the research question:
\begin{center}
\emph{How artificial systems, like (humanoid) robots, can execute complex movements in a versatile and creative manner?}
\end{center}
Beyond their biological motivation, \acp{dmp} have a simple and elegant formulation, guarantee convergence to a given target, are sufficiently flexible to create complex behaviors, are capable of reacting to external perturbations in real-time, and can be learned from data using efficient algorithms. These properties explain the ``success'' of \acp{dmp} in robotic applications, where they have established as a prominent tool for learning and generation of motor commands. Since their formulation in the pioneering work from \citeauthor{Ijspeert2001a} \citep{Ijspeert2001a, Ijspeert2002Movement}, \acp{dmp} have been successfully exploited in a variety of applications, becoming de facto the first approach that novices in the \ac{il} field use on their robots.

\subsection{Existing surveys and tutorials}
\begin{table*}[t]
	\small\sf\centering
	\caption{Comparison between existing reviews and tutorial about \acp{dmp} and our tutorial survey.}
    \input{existing_survey_table.tex}
	\label{tab:sota}
\end{table*} 
The popularity of \acp{dmp} resulted in a large amount of work that use, modify, or extend the original formulation of Ijspeert and colleagues. In this paper, we name \textit{classical \acp{dmp}} the \ac{dmp} formulation initially presented in \citep{Ijspeert2001a} and further refined in \citep{Ijspeert2002Movement, Ijspeert2002Learning}.
As shown in \tabref{tab:sota}, some tutorials and surveys already tried to categorize and review existing work on \acp{dmp}.

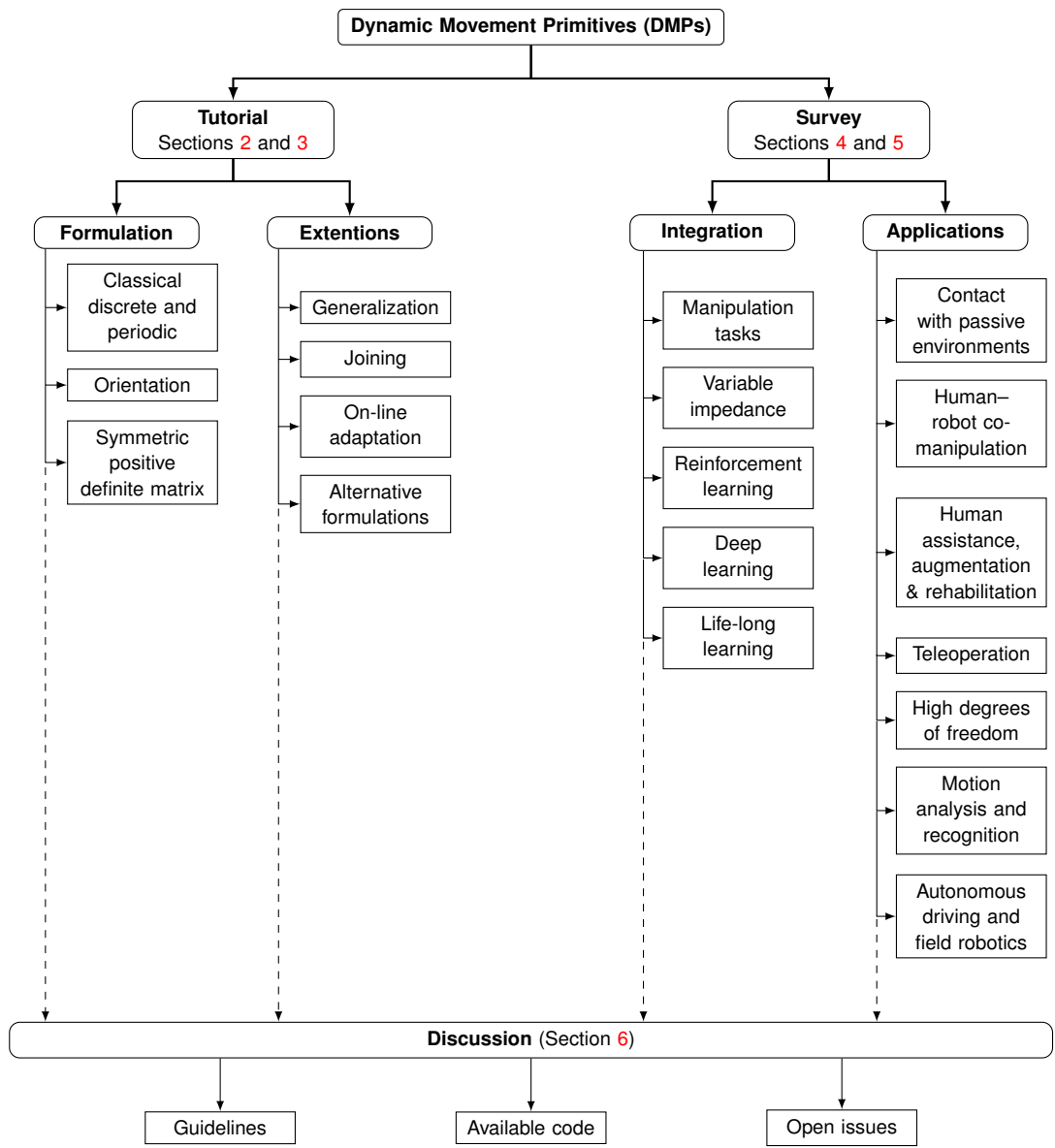
\begin{figure*}[!t]
	\centering
	\input{taxonomy}
	\caption{The structure of this tutorial survey on \acp{dmp}.}
	\label{fig:taxonomy}
\end{figure*}

\cite{schaal2007dynamics} presented the classical \acp{dmp} as an attempt to unify nonlinear dynamical systems and optimal control theory, \ie the two prominent frameworks used to derive computational models of neuro-biological motor theories~\citep{mussa1999modular, flash2005motor}. 
In their tutorial paper, \cite{Ijspeert2013Dynamical} presented a homogeneous formulation of rhythmic and discrete \acp{dmp} together with some extensions including coupling terms, generalization to different goal, and online adaptation for collision avoidance. They also described possible applications in \ac{il} and motion recognition methods. In the same year, \cite{pastor2013dynamic} published their tutorial on classical \acp{dmp} with a special focus on online adaptation of the \ac{dmp} attractor landscape by integrating the perceptual information into the action generation process. Later on, \cite{denivsa2016review} reviewed the so-called \acfi{cmp}, which was first introduced by \cite{petric2014onlinelearning}. \acp{cmp} combine classical \ac{dmp} to generate the desired kinematic path and \textit{torque primitives}---a weighted summation of Gaussian basis functions---to generate task-specific dynamics. As shown in the review, \acp{cmp} are capable of accurately tracking the kinematic path in a compliant manner, which makes them well suited for tasks that require interaction of the robot with the environment.

However the above-mentioned reviews and tutorials primarily focused on the methods and advancements within their respective research group and/or focused on a specific problem or field of application. On the other hand, the \acp{dmp} related literature is extensive and broad, with contributions from many research groups that made advancements in several important fields of application. Therefore, the proposed survey and tutorial on \acp{dmp} aims to scan a wider range and present a tutorial with unified and structured formulations for various \acp{dmp} methods and advancements up to date. This should make it clearer for the users to see the differences and connections between various methods, and can contribute to easier application. In addition, we provide a more comprehensive and categorised survey of all major \acp{dmp} application areas in robotics. This can help to inspire the readers to apply the \acp{dmp} in various areas.

In the tutorial part, we present mathematical formulations, implementation details, and potential issues of existing \ac{dmp} formulations starting from the classical \acp{dmp} presented in \citep{Ijspeert2002Movement, Ijspeert2002Learning} up to recent extensions of \acp{dmp} to Riemannian geometry and \ac{spd} matrices \citep{abudakka2020Geometry}. In the survey part, we meticulously review existing literature on \acp{dmp} in a comprehensive and methodological manner by focusing on the quality and significance of their continuations without putting a bias on any particular research group. Details on the systematic review procedure are given as follows. 

\subsection{Systematic review process}
We preformed am automatic search for documents containing the string 
$$\texttt{Dynamic Movement Primitive}$$ in Scopus on $25$ November $2020$ returned $1223$ papers. We found that Scopus lists papers only from $2004$ on. Therefore, we manually track related work from $2001$ (preliminary work on \acp{dmp}) to $2003$. We further refined the search on $01$ February $2021$ to include last minute papers.

We manually inspect all the papers and removed the ones that do not explicitly use \acp{dmp} and that only compare against \ac{dmp} in their literature review. The first and foremost selection criteria were the technical quality of work and the significance of the contribution with respect the \ac{dmp} state-of-the-art prior to the publication of any particular paper. In other words, we asked the question 'did the paper make a significant step change in the field?'. Therefore, we discarded papers that presented similar (or same) ideas multiple times, or that made insignificant improvements to the state-of-the-art. If multiple papers presented the same/similar idea, we included the one with the most comprehensive technical quality, and if the quality was similar, the next deciding factors were publication in more prestigious journals/venues or the most cited ones. This manual selection led to the 276  papers on \acp{dmp} (out of a total of \total{citenum} references) analyzed in this work.

\subsection{A taxonomy of \texorpdfstring{\ac{dmp}}{} related research}
The systematic review of \ac{dmp} literature lead to the taxonomy shown in \figref{fig:taxonomy}, which also describes the structure of this paper. \acp{dmp} are placed at the root of the tree and branch into two nodes, namely the \textit{tutorial} and the \textit{survey}. In the tutorial part we present different \ac{dmp} formulations and extensions in rigorous mathematical terms.

The tutorial part spans Sections~\ref{sec:formulation}~and~\ref{sec:advanceMethods}. \secref{sec:formulation} embraces \acp{dmp} formulations for \textit{discrete and periodic} motions, \textit{orientation} trajectories, and \textit{\acp{spd}} matrices. \secref{sec:advanceMethods} discusses extensions of the \ac{dmp} formalism to account for skills \textit{generalization}, \textit{joining} of multiple primitives, \textit{online adaptation} based on force feedback or reference velocity. The section ends with a short description of \ac{dmp} related formulations.

The survey part spans Sections~\ref{sec:integration}~and~\ref{sec:applications}. \secref{sec:integration} presents \acp{dmp} integration in larger executive frameworks for \textit{manipulation} and \textit{variable impedance} tasks,  \textit{reinforcement}, \textit{deep}, and \textit{life-long learning}.  \secref{sec:applications} presents \acp{dmp} in different robotic applications including \textit{physical interaction}, \textit{co-manipulation}, \textit{rehabilitation}, \textit{teleoperation}, \textit{motion recognition}, \textit{humanoids} and \textit{field robotics}, and \textit{autonomous driving}.

The paper ends with a discussion (\secref{sec:discussion}) of presented approaches with the aim of providing, where possible, guidelines to select the most suitable \ac{dmp} approach for specific needs. We have also collected available \ac{dmp} implementations (see \tabref{tab:source_code}) and contributed to the community with further open source implementations available at \url{https://gitlab.com/dmp-codes-collection}. \secref{sec:discussion} terminates with a discussion on open issues and possible research directions.

\subsection{Contribution overview}
Our paper has several key contributions that are summarized as follows. 

Concerning the tutorial part:
\begin{itemize}
    \item We present the classical \ac{dmp} formulation and existing variations of this formulation in a unified manner with rigorous mathematical terms, providing implementation details and discussing advantages and limitations of different approaches (\secref{sec:formulation}).
    \item We describe \textit{advanced} approaches where \acp{dmp} are integrated into sophisticated control and/or larger executive frameworks (\secref{sec:advanceMethods}). 
    \item We release to the community several implementations of described approaches. Detailed information on these code repositories are provided in \tabref{tab:source_code} and \secref{sec:discussion}. Moreover, we search for existing open-source implementations of the presented formulations and list them in our repository (\secref{sec:code}).
\end{itemize}

Concerning the survey part:
\begin{itemize}
    \item We perform a systematic literature search to provide a comprehensive and unbiased review of the topic (Sections~\ref{sec:integration}~and~~\ref{sec:applications}). 
    \item We categorize existing work on \acp{dmp} into different streams and highlight prominent approaches in each category (\figref{fig:taxonomy} and Sections~\ref{sec:integration}~and~~\ref{sec:applications}). 
    \item We present guidelines to select the the most suitable approach for different applications, discuss limitations inherent to the \ac{dmp} formalism, and highlight open issues and possible research directions (\secref{sec:discussion}).
\end{itemize}

%% file: existing_survey_table.tex
\begin{tabular}{m{0.17\textwidth}m{0.2\textwidth}m{0.55\textwidth}}
	\toprule
	\textbf{Survey/Tutorial} & \textbf{Topics} & \textbf{Description} \\
	\midrule
	\citep{schaal2007dynamics} & $\bullet$ Classical \acsp{dmp} \newline
            					$\bullet$ Online adaptation \newline
            					$\bullet$ Optimization & A tutorial that provides a unifying view on the two main approaches used to develop computational motor control theories, namely differential equations and optimal control. In this work, discrete and rhythmic \acsp{dmp} \citep{Ijspeert2002Movement, Ijspeert2002Learning} are presented as a computational model of the motor primitives theory~\citep{mussa1999modular} that unifies nonlinear differential equations and optimal control. The tutorial has a section dedicated to \ac{dmp} parameters optimization beyond \acsp{il}. \citeauthor{schaal2007dynamics} show how to optimize \ac{dmp} parameters to minimize various costs describing, for instance, the total jerk of the trajectory or the end-point variance. \\
	\midrule	
	\citep{Ijspeert2013Dynamical} & $\bullet$ Classical \acsp{dmp}  \newline
            											$\bullet$ Generalization  \newline
            											$\bullet$ Online adaptation  \newline
            											$\bullet$ Coupling terms  
         & A tutorial on classical \acsp{dmp} that presents both discrete and rhythmic formulations, mostly developed in \citep{Ijspeert2002Movement, Ijspeert2002Learning, ijspeert2002learningAttractor}, and their application in \ac{il} and movement recognition. The tutorial also presents extensions of the classical \ac{dmp} formulation to prevent high accelerations at the beginning of the motion, to avoid collisions with unforeseen obstacles \citep{Pastor2009}, and to generalize both in space (\eg reach a different goal) and time (e.g., produce longer/shorter trajectories). \\
	\midrule
	\citep{pastor2013dynamic} & $\bullet$ Classical \acsp{dmp} \newline
            									$\bullet$ Online adaptation  \newline
            									$\bullet$ Coupling terms \newline
            									$\bullet$ Impedance learning
            									 & A tutorial on classical \acsp{dmp} that presents both discrete and rhythmic formulations, mostly developed in \citep{Ijspeert2002Movement, Ijspeert2002Learning, ijspeert2002learningAttractor}. The tutorial also presents extensions of the classical \ac{dmp} formulation to avoid collisions with unforeseen obstacles \citep{Pastor2009} and to learn impedance control policies via \ac{rl} \citep{buchli2010variable}. The key difference between this tutorial and the one from \citep{Ijspeert2013Dynamical} is the section dedicated to sensory association and online, context-aware adaptation of \ac{dmp} trajectories using the associative skill memory framework developed in \citep{Pastor2011skill, Pastor2011Online}.\\
	\midrule
	\citep{denivsa2016review} & $\bullet$ Classical \acsp{dmp} \newline
            									$\bullet$ \acsp{cmp} & A tutorial on \acsp{cmp}, a framework developed to generate compliant robot behaviors that accurately track a reference trajectory. \acsp{cmp} exploit classical \acsp{dmp} to generate the desired kinematic landscape and encode task-dependent dynamics as a combination of Gaussian basis functions (torque primitives). The tutorial show how to learn torque primitives from training data, how to generalize \acsp{cmp} to new situations, and how to combine existing \acsp{cmp} to synthesize new robot motions. \\
	\bottomrule
\end{tabular}
\\[2pt]
\begin{tabular}{m{0.17\textwidth}m{0.2\textwidth}m{0.55\textwidth}}
    \toprule
    \textbf{Survey and Tutorial} & \textbf{Topics} & \textbf{Description} \\
	\midrule
    This paper  & \ac{dmp} tutorial \newline
            $\bullet$ Classical \newline
            $\bullet$ Orientation  \newline
            $\bullet$ SPD \newline
            $\bullet$ Joining \newline
            $\bullet$ Generalization \newline
            $\bullet$ Online adaptation \newline  \newline
        \ac{dmp} survey \newline
            $\bullet$ (Co-)Manipulation \newline
            $\bullet$ Variable impedance \newline
            $\bullet$ Physical interaction \newline
            $\bullet$ Rehabilitation \newline
            $\bullet$ Teleoperation \newline
            $\bullet$ Motion recognition \newline
            $\bullet$ Reinforcement, deep, and lifelong learning
            & This tutorial survey conducts a wide scan of the existing \ac{dmp} literature with the aim of categorizing and presenting the published work in the field. The main objective of this comprehensive literature review is give the reader an exhausting overview on \ac{dmp} related research, on its major achievements, as well as on open issues and possible research directions. Our tutorial survey also provides a structured and unified formulation for different methods developed starting from the classical \acp{dmp} proposed by \citep{Ijspeert2002Movement, Ijspeert2002Learning}. We believe that such formulation contributes to easier the understanding of different methods and extension that can be found in the literature, clarifying connections and differences among the existing approaches. The tutorial survey also provides an analysis on pros and cons of various methods and a discussion with guidelines for different application scenarios. \\ 
    \bottomrule
\end{tabular}

%% file: taxonomy.tex
\usetikzlibrary{calc}
\tikzset{
  basic/.style  = {draw, text width=5cm, font=\sffamily\footnotesize, rectangle},
  root/.style   = {basic, rounded corners=2pt, thin, align=center, fill=white},
  level-2/.style = {basic, rounded corners=4pt, thin,align=center, fill=white, text width=2.5cm},
  level-3/.style = {basic, rounded corners=4pt, thin,align=center, fill=white, text width=2cm},
  level-4/.style = {basic, thin, align=center, fill=white, text width=1.8cm},
  level-discussion/.style = {basic, rounded corners=4pt, thin,align=center, fill=white, text width=0.8\textwidth}
}

\begin{tikzpicture}[
  level 1/.style={sibling distance=23em, level distance=4em},
  level 2/.style={sibling distance=9em, level distance=4em},
  level 3/.style={sibling distance=5em, level distance=3em},
  edge from parent/.style={->,solid,black,thick,sloped,draw}, 
  edge from parent path={(\tikzparentnode.south) -- (\tikzchildnode.north)},
  >=latex, node distance=1.2cm, edge from parent fork down]

\node[root] (c0) {\textbf{Dynamic Movement Primitives (DMPs)}}
  child {node[level-2] (c1) {\textbf{Tutorial} \\ Sections \ref{sec:formulation}
 and \ref{sec:advanceMethods}}
    child {node[level-3] (c11) {\textbf{Formulation}}}
    child {node[level-3] (c12) {\textbf{Extentions}}}
  }
  child {node[level-2] (c2) {\textbf{Survey} \\ Sections \ref{sec:integration} and \ref{sec:applications}}
    child {node[level-3] (c21) {\textbf{Integration}}}
    child {node[level-3] (c22) {\textbf{Applications}}}
   };

 \begin{scope}[every node/.style={level-4}]
 \node [below of = c11, xshift=10pt, yshift=5pt] (c111) {Classical discrete and periodic};
 \node [below of = c111, yshift=4pt] (c112) {Orientation};
 \node [below of = c112, yshift=4pt] (c113) {Symmetric positive definite matrix};
 
 \node [below of = c12, xshift=10pt, yshift=5pt] (c121) {Generalization};
 \node [below of = c121, yshift=14pt] (c122) {Joining};
 \node [below of = c122, yshift=8pt] (c123) {On-line adaptation};
 \node [below of = c123, yshift=4pt] (c124) {Alternative formulations};

 \node [below of = c21, xshift=10pt] (c211) {Manipulation tasks};
 \node [below of = c211, yshift=4pt] (c212) {Variable impedance};
 \node [below of = c212, yshift=3pt] (c213) {Reinforcement learning};
 \node [below of = c213, yshift=3pt] (c214) {Deep \\ learning};
 \node [below of = c214, yshift=3pt] (c215) {Life-long learning};
 
 \node [below of = c22, xshift=10pt] (c221) {Contact with passive environments};
 \node [below of = c221, yshift=-6pt] (c222) {Human--robot co-manipulation};
 \node [below of = c222, yshift=-16pt] (c223) {Human assistance, augmentation \& rehabilitation};
 \node [below of = c223, yshift=-6pt] (c224) {Teleoperation};
 \node [below of = c224, yshift=9pt] (c225) {High degrees of freedom};
 \node [below of = c225, yshift=-1pt] (c226) {Motion analysis and recognition};
 \node [below of = c226, yshift=-7pt] (c227) {Autonomous driving and field robotics};
 
 \end{scope}

 \begin{scope}
   \node [below of = c0, yshift=-360pt, style={level-discussion}] (c216) {\textbf{Discussion} (Section \ref{sec:discussion})};
   \node [below of = c216, xshift=-120pt, yshift=0pt, style={level-4}] (c217) {Guidelines};
	 \node [below of = c216, xshift=0pt, yshift=0pt,style={level-4}] (c218) {Available code};
 	 \node [below of = c216, xshift=120pt, yshift=0pt,style={level-4}] (c219) {Open issues};

 \end{scope}

  \draw[<-] (c217.north) -- (c217.north|-c216.south);
  \draw[<-] (c218.north) -- (c218.north|-c216.south);
  \draw[<-] (c219.north) -- (c219.north|-c216.south);
  
  \draw[dashed,->] ([yshift=-85pt]c11.193) -- (c11.193|-c216.north);
  \draw[dashed,->] ([yshift=-101pt]c12.193) -- (c12.193|-c216.north);
  \draw[dashed,->] ([yshift=-152pt]c21.195) -- (c21.195|-c216.north);
  \draw[dashed,->] ([yshift=-260pt]c22.195) -- (c22.195|-c216.north);

  \draw[->] (c11.193) |- (c111.west);
  \draw[->] ([yshift=-23pt]c11.193) |- (c112.west);
  \draw[->] ([yshift=-53pt]c11.193) |- (c113.west);

  \draw[->] (c12.193) |- (c121.west);
  \draw[->] ([yshift=-23pt]c12.193) |- (c122.west);
  \draw[->] ([yshift=-43pt]c12.193) |- (c123.west);
  \draw[->] ([yshift=-69pt]c12.193) |- (c124.west);
   
  \draw[->] (c21.195) |- (c211.west);
  \draw[->] ([yshift=-27pt]c21.195) |- (c212.west);
  \draw[->] ([yshift=-57pt]c21.195) |- (c213.west);
  \draw[->] ([yshift=-88pt]c21.195) |- (c214.west);
  \draw[->] ([yshift=-120pt]c21.195) |- (c215.west);
   
  \draw[->] (c22.195) |- (c221.west);
  \draw[->] ([yshift=-27pt]c22.195) |- (c222.west);
  \draw[->] ([yshift=-67pt]c22.195) |- (c223.west);
  \draw[->] ([yshift=-117pt]c22.195) |- (c224.west);
  \draw[->] ([yshift=-157pt]c22.195) |- (c225.west);
  \draw[->] ([yshift=-183pt]c22.195) |- (c226.west);
  \draw[->] ([yshift=-218pt]c22.195) |- (c227.west);
  
\end{tikzpicture}

%% file: DMP_Formulation.tex
\begin{table*}[t]
	\small\sf\centering
	\caption{Description of key notations and abbreviations. Indices, super/subscripts, constants, and variables have the same meaning over the whole text.}
	\input{notation.tex}
	\label{tab:notation}
\end{table*} 

\section{Formulation of \texorpdfstring{\acp{dmp}}{} types}
\label{sec:formulation}

In this section, we will provide a complete description of the standard formulation of \acp{dmp}. Specifically, point attractors formulation---to encode discrete point-to point motions---in \secref{subsec:discrete}, and cycle attractors formulation---to encode rhythmic-patterns motions---in \secref{subsec:periodic}. For a better understanding, we have summarized the key notations and the used abbreviations in \tabref{tab:notation}.

\subsection{Discrete \texorpdfstring{\acs{dmp}}{}}
\label{subsec:discrete}

The discrete \ac{dmp} is used to encode a point-to-point motion into a stable dynamical system. In the following subsections, we will go through the formulation and main features of discrete \acp{dmp} starting by the classical one operating in $\mathbb{R}$ space  (\secref{subsec:discrete:classical}), then passing by Cartesian space---$\bm{\mathcal{S}}^3$ and $\bm{\mathcal{SO}}(3)$---in \secref{subsec:discrete:rotation}, and ending by \ac{dmp} formulation for \ac{spd} space ($\spd$) in \secref{subsec:discrete:SPD}. 

\subsubsection{Classical \texorpdfstring{\acs{dmp}}{}}
\label{subsec:discrete:classical}

The classical discrete \acp{dmp} were first introduced by \cite{Ijspeert2002Movement}. A \ac{dmp} for a single \ac{dof} trajectory $y$ of a discrete movement (point-to-point) is defined by the following set of nonlinear differential equations \citep{Ijspeert2002Movement,Ijspeert2013Dynamical}
\begin{align}
\tau \dz &= \alpha_z(\beta_z(g-y)-z) + f(x), \label{eq:dmp_discrete1}\\
\tau \dy &= z, \label{eq:dmp_discrete2}\\
\tau \dx &= \alpha_x x, \label{eq:dmp_discrete3}
\end{align}
where $x$ is the phase variable and $z$ is an auxiliary variable. Parameters $\alpha_z$ and $\beta_z$ define the behavior of the second order system described by \eqref{eq:dmp_discrete1} and \eqref{eq:dmp_discrete2}. With the choice $\tau > 0$, $\alpha_z = 4\beta_z$ and $\alpha_x > 0$, the convergence of the underlying dynamic system to a unique attractor point at $y = g$, $z = 0$ is ensured \citep{Ijspeert2013Dynamical}. Alternatively, the gains $\alpha_z$ and $\beta_z$ can be learned from training data while preserving the convergence of the system \citep{Tan2016Applying}. In the \ac{dmp} literature, equations \eqref{eq:dmp_discrete1}--\eqref{eq:dmp_discrete2}, as well as their periodic counterpart \eqref{eq:dmp_periodic1}--\eqref{eq:dmp_periodic2}, are called the \textit{transformation system}, while \eqref{eq:dmp_discrete3} (or \eqref{eq:dmp_periodic3}) is the \textit{canonical system}.
$f(x)$ is defined as a linear combination of $N$ nonlinear \acp{rbf}, which enables the robot to follow any smooth trajectory from the initial position $y_0$ to the final configuration $g$
\begin{align}
f(x) &= \frac{\sum_{i=1}^Nw_i\Psi_i(x)}{\sum_{i=1}^N\Psi_i(x)}x, \label{eq:fx_discrete} \\
\Psi_i(x) &= \exp\left(-h_i\left(x-c_i\right)^2\right), \label{eq:psi_discrete}
\end{align}
where $c_i$ are the centers of Gaussian basis functions distributed along the phase of the movement and $h_i$ their widths. For a given $N$ and setting $\tau$ equal to the total duration of the desired movement, we can define $c_i=\text{exp}\big(-\alpha_x\frac{i-1}{N-1}\big)$, $h_i=\frac{1}{(c_{i+1} - c_i)^2}$ and $h_N=h_{N-1}$ where $i = 1,\dots,N$. For each \ac{dof}, the weights $w_i$ should be adjusted from the measured data so that the desired behavior is achieved. The selection of the number of weights should be based on the desired resolution of the trajectory. For controlling a robotic system with more than one \ac{dof}, we represent the movement of every \ac{dof} with its own equation system \eqref{eq:dmp_discrete1}--\eqref{eq:dmp_discrete2}, but with the common phase \eqref{eq:dmp_discrete3} to synchronize them.

\begin{figure}[t]
 	\centering
 	\includegraphics[width=\linewidth]{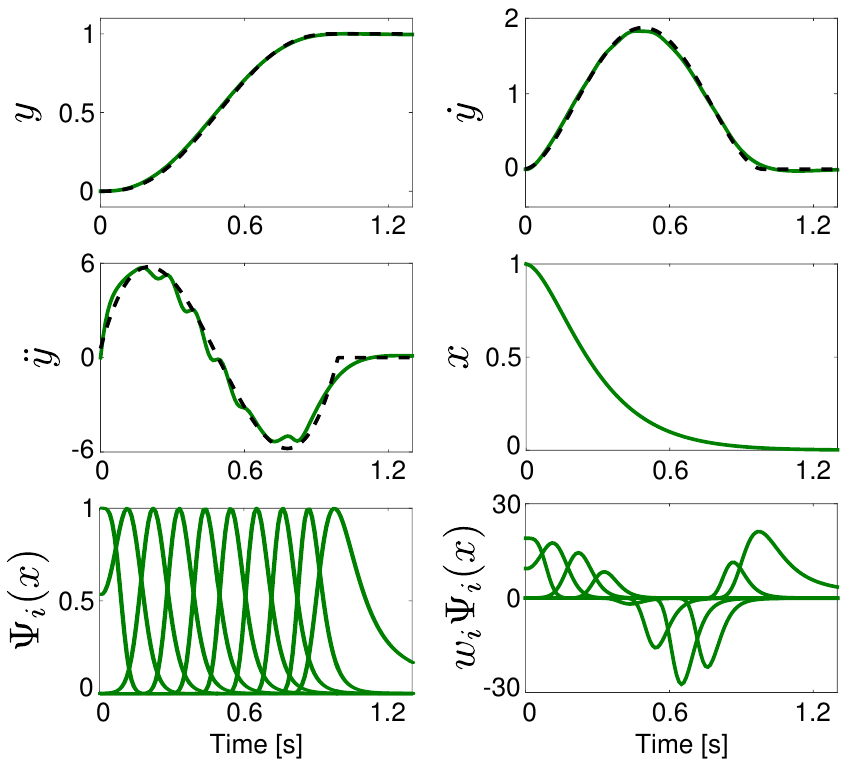}
 	\caption{A classical \ac{dmp} is used to generate a discrete motion connecting $x_0 = 0$ and $g=1$ (green line in the top left panel). The training data (black dashed lines) are obtained from a minimum jerk trajectory connecting $x_0$ and $g$ in $T=1\,$s and used to learn the weights $w_i$ of $10$ Gaussian basis functions equally distributed in time. The results of the parameters learning procedure are shown in the bottom right panel. The exponentially decaying phase variable is used as shown in the middle right panel. Results are obtained with the open source implementation available at \url{www-clmc.usc.edu/software/git/gitweb.cgi?p=matlab/dmp.git}. }
    \label{fig:classic_dmp_example}
\end{figure}

\paragraph{Learning the forcing term}\label{par:learning_forcing}
For a discrete motion, given a demonstrated trajectory $y_d(t_\jmath)$, $t_\jmath = 1,\ldots,\mathfrak{T}$ and its time derivatives $\dot{y}_d(t_\jmath)$ and $\ddot{y}_d(t_\jmath)$, it is possible to invert~\eqref{eq:dmp_discrete1} and approximate the desired shape $f_d$ as
\begin{equation}
f_d(t_\jmath) = \tau^2 \ddot{y}_d(t_\jmath)  -  \alpha_z \left( {\beta_z  \left(g - y_d(t_\jmath)\right)  - \tau \dot{y}_d(t_\jmath)  } \right).
\label{eq:forcingterm}
\end{equation}
By stacking each $f_d(t_\jmath)$ and $w_i$ into the column vectors  $\mathfrak{F} = 
	\begin{bmatrix}
		f_d(t_1),
		\ldots,
		f_d(t_\mathfrak{T})
	\end{bmatrix}\trsp$ and $\bm{w} = 
	\begin{bmatrix}
		w_1, \ldots, w_N
	\end{bmatrix}\trsp$,
we obtain the following linear system
\begin{equation}
	\bm{\Upphi}\bm{w}=\mathfrak{F},
\end{equation}
where
\begin{equation}
	\bm{\Upphi} = 
	\begin{bmatrix}
		\frac{\Psi_1(x_1)}{\sum_{i=1}^N\Psi_i(x_1)}x_1 & \cdots & \frac{\Psi_N(x_1)}{\sum_{i=1}^N\Psi_i(x_1)}x_1 \\
		\vdots & \ddots & \vdots \\
		\frac{\Psi_1(x_\mathfrak{T})}{\sum_{i=1}^N\Psi_i(x_\mathfrak{T})}x_\mathfrak{T} & \cdots & \frac{\Psi_N(x_\mathfrak{T})}{\sum_{i=1}^N\Psi_i(x_\mathfrak{T})}x_\mathfrak{T}
	\end{bmatrix}.
\end{equation}
\ac{lwr}~\citep{atkeson1997locally,Schaal1998,ude2010task} is a popular approach used to update the weights $w_{i}$. \ac{lwr} uses the error between the desired trajectory shape and currently learned shape and a \textit{forgetting factor} $\lambda$ to update the weights as 
\begin{align}
\bm{P}_{\jmath}&=\frac{1}{\lambda}\left(\bm{P}_{\jmath-1}-\frac{\bm{P}_{\jmath-1}\bm{\varphi}_\jmath\bm{\varphi}_\jmath\trsp\bm{P}_{\jmath-1}}{\lambda+\bm{\varphi}_\jmath\trsp\bm{P}_{\jmath-1}\bm{\varphi}_\jmath}\right),\label{eq:discrete:reg1}\\
\bm{w}_\jmath &= \bm{w}_{\jmath-1} +(f_d(t_\jmath) - \bm{\varphi}_\jmath\trsp \bm{w}_{\jmath-1})\bm{P}_\jmath\bm{\varphi}_\jmath. \label{eq:discrete:reg2}
\end{align}
In the previous equations $\bm{w}_\jmath = \bm{w}(t_{\jmath})$ and $\bm{\varphi}_\jmath$ is the column vector obtained by transposing the $\jmath$-th row of $\bm{\Upphi}$. The initial value of the parameters is $\bm{P}_0 = \bm{I}, \bm{w}_0 = \bm{0}$.  A discrete \ac{dmp} learned on synthetic data is shown in \Figref{fig:classic_dmp_example}.

\ac{lwr} has been the standard method to learn the weights of \acp{dmp} and therefore $f(x)$. As an alternative to \ac{lwr}, \citep{krug2013representing} have shown that learning a forcing term defined as in \eqref{eq:fx_discrete} can be formulated as a quadratic optimization problem and efficiently solved.  

In general, the problem of learning and retrieving $f(x)$ can be in principle solved with any regression technique \citep{Stulp13_hum}. For instance, \cite{Wang2016Dynamic} modified $f(x)$ in \eqref{eq:fx_discrete} by considering a bias term $b_i$, \ie $w_ix + b_i$, and used truncated kernels ($\Psi_i$ vanishes if $x - c_i$ is smaller than a threshold). This formulation, called \ac{dmp}+, produces more accurate trajectories than the original \ac{dmp}. Moreover, a learned trajectory can be modified by updating only a subset of the weights. Other work focused on using multiple demonstrations to increase the generalization power of the learned primitive. To learn a suitable forcing term from multiple demonstrations, some authors used \ac{gmm} \citep{yin2014learning,Pervez2017Novel} and \ac{gmr} \citep{cohn1996active}, while others adopted \ac{gp} \citep{Fanger2016Gaussian, Umlauft2017Bayesian} \citep{Rasmussen2006}, or exploited a deep \ac{nn} \citep{Pervez2017, Pahic2020} developed originally in \citep{lecun2015deep}.

\paragraph{Phase stopping and goal switching}\label{par:phase_stopping}
The phase variable $x$ in \eqref{eq:dmp_discrete3} provides the ability to manipulate time during the execution of \ac{dmp} equations. Moreover, \ac{dmp} provides the ability to slow-down or even stop the execution through the phase-stopping mechanism \citep{Ijspeert2002Movement}
\begin{equation}
\tau \dx = - \frac{\alpha_x\, x}{1+\alpha_{yx}||\tilde{y}-y||}
\label{eq:phaseStopping}
\end{equation}

Moreover, \acp{dmp} provide an elegant way to adapt the trajectory generation in real-time  through goal switching mechanisms \citep{Ijspeert2013Dynamical}
\begin{equation}
\tau {\Dot{g}} = \alpha_g(g_0 - g)
\label{eq:goalSwitching}
\end{equation}

\acp{dmp} in its standard formulation are not suitable for direct encoding of skills with specific geometry constraints, such as orientation profiles (represented in either unit quaternions or rotation matrices), stiffness/damping and manipulability profiles (encapsulated in full \ac{spd} matrices). For instance, direct integration of unit quaternions, does not ensure the unity of the quaternions norm. Any representation of orientation that does not contain singularities is non-minimal, which means that additional constraints need to be taken into account during integration.

\begin{figure}[t]
	\centering
	\includegraphics[width=\linewidth]{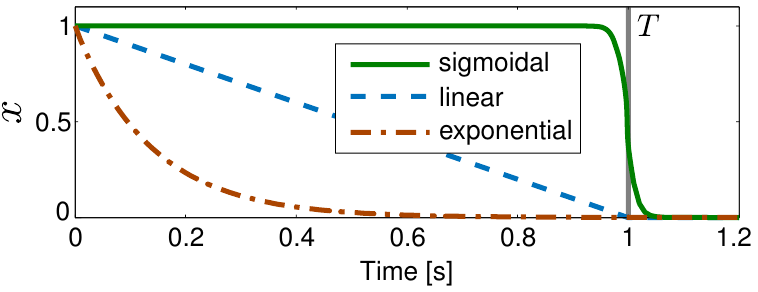}
    \caption{Possible phase variables used in different discrete \ac{dmp} formulations. All the different possibilities ensure that $x, s, p \rightarrow 0$ for $t \rightarrow +\infty$ (for $t > T$ in practice).}
    \label{fig:phase_dmp}
\end{figure} 

\paragraph{Alternative phase variables}
\label{par:phase_variables}

Equation \eqref{eq:dmp_discrete3} describes an exponential decaying phase variable that has been widely used in the \ac{dmp} literature. The main drawback of the exponential decaying phase is that it rapidly drops to very small values towards the end of the motion. This ``forces'' the learning algorithm to exploit relatively high weights $w_i$ to accurately reproduce the last part of the demonstration \citep{Samant2016Adaptive}. As an example, in \Figref{fig:phase_dmp}  the exponential decaying phase (brown dot-dashed line) is very small already after $0.6\,$s, while the expected time duration of the motion is $T=1\,$s. 

To overcome this limitation, \cite{kulvicius2011modified} propose the sigmoidal decay phase $s$ (green solid line in \Figref{fig:phase_dmp}), obtained by integrating
\begin{equation} 
\dot{s} = - \frac{\alpha_{ s }e^{({ \alpha_{s} }/{ \delta_{ t }})(\tau T-t)}}
{[ 1 + e^{({ \alpha_{s} }/{ \delta_{ t }})(\tau T-t)}]^{ 2 }},
 \label{eq:sigmoidal_decay}
\end{equation}
where~$ \alpha_{ s }$ defines the steepness of $s$ centered at time $T$ and $\delta t$ is the sampling time. As shown in \Figref{fig:phase_dmp}, $s = 1$ for $t < T -\delta_s$, where the time $\delta_s$ depends on the steepness $\alpha_{ s }$, and then it decays to~$s=0$.

The sigmoidal decay in \Figref{fig:phase_dmp} has a tail effect since it vanishes after $T+\delta_s\,$s, where $\delta_s$ depends on the tunable parameter $\alpha_{ s }$. The piece-wise linear phase $l$ (blue dashed line in \Figref{fig:phase_dmp}), proposed by \citep{Samant2016Adaptive}, linearly decays from $1$ to $0$ in exactly $T\,$s and then remains constant. $p$ is obtained by integrating
\begin{equation} 
\tau \dot{p}= 
\begin{cases}
-\frac{1}{T}, & p \geq 0\\
0, &\text{otherwise}
\end{cases}
\label{eq:linear_decay}
\end{equation}
where $p(0) = 1$ and $T$ is the time duration of the motion.

\subsubsection{Orientation  \texorpdfstring{\ac{dmp}}{}}
\label{subsec:discrete:rotation}
The classical \ac{dmp} formulation described in Section~\ref{subsec:discrete:classical} applies to single \ac{dof} motions. Multidimensional motions are generated independently and synchronized with a common phase. In other words, equations \eqref{eq:dmp_discrete1} and \eqref{eq:dmp_discrete2} are repeated for each \ac{dof} while the phase variable in~\eqref{eq:dmp_discrete3} is shared. This works when the evolution of different \ac{dof} is independent, like for joint space or Cartesian position trajectories. Unlike Cartesian position, the elements of orientation representations like unit quaternion or rotation matrix are constrained. In this section, we present approaches that extend the classical \ac{dmp} formulation to represent Cartesian orientations.

\paragraph{Quaternion \texorpdfstring{\ac{dmp}}{}}
\label{subsubsec:discrete:quaternion}

\begin{figure}[t]
 \includegraphics[width=\columnwidth]{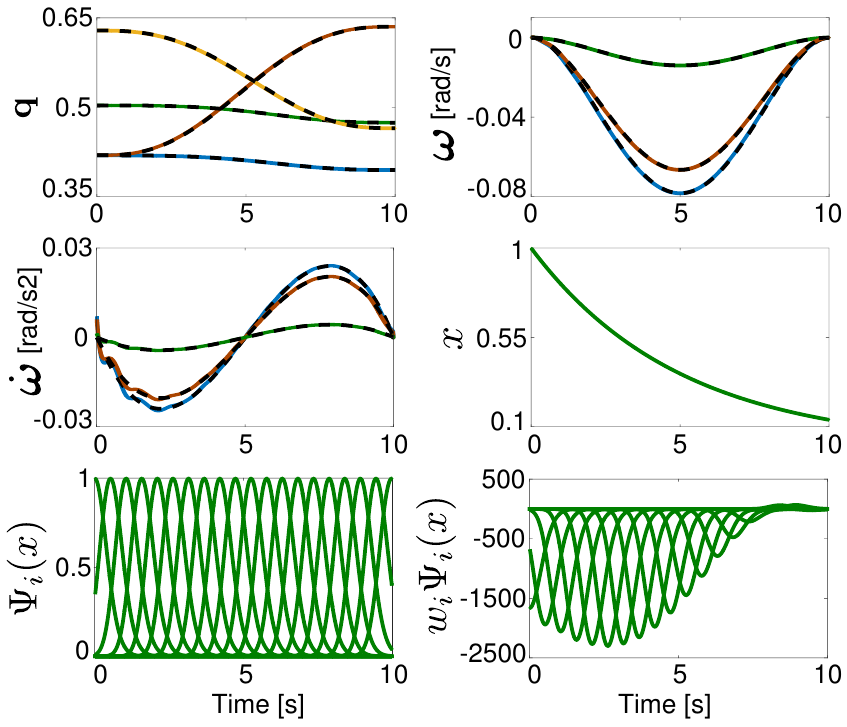}
	\caption{A unit quaternion \ac{dmp} is used to generate a discrete motion connecting $\q_1$ and $\bm{g}_q$. The training data (black dashed lines) are obtained from a minimum jerk trajectory connecting $\q_1$ and $\bm{g}_q$ in $T=10\,$s and used to learn the weights $w_i$ of $20$ Gaussian basis functions equally distributed in time. The results of the parameters learning procedure are shown in the bottom right panel. The exponentially decaying phase variable is used as shown in the middle right panel. Results are obtained with the open source implementation available at \url{https://gitlab.com/dmp-codes-collection}. }
    \label{fig:quaternion_dmp_example}
\end{figure} 

Unit quaternion $\q = \upnu+\bm{u} \in \bm{\mathcal{S}}^3$ provides a representation of the orientation of the robot's end-effector \citep{chiaverini1999unit}. $\bm{\mathcal{S}}^3$ is a unit sphere in $\mathbb{R}^4$, $\upnu\in \mathbb{R}$, and $\bm{u}\in \mathbb{R}^3$. 
\cite{AbuDakka2015Adaptation} rewrote \ac{dmp} equations \eqref{eq:dmp_discrete1} and \eqref{eq:dmp_discrete2} for direct unit quaternion encoding as follows
\begin{align}
\tau \bm{\Dot{\eta}} &= \alpha_z(\beta_z2 \, \LogQ(\bm{g}_q*\overline{\q})-\bm{\eta}) + \bm{f}_q(x), \label{eq:quat:dw}\\
\tau \dq &= \frac{1}{2}\bm{\eta}*\q, \label{eq:quat:dq}
\end{align}
where $\bm{g}_q \in \bm{\mathcal{S}}^3$ denotes the goal orientation, the quaternion conjugation is defined as $\overline{\q}=\overline{\upnu + \bm{u}}=\upnu - \bm{u}$, and $*$ denotes the the quaternion product
\begin{align*}
\begin{split}
\q_1 * \q_2 &= (\upnu_1 + \bm{u}_1) * (\upnu_2 + \bm{u}_2)\\
&= (\upnu_1 \upnu_2 - \bm{u}\trsp_1 \bm{u}_2) + (\upnu_1 \bm{u}_2 + \upnu_2 \bm{u}_1 + \bm{u}_1 \times \bm{u}_2).
\end{split}
\end{align*}
$\bm{\eta}\in \mathbb{R}^3$ is the scaled angular velocity $\bm{\omega}$ and treated as unit quaternion with zero scalar $(\upnu=0)$ in \eqref{eq:quat:dq}. The function $\LogQ(\cdot):\bm{\mathcal{S}}^3\mapsto\mathbb{R}^3$ is given as
\begin{equation}
\LogQ(\q) = 
\left\{ 
\begin{aligned}
&\arccos(\upnu) \frac{\bm{u}}{||\bm{u}||}, && \bm{u}\neq \bm{0} \\
&[0\;\; 0\;\; 0]{\trsp}, && \text{otherwise},
\end{aligned}  
\right.
\label{eq:quat:log}
\end{equation}
where $||\cdot||$ denotes $\ell_2$ norm. 

Early attempt to encode unit quaternion profiles using \ac{dmp} was presented by \cite{Pastor2011Online}. Unlike \citeauthor{AbuDakka2015Adaptation}'s formulation, \citeauthor{Pastor2011Online}'s does not take into account the geometry of $\bm{\mathcal{SO}}(3)$ as they just used the vector part of the quaternion product $(\bm{g}_q*\overline{\q})$ in \eqref{eq:quat:dw} instead of $2 \, \LogQ(\bm{g}_q*\overline{\q})$ which defines the angular velocity $\bm{\omega}$ that rotates quaternion $\q$ into $\bm{g}_q$ within a unit sampling time.

Equation \eqref{eq:quat:dq} can be integrated as 
\begin{equation}
{\q}(t+\delta t)=\ExpQ\left(\frac{\delta t}{2}\frac{\bm{\eta}(t)}{\tau}\right)*\q(t),
\end{equation}
where $\delta_t>0$ denotes a small constant. The function $\ExpQ(\cdot):\mathbb{R}^3\mapsto\bm{\mathcal{S}}^3$ is given
\begin{equation}
\ExpQ(\bm{\omega})=
\left\{ 
\begin{aligned}
&\cos(||\bm{\omega}||)+\sin(||\bm{\omega}||)\frac{\bm{\omega}}{||\bm{\omega}||}, && \bm{\omega}\neq \bm{0} \\
&1+[0\;\; 0 \,\;0]{\trsp}, && \text{otherwise}.
\end{aligned}  
\right.
\label{eq:quat:exp}
\end{equation}

Both mappings become one-to-one, continuously differentiable and inverse to each other if the input domain of the mapping $\LogQ(\cdot)$ is restricted to $ \bm{\mathcal{S}}^3$ except for $-1 + [0\, 0\, 0]\trsp$, while the input domain of the mapping $\ExpQ(\bm{\omega})$ should fulfill the constraint $||\bm{\omega}||<\pi$ \citep{AbuDakka2015Adaptation}. An exemplar unit quaternion \ac{dmp} is shown in \Figref{fig:quaternion_dmp_example}.

Phase-stopping \eqref{eq:phaseStopping} can be rewritten as follows
\begin{align}
\tau \dx = - \frac{\alpha_x\, x}{1+\alpha_{qx}\text{d}(\tilde{\q},\q)}
\label{eq:quat:phaseStopping}
\end{align}
where 
\begin{align*}
\text{d}(\tilde{\q},\q)=
\left\{ 
\begin{aligned}
&2\,\pi, && \q_1*\overline{\q_2} = 1+[0\;\; 0 \,\;0] \\
&2\,|\LogQ(\q_1*\overline{\q_2})||,&& \text{otherwise}
\end{aligned}
\right.
\end{align*}

\cite{Ude2014Orientation} extended \ac{dmp} quaternions-based formulation by rewriting \eqref{eq:goalSwitching} to include goal switching mechanism. 
\begin{equation}
\tau \bm{\Dot{g}}_q = \alpha_{qg}\LogQ(\bm{g}_{q,new} - \overline{\bm{g}}_q) * \bm{g}_q
\label{eq:quat:goalSwitching}
\end{equation}
so that $\bm{g}_q$ is continuously changing onto $\bm{g}_{q,new}$ in real-time. Equation \eqref{eq:quat:goalSwitching} should be integrated using \eqref{eq:quat:exp} along with \eqref{eq:quat:dw} and \eqref{eq:quat:dq}.

As shown by \cite{saveriano2019merging} using Lyapunov arguments, both the quaternion \ac{dmp} formulations in \citep{Pastor2011Online} and in \citep{AbuDakka2015Adaptation, Ude2014Orientation} asymptotically converge to the target quaternion $\bm{g}_q$ with zero velocity. 

\paragraph{Rotation Matrix \texorpdfstring{\ac{dmp}}{}}
\label{subsubsec:discrete:rotMatrix}

\begin{figure}[t]
 \includegraphics[width=\columnwidth]{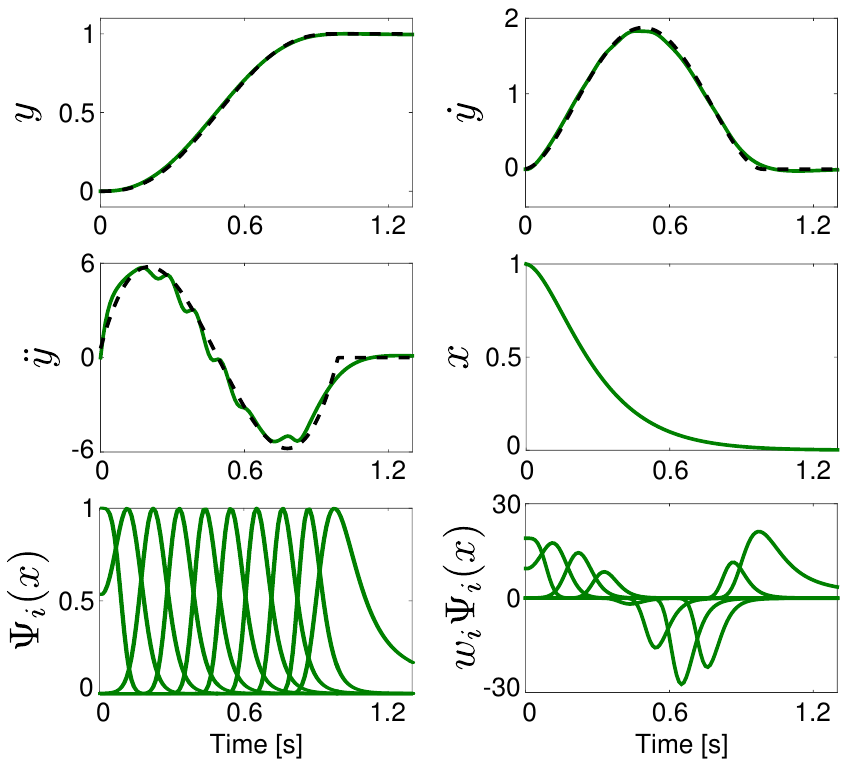}
	\caption{A rotation matrix \ac{dmp} is used to generate a discrete motion connecting $\R_1$ and $\R_g$. The training data (black dashed lines) are obtained from a minimum jerk trajectory connecting $\R_1$ and $\R_g$ in $T=10\,$s and used to learn the weights $w_i$ of $20$ Gaussian basis functions equally distributed in time. The results of the parameters learning procedure are shown in the bottom right panel. The exponentially decaying phase variable is used as shown in the middle right panel.} 
    \label{fig:rotation_matrix_dmp_example}
\end{figure}

In their work on orientation \acp{dmp}, \cite{Ude2014Orientation} extended \acp{dmp} formulation in order to encode orientation trajectories represented in the form of rotation matrices $\R(t)\in\bm{\mathcal{SO}}(3)$. Therefore, they rewrote \eqref{eq:dmp_discrete1} and \eqref{eq:dmp_discrete2} in the form
\begin{align}
\tau \bm{\Dot{\eta}} &= \alpha_z(\beta_z \, \LogR(\R_g\R\trsp)-\bm{\eta}) + \bm{f}_R(x), \label{eq:rot:dw}\\
\tau \dR &= [\bm{\eta}]_\times \R, \label{eq:rot:dq}
\end{align}
where $\R_g$ represents the goal orientation. 
$[\bm{\eta}]_\times$ is a skew symmetric matrix, such as $[\bm{\eta}]_\times = -[\bm{\eta}]_\times$.
The relation between the angular velocity and 1st-time-derivative of the rotation matrix is given by
\begin{equation}
    [\bm{\omega}]_\times =
    \begin{bmatrix}
        0   &   -\omega_z   &   \omega_y    \\
        \omega_z    &   0   &   -\omega_x   \\
        -\omega_y   &   \omega_x    &   0   
    \end{bmatrix}
    = \dot{\R}\R\trsp.
\end{equation}
The function $\LogR(\cdot):\bm{\mathcal{SO}}(3)\mapsto\mathbb{R}^3$ is given as
\begin{align}
\LogR(\R) = 
\left\{ 
\begin{aligned}
&[0, 0, 0]\trsp, && \R = \bm{I} \\
&\omega = \theta \bm{n}, && \text{otherwise},
\end{aligned}  
\right.\label{eq:rot:log}
\end{align}
\begin{equation*}
    \theta = \arccos{\left(\frac{\text{trace}(\R)-1}{2}\right)},\ 
\bm{n}=\frac{1}{2\sin{(\theta)}}
\begin{bmatrix}
r_{32} - r_{23}\\
r_{13} - r_{31}\\
r_{21} - r_{12}
\end{bmatrix}
\end{equation*}

The generated rotation matrices can be obtained by integrating \eqref{eq:rot:dq} as follows
\begin{equation}
    \R(t+\delta t) = \ExpR\left(\delta t\frac{[\bm{\eta}]_\times}{\tau}\right)\R(t).
\end{equation}
The function $\ExpR(\cdot):\mathbb{R}^3\mapsto\bm{\mathcal{SO}}(3)$ is given as
\begin{equation}
\begin{split}
\ExpR\left(t[\bm{\omega}]_\times\right)=
\bm{I} 
&+ \sin(\theta)\frac{[\bm{\omega}]_\times}{||\bm{\omega}||}\\
&+ (1-cos(\theta))\frac{[\bm{\omega}]_\times^2}{||\bm{\omega}||^2}.
\end{split}  
\label{eq:rot:exp}
\end{equation}
where $\theta(t)=t||\bm{\omega}||$ express the rotation angle within time $t$. An exemplar rotation matrix \ac{dmp} is shown in \Figref{fig:rotation_matrix_dmp_example}.




\subsubsection{\texorpdfstring{\ac{spd}}{} matrices}
\label{subsec:discrete:SPD}

\begin{figure}[t]
	\centering
	\includegraphics[width=\linewidth]{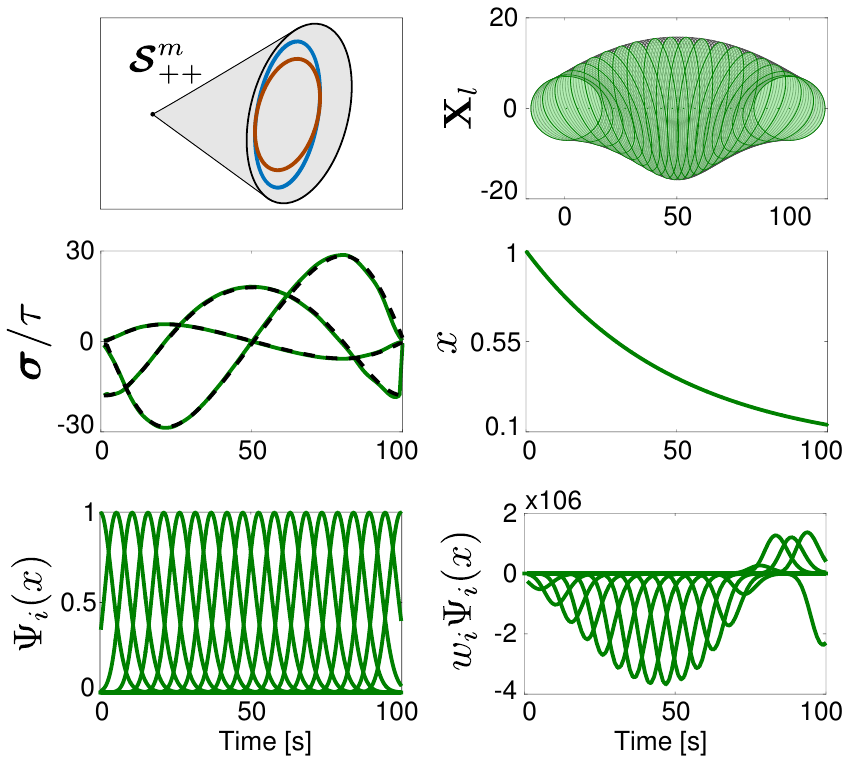}
	\caption{An \ac{spd} \ac{dmp} is used to generate a discrete motion connecting $\X_1$ and $\X_g$. The training data (black dashed lines) are obtained from a minimum jerk trajectory connecting $\X_1$ and $\X_g$ in $T=100\,$s and used to learn the weights $w_i$ of $20$ Gaussian basis functions equally distributed in time. The cone in the upper left corner represents the manifold of \ac{spd} data and includes the geodesic of the \ac{spd} profile.
	The results of the parameters learning procedure are shown in the bottom right panel. The exponentially decaying phase variable is used as shown in the middle right panel. Results are obtained with the open source implementation available at \url{https://gitlab.com/dmp-codes-collection}.}
    \label{fig:spd_dmp_example}
\end{figure} 
\cite{abudakka2020Geometry} generalized \ac{dmp} formulation in order to encode robotic manipulation data profiles encapsulated in form of \ac{spd} matrices. By defining $\X \in \spd$ as an arbitrary \ac{spd} matrix and $\bm{\Xi} = \{t_\jmath, \X_\jmath\}_{\jmath=1}^{\mathfrak{T}}$ as the set of \ac{spd} matrices in one demonstration, where $\spd$ defines the set of $m\times m$ \ac{spd} matrices. Afterwards, we can rewrite \eqref{eq:dmp_discrete1} and \eqref{eq:dmp_discrete2} as follows
\begin{align}
\begin{split}
\tau\dot{\bm{\sigma}} & =  \alpha_z(\beta_z \bm{vec}(\mathbb{B}_{\X_\jmath\mapsto\X_1} ({\LogS_{\X_\jmath}(\X_g)})) - \bm{\sigma}) \\
&\quad + \bm{\mathcal{F}}_{+}(x), \label{eq:spd:accel}
\end{split}\\
\tau\dot{\bm{\xi}} & =  \bm{\sigma}, \label{eq:spd:velocity}
\end{align}
where $\bm{\sigma} = \bm{vec}(\bm{\Sigma})$ is the Mandel representation of the symmetric matrix $\bm{\Sigma}$, where $\bm{\Sigma}$ is the time derivative of $\bm{\Xi}$ so that $\bm{\Sigma}\equiv\bm{\dot{\Xi}} = (\LogS_{\X_{\jmath-1}}(\X_\jmath))/\delta t$. The function ${\text{Log}_{\X_{\jmath-1}}(\X_\jmath)\!:\!\mathcal{M}\mapsto\mathcal{T}_{\X_{\jmath-1}}\mathcal{M}}$ maps a point $\X_\jmath$ in the manifold $\mathcal{M}$ to a point in the tangent space$\bm{\Delta} \in \mathcal{T}_{\X_{\jmath-1}}\mathcal{M}$. $vec(\cdot)$ is a function that transforms a symmetric matrix into a vector using Mandel's notation, \eg a vectorization of a $2\times 2$ symmetric matrix is
\begin{equation}
vec\Bigg(	\begin{pmatrix} 		a & b \\		b & d 	\end{pmatrix}\Bigg) = 
\begin{pmatrix} 		a \\		d \\ \sqrt{2}b 	\end{pmatrix}.
\end{equation}
$\bm{\xi}$ is the vectorization of $\bm{\Xi}$. $\X_g\in \spd$ represents the goal \ac{spd} matrix. $vec(\mathbb{B}_{\X_\jmath\mapsto\X_1} ({\LogS_{\X_\jmath}(\X_g)}))$ is the vectorization of the transported symmetric matrix ${\LogS_{\X_\jmath}(\X_g)}$ over the geodesic from $\X_\jmath$ to $\X_1$. Then we integrat \eqref{eq:spd:velocity} as
\begin{equation}
\bm{\hat{X}}(t+\delta t)=\ExpS_{\X(t)}\Bigg(\frac{\mathbb{B}_{\X_1\mapsto\X(t)} (mat(\bm{\sigma}(t)))}{\tau}\delta t\Bigg).
\end{equation}
where the function $mat(\cdot)$ is the inverse of $vec(\cdot)$ and denotes to the matricization using Mandel's notation. $\bm{\hat{X}}\in \spd$ represents the new \ac{spd}-matrices-based robot skills. The function $\ExpS_{\X_{\jmath-1}}(\bm{\Delta})\!:\!{\mathcal{T}_{\X_{\jmath-1}}\mathcal{M}\mapsto\mathcal{M}}$ maps a point $\bm{\Delta}\in \mathcal{T}_{\X_{\jmath-1}}\mathcal{M}$ to a point ${\X_\jmath\in\mathcal{M}}$, so that it lies on the geodesic starting from ${\X_{\jmath-1}\in \spd}$ in the direction of $\bm{\Delta}$. 
An exemplar \ac{spd} \ac{dmp} is shown in \Figref{fig:spd_dmp_example}.

Moreover, \cite{abudakka2020Geometry} rewrote \eqref{eq:goalSwitching} for smooth goal adaptation in case of sudden goal switching as follows
\begin{equation}
		\tau \bm{\dot{g}_+} = \alpha_g\LogS_{\bm{g}_{new}^+}(\bm{g}_+).
\end{equation}
so $\bm{g}$ now is updated continually.

\subsection{Periodic \texorpdfstring{\acs{dmp}}{}}
\label{subsec:periodic}

The periodic \ac{dmp} (sometimes called rhythmic \ac{dmp}) are used when the encoded motion follows a rhythmic pattern.

\subsubsection{Classical \texorpdfstring{\acs{dmp}}{}}
\label{subsec:periodic:classical}

\begin{figure}[t]
	\centering
	\includegraphics[width=\linewidth]{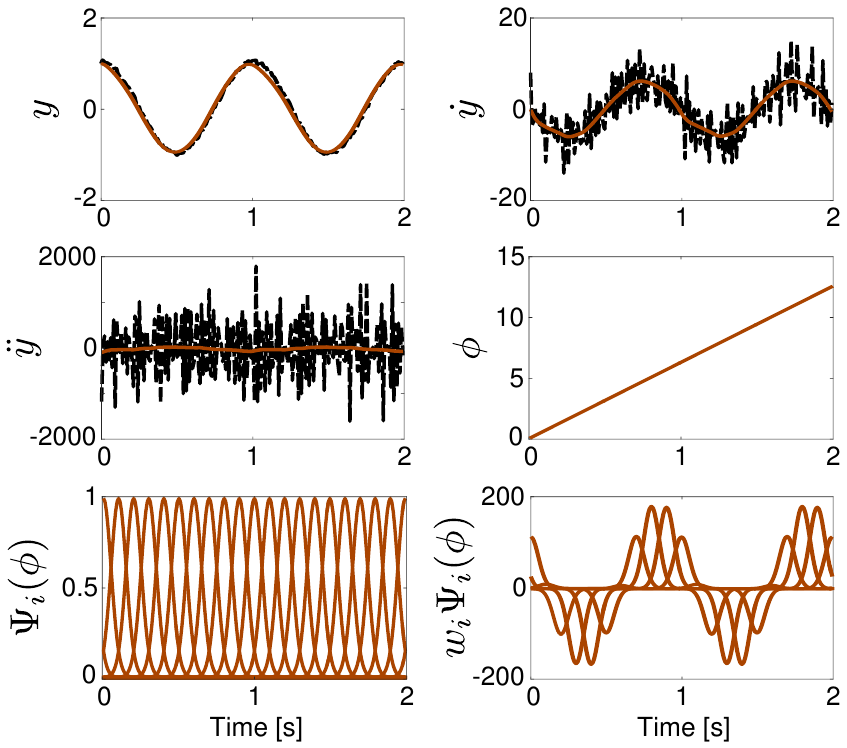}
	\caption{A classical \ac{dmp} is used to reproduce a rhythmic motion (brown solid line in the top left panel). The desired trajectory is obtained by adding Gaussian noise to $y_d = \cos{(2\pi t)}$ with $t \in [0, 2\,]s$ and computing the numerical derivatives with $\delta t = 0.01\,$s (black dashed lines). The forcing term is obtained as the weighted summation of $20$ Gaussian basis equally distributed in time (bottom left panel). The results of the parameters learning procedure are shown in the bottom right panel.
		Results are obtained with the open source implementation available at \url{www-clmc.usc.edu/software/git/gitweb.cgi?p=matlab/dmp.git}. }
	\label{fig:rythmic_dmp_example}
\end{figure} 

The classical periodic (or rhythmic) \acp{dmp} were first introduced by \cite{Ijspeert2002Learning}, where they redefined the second order differential equation system described in \eqref{eq:dmp_discrete1} and \eqref{eq:dmp_discrete2} as follows
\begin{align}
\dot z &= \Omega \left( {\alpha \left( {\beta  \left( - y\right) - z} \right) +f(\phi)} \right),
\label{eq:dmp_periodic1} \\
\dot y &= \Omega z,	\label{eq:dmp_periodic2} \\
\tau\dot{\phi} &= 1, \label{eq:dmp_periodic3}
\end{align}
where $\Omega$ is the frequency and $y$ is the desired periodic trajectory that we want to encode with a \ac{dmp}. The main difference between periodic \acp{dmp} and point-to-point \acp{dmp} is that the time constant related to trajectory duration is replaced by the frequency of trajectory execution (refer to~\citep{Ijspeert2013Dynamical,Ijspeert2002Learning} for details). In addition, the periodic \acp{dmp} must ensure that the initial phase ($\phi=0$) and the final one ($\phi=2\pi$) coincide in order to achieve smooth transition during the repetitions.

Similar to \eqref{eq:fx_discrete}, $f(\phi)$ is defined with $N$ Gaussian kernels according to the following equation
\begin{align}
f(\phi)&=\frac{{\sum_{i = 1}^N {\Psi_i(\phi) w_i r} }}{{\sum_{i = 1}^N {\Psi_i(\phi) } }}, \label{eq:fx_periodic}\\
\Psi_i(\phi)  &=  \exp\left(h\left( {\cos \left( {\phi  - c_i } \right) - 1} \right)\right),
\label{eq:psi_periodic}
\end{align}
where the weights are uniformly distributed along the phase space, and $r$ is used to modulate the amplitude of the periodic signal~\citep{Ijspeert2002Learning,Gams2009online} (if not used, it can be set to $r=1$~\citep{peternel2016adaptive}).

\input{summary_formulation_table}

Similarly to discrete \acp{dmp}, \ac{lwr}~\citep{Schaal1998} can be used to update the weight to learn a desired trajectory. In a standard periodic \ac{dmp} setting~\citep{Ijspeert2002Learning,Gams2009online}, the desired shape $f_d$ is approximated by solving
\begin{equation}
f_d(t_{\jmath}) = \frac{\ddot{y}_d(t_{\jmath})}{\Omega^2}  -  \alpha_z \left( {\beta_z  \left( - y_d(t_{\jmath})\right)  - \frac{\dot{y}_d(t_{\jmath})}{\Omega}  } \right),
\label{eq:goal}
\end{equation}
where $y_d$ is some demonstrated input trajectory that needs to be encoded. The weights $w_{i}$  can be updated using the recursive least-squares method~\citep{Schaal1998} with forgetting factor $\lambda$ based on the error between the desired trajectory shape and currently learned shape
\begin{align}
w_{i}(t_{\jmath+1}) & = w_{i}(t_{\jmath})+\Psi_{i}P_{i}(t_{\jmath+1})re_r(t_{\jmath})\label{eq:reg1},\\
e_r(t_{\jmath})&=f_{d}(t_{\jmath})-w_{i}(t_{\jmath})r\label{eq:reg3},\\
P_{i}(t_{\jmath+1})&=\frac{1}{\lambda}\left(P_{i}(t_{\jmath})-\frac{P_{i}(t_{\jmath})^{2}r^{2}}{\frac{\lambda}{\Psi_{i}}+P_{i}(t_{\jmath})r^{2}}\right).\label{eq:reg2}
\end{align}
The initial value of the parameters is $w_{i}(0)=0$ and $P_{i}(0)=1$. The forgetting factor determines the rate of weight changes. Refer to~\citep{Schaal1998} for details on parameter setting. An exemplar rhythmic \ac{dmp} is shown in \Figref{fig:rythmic_dmp_example}.

The classical periodic \ac{dmp} described by \eqref{eq:dmp_periodic1}--\eqref{eq:dmp_periodic3} does not encode the transit motion needed to start the periodic one. Transients are important in several applications like humanoid robot walking where usually the first step made from a rest position is a transient needed to start the periodic motion.
To overcome this limitation, \citep{ernesti2012encoding} modify the classical formulation  of periodic \acp{dmp} to explicitly consider transients as motion trajectory that converge towards the limit cycle (\ie periodic) one.

\subsection{Formulation summary}
\label{subsec:formSummary}

A summary for the existence \ac{dmp} formulations mentioned in the earlier sections is shown in  \tabref{tab:comparison}. The table shows the variations of the formulation in its standard shape based on the space that they are applied to. However, the modifications of this standard shape (\eg adding a coupling term) is discussed in the next section as an extension of the \ac{dmp} formulations.

%% file: notation.tex
\resizebox{\linewidth}{!}{%
		{\renewcommand\arraystretch{1} 
			
			\centering
			\begin{tabular}{m{0.15\linewidth}m{0.01\linewidth}m{0.34\linewidth}||m{0.155\linewidth}m{0.01\linewidth}m{0.335\linewidth}}
				\noalign{\hrule height 1.5pt}
				\rowcolor{Light0}
				$N$ & $\triangleq$ & \# of nonlinear basis functions
				&
				$i$ & $\triangleq$ & index $: i=1,2,\ldots,N$ \\
				\rowcolor{Light0}
				$J$ & $\triangleq$ & \# of joints or \acp{dof}
				&
				$j$ & $\triangleq$ & index $: j=1,2,\ldots,J$ \\
				\rowcolor{Light0}
				$L$ & $\triangleq$ & \# of demonstrations or \acp{dmp}
				&
				$l$ & $\triangleq$ & index $: l=1,2,\ldots,L$ \\
				\rowcolor{Light0}
				$V$ & $\triangleq$ & \# of via-points or via-goals
				&
				$v$ & $\triangleq$ & index $: v=1,2,\ldots,V$ \\
				\rowcolor{Light0}
				$\mathfrak{T}$ & $\triangleq$ & \# of datapoints
				&
				$\jmath$ & $\triangleq$ & index $: \jmath=1,2,\ldots, \mathfrak{T}$ \\
				\rowcolor{Light0}
				$m$ & $\triangleq$ & dimensions of $\spd$
				&
				$n$ & $\triangleq$ &  dimensions of $\mathbb{R}^n$ \\
				\rowcolor{Light0}
				$\{\cdot\}_d$ & $\triangleq$ & subscript for desired value
				&
				$\{\cdot\}_q$ or $\{\cdot\}^q$ & $\triangleq$ & quaternion related variable \\
				\rowcolor{Light0}
				$\{\cdot\}_R$ or $\{\cdot\}^R$ & $\triangleq$ & rotation matrix related variable
				&
				$\{\cdot\}_{++}$, $\{\cdot\}_+$ or $\{\cdot\}^+$ & $\triangleq$ & \ac{spd} related variable \\
				\rowcolor{Light0}
				$\{\cdot\}_{g}$ & $\triangleq$ & subscript for goal value
				&
				$\alpha_z$, $\beta_z$, $\alpha_{x}$, $\alpha_{s}$, $\alpha_{g}$, $\alpha_{yx}$, $\alpha_{qg}$ & $\triangleq$ & positive gains \\
				\rowcolor{Light0}
				$\tau$ & $\triangleq$ & time modulation parameter
				&
				$c_i,h_i$ & $\triangleq$ & centers and widths of Gaussians \\
				\rowcolor{Light0}
				$T$ & $\triangleq$ & time duration
				&
				$t$ & $\triangleq$ & continuous time \\
				\rowcolor{Light0}
				$\lambda$ 
				& $\triangleq$ & forgetting factor 
				&
				$r$ & $\triangleq$ & amplitude modulation parameter \\
				\rowcolor{Light1}
				$x$ & $\triangleq$ & phase variable
				&
				$y,\dy$ & $\triangleq$ & trajectory data and its 1st derivative \\
				\rowcolor{Light1}
				$s$ & $\triangleq$ & sigmoidal decay phase
				&
				$z,\dz$ & $\triangleq$ & scaled velocity and acceleration \\
				\rowcolor{Light1}
				$p$ & $\triangleq$ & piece-wise  linear  phase
				&
				$g$, $\bm{g}_q$, $\bm{g}_+$ & $\triangleq$ & attractor point (\textit{goal}) in different spaces \\
				\rowcolor{Light1}
				$\bm{\omega}$ & $\triangleq$ & angular velocity
				&
				$\hat{g}$, $\hat{\bm{g}}_q$ and $\tilde{g}$, $\tilde{\bm{g}}_q$ & $\triangleq$ & moving target and delayed goal function in different spaces \\
				\rowcolor{Light1}
				$\mathbf{\mathcal{Q}}_t,\ \mathbf{\mathcal{\dot{Q}}}_t$ & $\triangleq$ & joint position, its 1st time-derivative
				&
				$g_v$ & $\triangleq$ & intermediate attractor (\textit{via-goal})  \\
				\rowcolor{Light1}
				$\q,\dq$ & $\triangleq$ & unit quaternion, its 1st time-derivative
				&
				$\R,\dR$ & $\triangleq$ & rotation matrix, its 1st time-derivative \\
				\rowcolor{Light1}
				$f$, $\bm{f}_q$, $\bm{f}_R$, $\bm{f}_q$, $\bm{\mathcal{F}}_{+}$ & $\triangleq$ & forcing term for different spaces
				&
				$w_i$ & $\triangleq$ & adjustable weights \\
				\rowcolor{Light1}
				$\Psi_i$ & $\triangleq$ & basis functions
				&
				$\theta$ and $\bm{\vartheta}$ & $\triangleq$ &  an angle and learnable parameters \\
				\rowcolor{Light2}
				$\spd$ & $\triangleq$ & $m\times m$ \ac{spd} manifold
				&
				$\sym$ & $\triangleq$ & $m\times m$ symmetric matrices space\\
				\rowcolor{Light2}
				$\mathcal{M}$ & $\triangleq$ & a Riemannian manifold
				&
				$\X$ & $\triangleq$ & an arbitrary \ac{spd} matrix \\
				\rowcolor{Light2}
				${\mathcal{T}_{\bm{\Lambda}} \mathcal{M}}$ & $\triangleq$ & a tangent space of $\mathcal{M}$ at an arbitrary point $\bm{\Lambda}$
				&
				${\bm{M}}$ & $\triangleq$ & the mean of $\{\X_{t}\}_{t=1}^\mathfrak{T}$ \\
				\rowcolor{Light2}
				$\bm{\varrho} = \text{Log}_{\bm{\Lambda}}(\bm{\Upsilon})$ & $\triangleq$ & $\mathcal{M}\mapsto\mathcal{T}_{\bm{\Lambda}}\mathcal{M}$, maps an arbitrary point $\bm{\Upsilon}\in\mathcal{M}$ into $\bm{\varrho} \in \mathcal{T}_{\bm{\Lambda}} \mathcal{M}$
				&
				$\bm{\Upsilon}=\text{Exp}_{\bm{\Lambda}}(\bm{\varrho})$ & $\triangleq$ & $\mathcal{T}_{\bm{\Lambda}}\mathcal{M}\mapsto\mathcal{M}$, maps $\bm{\varrho} \in \mathcal{T}_{\bm{\Lambda}} \mathcal{M}$ into $\bm{\Upsilon}\in\mathcal{M}$ \\
				\rowcolor{Light2}
				$vec(\cdot)$ & $\triangleq$ & a function transforms $\sym$ into $\mathbb{R}^n$ using Mandel's notation.
				&
				$mat(\cdot)$ & $\triangleq$ & a function transforms $\mathbb{R}^n$ into $\sym$ using Mandel's notation.\\
				\rowcolor{Light3}
				$k$, $K$, $\Kp$, $\Ko$ & $\triangleq$ & different forms of stiffness gains
				&
				$D, \Dv$, $\Dw$ & $\triangleq$ & different forms of damping gains \\
				\rowcolor{Light3}
				$\bm{\mathfrak{M}}$ and $\bm{\mathcal{I}}$ & $\triangleq$ & mass and inertia matrices
				&
				$F$, $\fe$ and $\taue$ & $\triangleq$ & forces and external forces and torques \\
				\noalign{\hrule height 1.5pt} 
				\end{tabular}
				}}
				
				
        \resizebox{\linewidth}{!}{%
		{\renewcommand\arraystretch{1} 
				\begin{tabular}{m{0.06\linewidth}m{0.44\linewidth}||m{0.06\linewidth}m{0.44\linewidth}}
				\noalign{\hrule height 1.5pt}
				\rowcolor{Light0}
				\acs{dmp}	& \acl{dmp} & \acs{il}		&\acl{il}\\
			\rowcolor{Light0}
			\acs{rl}		& \acl{rl} &
			\acs{spd}	& \acl{spd}\\
			\rowcolor{Light0}
			\acs{dof}	& \acl{dof}&
			\acs{rbf}	& \acl{rbf}\\
			\rowcolor{Light0}
			\acs{lwr}	& \acl{lwr}&
			\acs{gmm}	& \acl{gmm}\\
			\rowcolor{Light0}
			\acs{gmr}	& \acl{gmr}&
			\acs{gp}		& \acl{gp}\\
			\rowcolor{Light0}
			\acs{nn}		& \acl{nn}&
			\acs{vmp}	& \acl{vmp}\\
			\rowcolor{Light0}
			\acs{promp}	& \acl{promp}&
			\acs{lfd}	& \acl{lfd}\\
			\rowcolor{Light0}
			\acs{gpr}	& \acl{gpr}&
			\acs{momp}	& \acl{momp}\\
			\rowcolor{Light0}
			\acs{emg}	& \acl{emg}&
			\acs{ilc}	& \acl{ilc}\\
			\rowcolor{Light0}
			\acs{vic}	& \acl{vic}&
			\acs{vilc}	& \acl{vilc}\\
			\rowcolor{Light0}
			\acs{pi2}	& \acl{pi2}&
			\acs{cmaes}	& \acl{cmaes}\\
			\rowcolor{Light0}
			\acs{ccdmp}	& \acl{ccdmp}&
			\acs{rbfnn}	& \acl{rbf}-\acl{nn}\\
			\rowcolor{Light0}
			\acs{power}	& \acl{power}&
			\acs{hrl}	& \acl{hrl}\\
			\rowcolor{Light0}
			\acs{aedmp}	& \acl{aedmp}&
			\acs{cnn}	& \acl{cnn}\\
			\rowcolor{Light0}
			\acs{gpdmp}	& \acl{gpdmp}&
			\acs{uav}	& \acl{uav}\\
			\noalign{\hrule height 1.5pt}
			\end{tabular}
	}}

%% file: summary_formulation_table.tex
\begin{table*}[t]
	\caption{Summary of \ac{dmp} basic formulations.} 
	\resizebox{\textwidth}{!}{%
		{\renewcommand\arraystretch{1.2} 
			
			\centering
			\begin{tabular}{|m{0.1\textwidth}|m{0.05\textwidth}|m{0.45\textwidth}|m{0.1\textwidth}|m{0.27\textwidth}|}
				\noalign{\hrule height 1.5pt}
				\multicolumn{1}{|c}{\textbf{Type of movement}}& \multicolumn{1}{|c}{\textbf{Space}} &\multicolumn{1}{|c}{\textbf{System of equations}} & \multicolumn{1}{|c}{\textbf{Reference}} & \multicolumn{1}{|c|}{\textbf{Short description}} \\
				\noalign{\hrule height 1pt}
				\multirow{20}{*}{\textbf{Discrete}}&
				\begin{tabular}[c]{@{}l@{}}$\mathbb{R}$\end{tabular} &
				\begin{tabular}[c]{@{}l@{}}$\tau \dz = \alpha_z(\beta_z(g-y)-z) + f(x)$\\$\tau \dy = z$\\$\tau \dx = \alpha_x x$\end{tabular} & Eqs. \eqref{eq:dmp_discrete1}--\eqref{eq:dmp_discrete3}, \citep{Ijspeert2002Movement} &  A single \ac{dof}, discrete motion trajectory is encoded into a linear, second-order dynamical system with an additive, non-linear forcing term. Convergence to the desired goal $g$ is ensured by a vanishing phase variable $x$.\\ \cline{2-5}
				&
				$\bm{\mathcal{S}}^3$
				 &\begin{tabular}[c]{@{}l@{}}$\tau \bm{\Dot{\eta}} = \alpha_z(\beta_z2 \, \LogQ(\bm{g}_q*\overline{\q})-\bm{\eta}) + \bm{f}_q(x)$\\$\tau \dq = \frac{1}{2}\bm{\eta}*\q$\end{tabular}
				 &
				 Eqs. \eqref{eq:quat:dw}--\eqref{eq:quat:dq}, \citep{AbuDakka2015Adaptation} & A quaternion-based orientation trajectory ($3$ \acp{dof}) is encoded into a second-order dynamical system with an additive, non-linear forcing term. The error definition complies with the geometry of the unit quaternions space.  \\ \cline{2-5}
				& 
				$\bm{\mathcal{SO}}(3)$
				&  
				\begin{tabular}[c]{@{}l@{}}$\tau \bm{\Dot{\eta}} = \alpha_z(\beta_z \, \LogR(\R_g*\R\trsp)-\bm{\eta}) + \bm{f}_R(x)$\\$\tau \dR = [\bm{\eta}]_\times \R$\end{tabular} & Eqs. \eqref{eq:rot:dw}--\eqref{eq:rot:dq}, \citep{Ude2014Orientation}  & A rotation matrix-based orientation trajectory ($3$ \acp{dof}) is encoded into a second-order dynamical system with an additive, non-linear forcing term. The error definition complies with the geometry of the rotation matrices space.\\ \cline{2-5}
				& 
				$\spd$
				&
				\begin{tabular}[c]{@{}l@{}}$\tau\dot{\bm{\sigma}}  =  \alpha_z(\beta_z \bm{vec}(\mathbb{B}_{\X_l\mapsto\X_1} ({\LogS_{\X_l}(\X_g)})) - \bm{\sigma}) + \bm{\mathcal{F}}(x)$\\$\tau\dot{\bm{\xi}} = \bm{\sigma}$\end{tabular}  & Eqs. \eqref{eq:spd:accel}--\eqref{eq:spd:velocity}, \citep{abudakka2020Geometry} & An \ac{spd} matrices trajectory, $m(m+1)/2$ \acp{dof}, is encoded into a second-order dynamical system with an additive, non-linear forcing term. The error definition complies with the geometry of the \ac{spd} matrices space.\\ \cline{2-5}
				\hline \hline
				\multirow{1}{*}{\textbf{Periodic}}  & \begin{tabular}[c]{@{}l@{}}$\mathbb{R}$\end{tabular} & \begin{tabular}[c]{@{}l@{}}$\dot z = \Omega \left( {\alpha \left( {\beta  \left( - y\right) - z} \right) +f(\phi)} \right)$\\$\dot y = \Omega z$\\$\tau\dot{\phi} = 1$\end{tabular} & Eqs. \eqref{eq:dmp_periodic1}--\eqref{eq:dmp_periodic3}, \citep{Ijspeert2002Learning}  & A single \ac{dof}, periodic motion trajectory is encoded into a linear, second-order dynamical system with an additive, non-linear forcing term. The resulting system generates a stable limit cycle.\\
				\noalign{\hrule height 1.5pt}
			\end{tabular}
	}}
	\label{tab:comparison}
\end{table*}

%% file: DMP_Extentions.tex
\section{\texorpdfstring{\acp{dmp}}{} extensions}
\label{sec:advanceMethods}

\subsection{Generalization}
\label{subsec:splines}
A desirable property of motion primitives is the ability to generalize to unforeseen situations. In this section, we present approaches that allow to adapt \ac{dmp} motion trajectories to novel executive contexts.

\subsubsection{Start, goal, and scaling}\label{subsubsec:start_goal}
Classical \acp{dmp} are time invariant, meaning that time scaling $\varsigma \tau$ with $\varsigma>0$ generate topologically equivalent trajectories~\citep{Ijspeert2013Dynamical}. Using a simple modification of the transformation system, namely substituting~\eqref{eq:dmp_discrete1} with
\begin{equation}
\tau \dz = \alpha_z(\beta_z(g-y)-z) + \textcolor{darkgreen}{(g-y_0)}f(x), \label{eq:dmp_scale_invariance1}
\end{equation}
\cite{Ijspeert2013Dynamical} show that \ac{dmp} are also scale invariant, meaning that the scaling of the movement amplitude $\varsigma(g-y_0)$ with $\varsigma>0$ generates topologically equivalent trajectories. The purpose of the green color used in~\eqref{eq:dmp_scale_invariance1} is to highlight differences \wrt\eqref{eq:dmp_discrete1}. 
Apart from generating scaled---in time and space---versions of the demonstrated motion trajectory, classical \acp{dmp} also generalize to different initial/target states. However, the classical formulation---and its extension in~\eqref{eq:dmp_scale_invariance1}---may exhibit dangerous behaviors like over-amplification of the trajectory when reaching a different target and high accelerations when switching to a different target on-line~\citep{Pastor2009,Ijspeert2013Dynamical}. 
To alleviate the second issue, \cite{Ijspeert2013Dynamical} replaced hard goal switches with the smooth switching law as in \eqref{eq:goalSwitching}.
However, the over-amplification issue still remains. Moreover, a \ac{dmp} that uses~\eqref{eq:dmp_scale_invariance1} fails to learn motions with the same initial and target states (\ie $g=y_0, z_0 = 0 \rightarrow y(t) = y_0 = g ~\forall t$).  

In order to remedy those issues, \cite{Pastor2009} proposed to modify the transformation system as
\begin{equation}
    \tau \dz = \alpha_z(\beta_z(g-y \textcolor{darkgreen}{- (g -y_0)x + f(x)} )-z), \label{eq:dmp_pastor1}
\end{equation}
where the green color is used to highlight differences between~\eqref{eq:dmp_pastor1} and~\eqref{eq:dmp_discrete1}. The most important change in this formulation is the term $(g-y_0)x$ that has several benefits. It prevents high accelerations at the beginning of the motion ($g - y - (g - y_0)x = 0$ for $t=0$) or when the goal is close to the initial state. It allows to reproduce motions with the same initial and target states and it prevents over-amplifications and trajectory mirroring effects\footnote{As discussed by~\citep{Pastor2009}, a transformation system that uses~\eqref{eq:dmp_scale_invariance1} generates a mirrored trajectory while reaching a new goal $g_{new}$ every time the signs of $(g_{new}-y_0)$ and $(g-y_0)$ differ.} when changing the goal. \cite{hoffmann2009biologically} derived a multidimensional representation of \eqref{eq:dmp_pastor1} from the behavior of the spinal force fields in frogs.

The goal can also change over time and, in this case, the tracking performance of the \ac{dmp} mostly depends on the gains $\alpha_z$ and $\beta_z$. As proposed by \citep{koutras2020dynamic}, the tracking performance can be improved by adapting the temporal scaling $\tau$.

\cite{dragan2015movement} showed that \acp{dmp} solve a trajectory optimization problem in order to minimize a particular Hilbert norm between the demonstration and the new trajectory subject to start and goal constraints. In this light, \ac{dmp} adaptation capabilities to different start and goals can be improved by choosing (or learning) a proper Hilbert norm that reduces the deformation in the retrieved trajectory.

\subsubsection{Via-points}\label{subsubsec:via-points}
A via-point can be defined as a point in the state space where the trajectory has to pass. Failing to pass a via-point may cause the robot to fail the task execution. Therefore, having a motion primitive representation with the capability of modulating the via-points is of importance in robotic scenarios. It is not surprising that researchers have extended the \ac{dmp} formulation to consider intermediate via-points in the trajectory generation process. 

\cite{ning2011accurate, ning2012novel} extend the classical \ac{dmp} to satisfy position and velocity constraints at the beginning and at the end of a sample trajectory. Their approach to traverse via-points consists of creating a sample trajectory by combining locally-linear trajectories connecting the via-points. This sample trajectory is used to fit a \ac{dmp} that is constrained to pass the via-points.

\cite{weitschat2018safe} considered each via-point as an intermediate goal (\textit{via-goal}) $g_v$ for $v=1,\ldots,V$ to reach. The last via-goal $g_V$ corresponded to the target state of the \ac{dmp}. In their formulation, they defined a variable goal as
\begin{equation}
    g_{via}(x) = \sum_{v=1}^V\Psi_v(x)g_v,
    \label{eq:variable_via_goal}
\end{equation}
where $\Psi_v(x)$ are the Gaussian basis function centered at the time corresponding to the $v-$th via-goal. The effectiveness of the approach is demonstrated in a task were the robot has to reach a different target while preventing possible self-collisions of the end-effector with the robot body. To this end, authors place the via-goals along the trajectory used to learn the \ac{dmp}, forcing the generated trajectory to stay close to the demonstration while reaching the new target.

The problem of generalizing to via-point close (interpolation) and far (extrapolation) from the demonstration is faced by \citep{Zhou2019Learning}. Their approach, namely \textit{\acp{vmp}}, combines the benefits of \ac{dmp} and \acp{promp} \citep{Paraschos2013}. Authors assumed that the motion trajectory is generated as
\begin{equation}
    y_{vmp}(x) = e(x) + f_{vmp}(x), 
    \label{eq:y_vmp}
\end{equation}
where $x$ is the phase variable defined as in~\eqref{eq:dmp_discrete3} and the elementary trajectory $e(x)$ can be defined as the linear attractor $e(x) = (y_0 - g)x$. The shape modulation term $f_{vmp}(x)$ is defined as
\begin{equation}
 f_{vmp}(x) = \sum_{i=1}^Nw_i\Psi_i(x) + \epsilon_f
    \label{eq:f_vmp}
\end{equation}
where the Gaussian kernels $\Psi_i(x)$ are defined as in~\eqref{eq:psi_discrete}, $w_i$ are learnable weights, and $\epsilon_f$ is the Gaussian noise. As detailed in \citep{Paraschos2013}, learning the shape modulation term $f_{vmp}(x)$ means \acp{lfd}  the  prior  probability distribution  of the weights $w_i$. Having separated the generated trajectory into two parts like in~\eqref{eq:y_vmp} allows to adopt different strategies to pass a via-point $y_{v}$ at $x_{v}$. \cite{Zhou2019Learning} proposed to modify the shape modulation term for interpolation cases--when the via-point is ``close'' to the demonstrations. In extrapolation cases, instead, the elementary trajectory $e(x)$ is rewritten as the polygonal line connecting $y_0$, $y_{v}$, and $g$. This approach easily generalizes to the case of multiple via-points. \acp{vmp} are experimentally compared with \acp{promp}, showing better performance especially in extrapolation cases.  

\begin{figure*}[t]
	\centering
	\includegraphics[width=\linewidth]{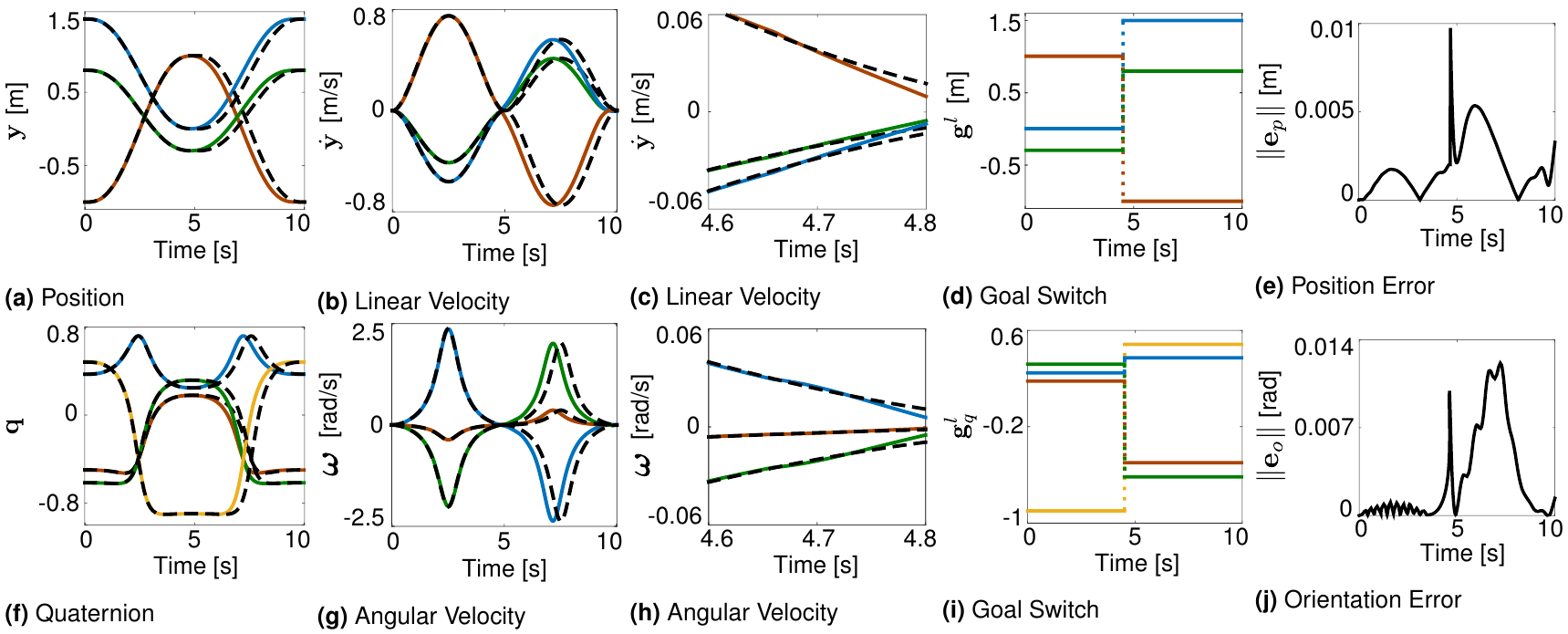}
	\caption{Results obtained by applying the zero velocity switch approach to join two \acp{dmp} trained on synthetic data. The training trajectory for the position and the orientation are shown as a black dashed lines in (a)-(b) and (f)-(g) respectively. Results are obtained with the open source implementation available at \url{https://gitlab.com/dmp-codes-collection}.}
	\label{fig:Joining_zero_velocity}
\end{figure*}

\subsubsection{Task parameters}\label{subsubsec:task_paramters}
Reaching a different goal, or passing through via-points, may not be enough to successfully execute a task in a different context. Approaches presented in this section adapt the \ac{dmp} motion to new situations by adjusting the weights $w_i$ of the forcing term~\eqref{eq:fx_discrete}, that modifies the entire \ac{dmp} trajectory.

\cite{weitschat2013dynamic} considered that $L$ demonstrations are given, each encoded in a different \ac{dmp}. In order to generalize, for instance, to a new goal $g_{new}$, they proposed to interpolate the weights of nearby \acp{dmp}, \ie \acp{dmp} that reached points around $g_{new}$. In formulas
\begin{equation}
    \bm{w}_{new} = \frac{\sum\limits_{\forall o:d_o<d_{max}} \bm{w}_{o}d_o^{-1}}{\sum\limits_{\forall o:d_o<d_{max}} d_o^{-1}},
    \label{eq:weight_intepolation}
\end{equation}
where $o$ represents the indices of the nearby \acp{dmp} for which it holds that $d_o < d_{max}$. $d_o$ is the distance (or, more generally, a cost) between $g_{new}$ and $g_o$, $d_{max}$ is the maximum distance to consider 2 \acp{dmp} close. $\bm{w}_{new} = [w_{1,new},\cdots,w_{N,new}]\trsp$ and $\bm{w}_{o} = [w_{1,o},\cdots,w_{N,o}]\trsp$ are the new weights and the weight of nearby \acp{dmp}, respectively.

The approach by \cite{forte2011realtime, forte2012line} also assumes that $L$ demonstrations are given and that each demonstration is encoded in a different \ac{dmp}. Further, the authors exploited \ac{gp}~\citep{Rasmussen2006} to learn a mapping between the query points $q_l$ for $l=1,\ldots,L$ (\eg the goal of each \ac{dmp}) and the \ac{dmp} parameters $[\bm{w}_l, g_l, \tau_l]$. Given the new query point $q_{new}$, \ac{gpr}~\citep{Rasmussen2006} is used to retrieve the new set of parameters $[\bm{w}_{new}, g_{new}, \tau_{new}]$, that can be used to generate a \ac{dmp} motion. This approach builds on previous work \citep{gams2009generalization,ude2010task} where raw data from the $L$ demonstrations are stored in memory and \ac{lwr} is used to generate new \ac{dmp} weights. \cite{alizadeh2016learning} extend the approach in \citep{ude2010task} to retrieve the \ac{dmp} weights even when the task parameters are partially observable. Finally, \citep{Zhou2017} extend the approach in \citep{ude2010task} to consider task-specific costs while learning the mapping between query points and \ac{dmp} weights.  

Aforementioned approaches follow a $2$-steps procedure where first the shape parameters $\bm{w}$ are estimated given new task parameters and then execute the \ac{dmp}.  \cite{matsubara2010learning, Matsubara2011}  augmented the forcing term with a style parameter used to capture human variability across multiple demonstrations.   \cite{Stulp13_hum} proposed a $1$-step procedure where the \ac{dmp} forcing term~\eqref{eq:fx_discrete} is reformulated to explicitly depend on the task parameters. Their experiments shows that a $1$-step approach gives more freedom \wrt the used regression technique and increase the generalization performance. Along the same line, \cite{pervez2018learning} embedded task parameters directly in the forcing term. Authors proposed to use a mixture of Gaussians \citep{cohn1996active} to learn the mapping between the task parameters (\eg new goal, height of an obstacle, \etc) and the forcing term. 
Given a new query task parameter, regression over the mixture of Gaussians is used to retrieve the forcing term parameters and generate the \ac{dmp} motion. The approach is tested on a variety of tasks including sweeping and stricking and additionally compared with the approaches presented by~\citep{ude2010task,forte2012line, Stulp13_hum} showing better performance especially in extrapolation.

A \ac{momp} is proposed in \citep{Muelling10, mulling2013learning} and used to generalize table tennis skills like hitting and batting a ball. \ac{momp} uses an augmented state that contains robot position and velocity as well as the meta-parameters of the table tennis task like the expected hitting position and velocity. The adapted motion is generated by the weighted summation of $L$ \acp{dmp} and the responsibility of each \ac{dmp}, representing the probability that a particular \ac{dmp} is the correct one for the sensed augmented state, is also learned from data.  

In high \ac{dof} systems, like humanoid robots, it is non trivial to find a relationship between the task and the \ac{dmp} parameters. This is especially true when the \acp{dmp} are used to encode joint space trajectories. \cite{bitzer2009latent} showed that such a relationship is easier to find in a latent (lower dimensional) space obtained from training data. Therefore, they used dimensionality reduction techniques to find the latent space where to fit a \ac{dmp} and show that interpolation of \ac{dmp} weights in the latent space results in better generalization performance.

\subsection{Joining multiple \texorpdfstring{\acp{dmp}}{}}
\label{subsec:joiningDMPs}
An important and desired feature of any motion primitive representation is the possibility to combine basic movements to obtain more complex behaviors~\citep{schaal1999imitation}. We review here three prominent approaches developed to smoothly join a sequence of \acp{dmp}. In this tutorial, we name the approach by \cite{Pastor2009} as \textit{velocity threshold}, that in \citep{kober2010movement} as \textit{target crossing}, and that in \citep{kulvicius2011modified, kulvicius2012joining} as \textit{basis functions overlay}. Some of the presented approaches modify the \ac{dmp} formulations in \secref{subsec:discrete:classical}~and~~\ref{subsec:discrete:rotation}. The main differences are highlighted with green text. The $3$ approaches have been implemented in Matlab for both position (Section~\ref{subsec:discrete:classical}) and orientation (Section~\ref{subsec:discrete:rotation}) \acp{dmp}. The source code is included in our public repository (see Table~\ref{tab:source_code}). Results on synthetic data are shown in \figsref{fig:Joining_zero_velocity}~to~\ref{fig:Joining_overlay}.

\subsubsection{Velocity threshold}\label{subsubsec:zero_vel_switch}
A properly designed \ac{dmp} reaches the desired target with zero velocity and acceleration, \ie once a \ac{dmp} is fully executed the robot comes to a full stop. This also implies that the velocity ``close'' to the target is continuously decreasing. Using this property, \cite{Pastor2009} propose to combine successive \acp{dmp} by simply terminating the current \ac{dmp} when the velocity is below a certain threshold and then starting the following primitive. When executing a single \ac{dmp}, it is common practice to initialize its velocity to zero---the robot is assumed to be still. In principle, this initialization can be used to sequence multiple \acp{dmp} \citep{xu2004multiple, lioutikov2016learning}, but it may generate discontinuities if the robot does not fully stop in between two consecutive primitives. To prevent this discontinuities, \citeauthor{Pastor2009} initialized the state of the current \ac{dmp} with that of the previous one. 

The velocity threshold approach is simple and effective since it directly applies to the \ac{dmp} formulations in Sections~\ref{subsec:discrete:classical}~and~\ref{subsec:discrete:rotation}. For instance, \cite{saveriano2019merging} showed how to join multiple quaternion \acp{dmp}\footnote{\citeauthor{saveriano2019merging} used the multi-dimensional \ac{dmp} formulation developed in \citep{hoffmann2009biologically} for both position and quaternion \acp{dmp}. In this review paper, we reformulate the merging approaches in \citep{saveriano2019merging} to comply with the formulations in Section~\ref{subsec:discrete:classical}~and~\ref{subsubsec:discrete:quaternion}.} (see Section~\ref{subsubsec:discrete:quaternion}) with the velocity threshold approach. 

Results in \Figref{fig:Joining_zero_velocity} are obtained when velocity threshold is applied to merge $2$ \acp{dmp} separately trained to fit minimum jerk trajectories (black dashed lines). \figsref{fig:Joining_zero_velocity}a--\ref{fig:Joining_zero_velocity}e show the position and \figsref{fig:Joining_zero_velocity}f--\ref{fig:Joining_zero_velocity}j the orientation (unit quaternion) parts of the motion. The merged trajectory is generated by following the first \ac{dmp} until the distance from the via-point is below $0.01\,$[m] and $0.01\,$[rad]. As shown in \figsref{fig:Joining_zero_velocity}d~and~\ref{fig:Joining_zero_velocity}i, the switch occurs after about $4.7\,$[s]. \figsref{fig:Joining_zero_velocity}e~and~\ref{fig:Joining_zero_velocity}j shows that the desired trajectory is accurately reproduced. More or less accurate trajectories can be obtained by tuning the distance from the via-point. However, the value of this distance the time duration of the generated trajectory---a bigger (smaller) distance results in a shorter (longer) trajectory. For instance, in the considered case, the total motion ends after $9.5\,$[s] while the demonstration lasts for $10\,$[s]. Depending on the application, the time difference may cause failures, therefore, it has to be taken into account. Finally, the velocity threshold approach may generate discontinuities if the target of the current \ac{dmp} is far from the demonstrated initial point of the following primitive.

\subsubsection{Target crossing}\label{subsubsec:target_crossing}
\begin{figure}[t]
	\centering
	\includegraphics[width=\linewidth]{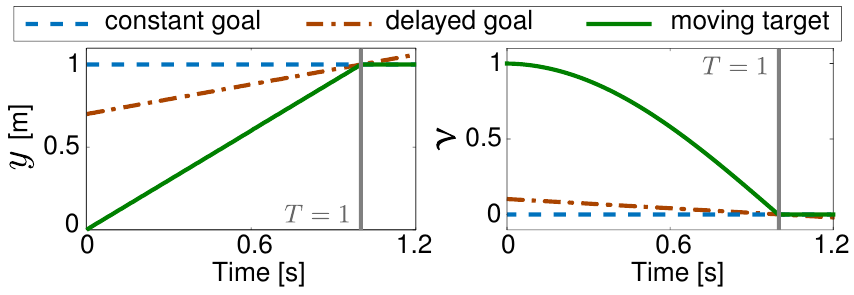}
	\caption{The constant goal, moving target, and delayed goal obtained obtained with $y(0) = 0\,$[m], $g = 1\,$[m], $\dot{\hat{y}} = 0.3\,$[m/s] (left), and $\q(0) = 1 + [0,0,0]\trsp$, $\bm{g}_q = 0 + [1,0,0]\trsp$, $\hat{\bm{\omega}} = [0.2,0.2,0.2]\trsp$[rad/s] (right). The sampling time is $\delta t = 0.01\,$[s]. Only the scalar part  $\upnu$ of the quaternion is shown for a better visualization.}
	\label{fig:goal_functions}
\end{figure}

There exist movements like hitting or batting that are correctly executed only if the target is reached with a non-zero velocity. To this end, \cite{kober2010movement} extend the classical \ac{dmp} formulation in \secref{subsec:discrete:classical} to let the \ac{dmp} to track a target moving at a given velocity. In their approach, the \ac{dmp} passes the target with a given velocity exactly after $T$ seconds. To achieve this, the acceleration in~\eqref{eq:dmp_discrete1} is re-written as
\begin{align}
\tau \dz = \textcolor{darkgreen}{(1-x)}\alpha_z\left(\beta_z(\textcolor{darkgreen}{\hat{g}} - y) \textcolor{darkgreen}{+ \tau(\dot{\hat{y}} - \dot{y})} \right) + f(x),
\label{eq:dmp_discrete_crossing1}
\end{align}
where $\dot{\hat{y}}_m$ is the desired velocity of the moving target $\hat{g}$, which is defined as
\begin{align}
&\hat{g} = \hat{g}(0) - \dot{\hat{y}} \frac{\tau \text{ln}(x)}{\alpha_x}, 
\label{eq:dmp_discrete_moving_target}\\
&\hat{g}(0) = g - \frac{T \dot{\hat{y}}}{\tau}.
\label{eq:dmp_discrete_initial_target}
\end{align}
By inspecting~\eqref{eq:dmp_discrete_moving_target} and~\eqref{eq:dmp_discrete_initial_target}, and considering that the term $- \tau\ln(x) / \alpha_x$  represents the elapsed time if $x$ is the phase defined in~\eqref{eq:dmp_discrete3}, it is possible to show that the moving target~$\hat{g}$ is designed to reach the goal $g$ after $T$ seconds, \ie $\hat{g}(T) = g$ (\figref{fig:goal_functions}-\emph{left}). The initial position of the moving target $\hat{g}(0)$ is obtained by moving the goal position $g$ for $T$ seconds at constant velocity $-\dot{\hat{y}}$. High accelerations at the beginning of the movement are avoided by the pre-factor~$(1-x)$ which is set to zero at the beginning of the motion ($x(0) = 1$). The approach by \cite{nemec2012action} combines a moving target and a particular initialization of the subsequent \ac{dmp} to ensure continuity of the movement up to second-order derivatives.

\begin{figure*}[t]
	\centering
	\includegraphics[width=\linewidth]{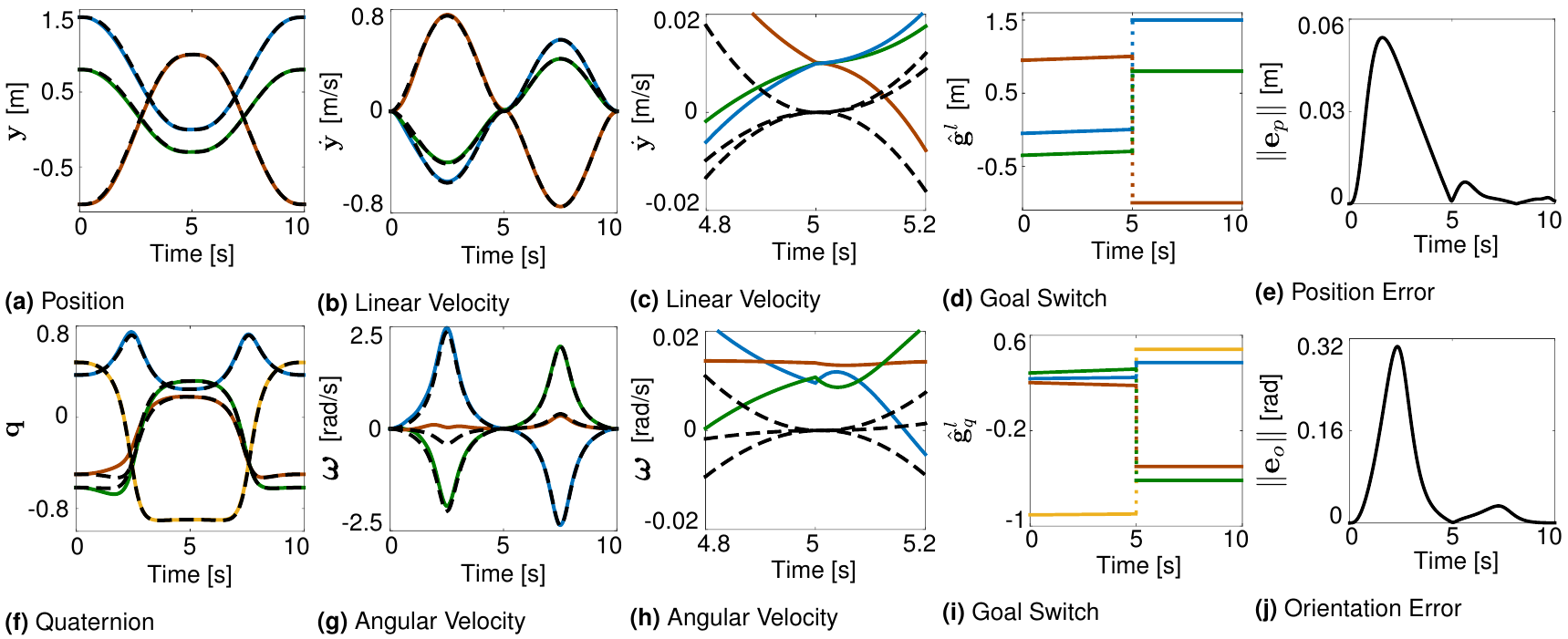}
	\caption{Results obtained by applying the target crossing approach to join two \acp{dmp} trained on synthetic data. The training trajectory for the position and the orientation are shown as a black dashed lines in (a)-(b) and (f)-(g) respectively. Results are obtained with the open source implementation available at \url{https://gitlab.com/dmp-codes-collection}.}
	\label{fig:Joining_crossing}
\end{figure*}

\cite{saveriano2019merging} extended this idea to quaternion \ac{dmp}. The angular acceleration in \eqref{eq:quat:dw} is modified as
\begin{align}
\begin{split}
\tau \bm{\Dot{\eta}} = &\textcolor{darkgreen}{(1-x})\alpha_z(\beta_z2 \, \LogQ(\textcolor{darkgreen}{\hat{\bm{g}}_{q}}*\overline{\q}) \textcolor{darkgreen}{+\tau(\hat{\bm{\omega}} - \bm{\omega})} \\
& + \bm{f}_q(x), \label{eq:quat:dw_crossing}
\end{split}
\end{align}
where $\hat{\bm{\omega}}$ is the angular velocity of the moving quaternion target $\hat{\bm{g}}_{q}$ and $2\,\LogQ(\hat{\bm{g}}_{q}*\overline{\q})$ measures the error between the current orientation $\q$ and $\hat{\bm{g}}_{q}$. The pre-factor~$(1-x)$ is used to avoid high angular accelerations at the beginning of the motion. The moving target for the quaternion \acp{dmp} is defined as
\begin{equation}
\begin{aligned}
 &\hat{\bm{g}}_{q} = \ExpQ\left(-\frac{\tau\ln(x)}{2\alpha_x} \hat{\bm{\omega}} \right) \ast \hat{\bm{g}}_{q}(0), \\
 &\hat{\bm{g}}_{q}(0) = \ExpQ\left(-\frac{T}{2} \hat{\bm{\omega}} \right) \ast \bm{g}_{q},
 \end{aligned}
\end{equation}
where~$\bm{g}_{q}$ is the goal quaternion, $T$ is the time duration of the \ac{dmp}, and the exponential map $\ExpQ(\cdot)$ is defined in~\eqref{eq:quat:exp}. As shown in \Figref{fig:goal_functions}-\emph{right}, the moving target $\hat{\bm{g}}_{q}$ reaches the goal orientation after $T$ seconds, \ie $\hat{\bm{g}}_{q}(T) = \bm{g}_{q}$. This can be easily verified by considering that the initial value of the moving target $\hat{\bm{g}}_{q}(0)$ is computed by moving the goal orientation $\bm{g}_{q}$ for $T$ seconds at the desired velocity $-\hat{\bm{\omega}}$.

The presented \textit{target crossing} approach allows to cross the target after $T$ seconds. Assuming to have two \acp{dmp} with time duration $T^1$ and $T^2$ respectively, one can join them by running the first \ac{dmp} for $T^1$ seconds and then switching to the second one. As for the velocity threshold approach, possible discontinuities at the switching point are prevented by initializing the state of \ac{dmp}\textsubscript{2} with the final state of \ac{dmp}\textsubscript{1}. This procedure can be repeated to join $L \geq 2$ consecutive \acp{dmp}.

Results in \Figref{fig:Joining_crossing} are obtained when the velocity threshold is applied for merging $2$  separately trained \acp{dmp} to fit the minimum jerk trajectories (black dashed lines). \figsref{fig:Joining_crossing}a--\ref{fig:Joining_crossing}e  show the position and \figsref{fig:Joining_crossing}f--\ref{fig:Joining_crossing}j the orientation (unit quaternion) parts of the motion. The merged trajectory is generated by following the first \ac{dmp} for $T^1=5\,$s and then switch to the second one. The required intermediate velocity is set to $0.01\,$m/s (rad/s for the orientation) in each direction. The generated trajectory reaches the goal in $10\,$s, \ie demonstration and execution times are the same. As required, the via-point is crossed at $T=5\,$s with the desired velocity (\figref{fig:Joining_crossing}c~and~\ref{fig:Joining_crossing}h). However, the non-zero crossing velocity introduce a deformation in the first part of the trajectory (\figref{fig:Joining_crossing}e~and~\ref{fig:Joining_crossing}j).

\begin{figure*}[t]
	\centering
	\includegraphics[width=\linewidth]{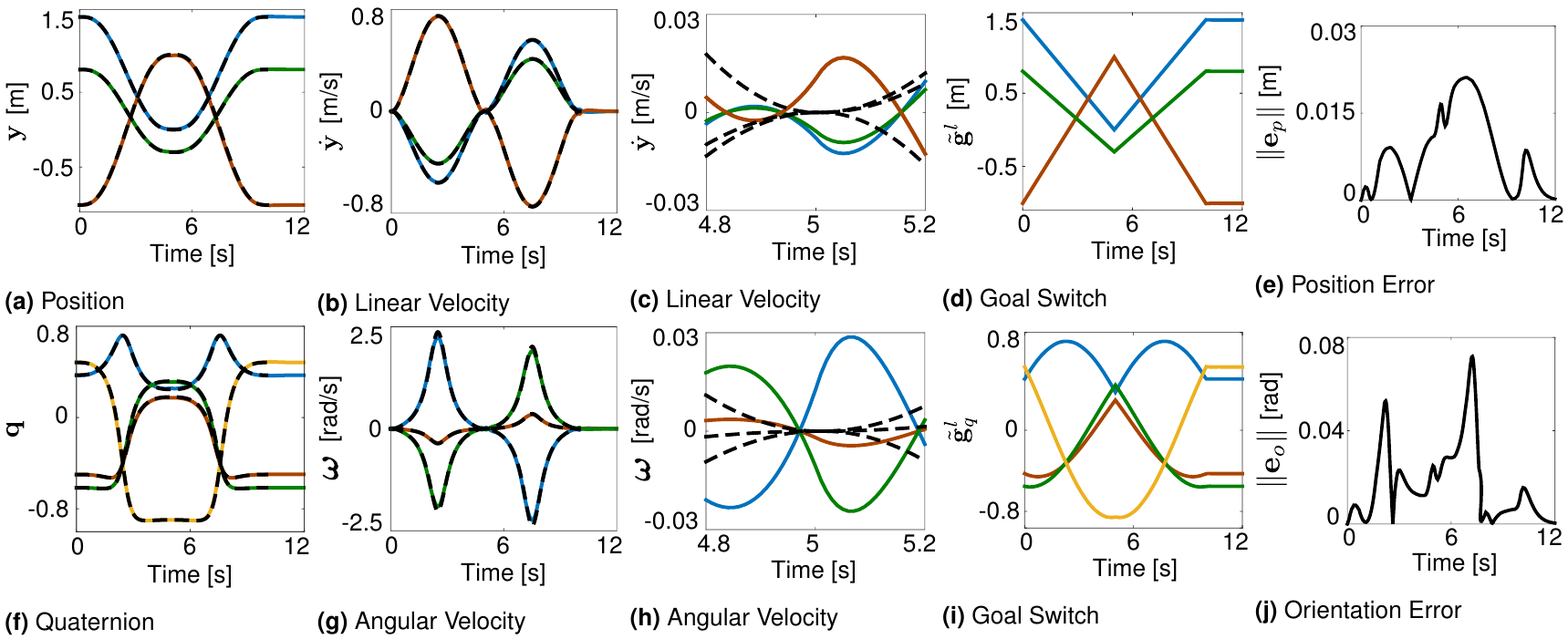}
	\caption{Results obtained by applying the basis functions overlay approach to join two \acp{dmp} trained on synthetic data. The training trajectory for the position and the orientation are shown as a black dashed lines in (a)-(b) and (f)-(g) respectively. Results are obtained with the open source implementation available at \url{https://gitlab.com/dmp-codes-collection}.}
	\label{fig:Joining_overlay}
\end{figure*}

\subsubsection{Basis functions overlay}\label{subsubsec:basis_functions_overlay}
The approach by \cite{kulvicius2011modified, kulvicius2012joining} combines multiple \acp{dmp} into a complex one, guaranteeing a smooth transition between the primitives by ensuring that the basis functions composing $f(x)$ in~\eqref{eq:fx_discrete} overlap at the switching instances. First of all, \citeauthor{kulvicius2012joining} adopted a sigmoidal phase variable in~\eqref{eq:sigmoidal_decay} instead of the exponentially decaying one~\eqref{eq:dmp_discrete3}. As discussed in \secref{par:phase_variables}, the sigmoidal phase is $\approx 1$ for the large part of the motion which makes it possible to use smaller forcing terms to reproduce the demonstrations. On the contrary, the exponential phase is close to zero already before $T\,$s (\figref{fig:phase_dmp}), which results in larger forcing terms.

The classical acceleration dynamics in~\eqref{eq:dmp_discrete1} is modified as
\begin{equation}
\tau \dz = \alpha_z(\beta_z(\textcolor{darkgreen}{\tilde{g}}-y)-z) + \textcolor{darkgreen}{f(s)},
\label{eq:dmp_discrete1_overlay}
\end{equation}
Similarly to target crossing, \citeauthor{kulvicius2012joining} used a moving target $\tilde{g}$ in the acceleration dynamics, but called it the \textit{delayed goal function}. The $\tilde{g}$ term in~\eqref{eq:dmp_discrete1_overlay} is obtained by integrating
\begin{equation} 
\tau \dot{\tilde{g}}= 
\begin{cases}
\frac{\delta t}{T}(g-y_0), & t \leq T\\
0, &\text{otherwise}
\end{cases}
\label{eq:linear_delayed_goal_function}
\end{equation}
with $\tilde{g}(0) = y_0$. The delayed goal function in \Figref{fig:goal_functions} moves linearly from $y_0$ to $g$ in $T$ seconds and then remains constant, \ie $\tilde{g}(t\geq T) = g$. 

The non-linear forcing term $f(s)$ is in green in~\eqref{eq:dmp_discrete1_overlay} because it slightly differs from the classical one in~\eqref{eq:fx_discrete}. $f(s)$ is defined as
\begin{equation}
\begin{split}
\textcolor{darkgreen}{f(s)} &= \frac{\sum_{i=1}^{N}w_{i}\textcolor{darkgreen}{\Psi_i(t)}}{\sum_{i=1}^{N}\textcolor{darkgreen}{\Psi_i(t)}}\textcolor{darkgreen}{s},\\
\textcolor{darkgreen}{\Psi_i(t)} &= \exp{\left(-\frac{(\textcolor{darkgreen}{\frac{t}{\tau T}}-c_i)^2}{\textcolor{darkgreen}{2\sigma^{2}_i}}\right)}, 
\end{split}
\label{eq:fx_discrete_overlay}
\end{equation}
where $\sigma_{ i }$ is the width and $c_{ i }$ is the center of the $i$-th basis function, and $s$ is obtained by integrating \eqref{eq:sigmoidal_decay}. The term $t/\tau T$ is used in \eqref{eq:fx_discrete_overlay} instead of the the phase variable $x$. Being $0\leq t/\tau T \leq 1$, the basis functions are equally spaced between $0$ and $1$. Finally, $\sigma_{i}$ are the widths of each kernel. They are constant and depend on the number of kernels.

Having presented the main differences with the canonical approach, it is possible to focus on how \cite{kulvicius2012joining} solved the problem of joining $L\geq 2$ \acp{dmp}. In general, each of the $L$ \acp{dmp} has a different time duration $T^l$, desired target $g^l$, and initial position $y^l_0$, from which  it is possible to compute the delayed goal functions by integrating
\begin{equation} 
\tau \dot{\tilde{g}}^l= 
\begin{cases}
\frac{\delta t}{T^l}(g^l-y^l_0), & \sum\limits_{ \kappa=1 }^{ l-1 }{ T^{ \kappa } \leq t \leq \sum\limits_{ \kappa=1 }^{ l }{ T^{ \kappa } }} \\
0, &\text{otherwise}
\end{cases} .
\label{eq:linear_delayed_goal_functions}
\end{equation}
Note that, being $\tilde{g}^l(0) = y_0$, the acceleration \eqref{eq:dmp_discrete1_overlay} is smooth at the beginning of the motion. For this reason, the term $(1-x)$ used in~\eqref{eq:dmp_discrete_crossing1} is not needed in~\eqref{eq:dmp_discrete1_overlay}.

Assuming that $L$ \acp{dmp} have been trained and that each \ac{dmp} has $N$ kernels, we can merge them into one \ac{dmp} as follows. The centers of the joined \ac{dmp} are computed as
\begin{equation} 
\check{c}^{ l }_{ i }=\begin{cases} \frac{ T^{ 1 }( i-1 )}{ T_{join}( N-1 )}, &l=1\\
\frac{ T^{ l }( i-1 )}{ T_{join}( N-1 ) } + \frac{ 1 }{ T_{join}} \sum\limits_{ \kappa=1 }^{ l-1 }{ T^{ \kappa } }, & \text{otherwise}
 \end{cases}, \label{eq:centres} 
\end{equation}
where $T^{l}$ is the duration of the $l$-th \ac{dmp}, and $T_{join} = \sum_{l=1}^{L}{ T^{l}}$ (duration of the joined motion). The widths of the joined \ac{dmp} are computed as
\begin{equation}
\check{\sigma}^{l}_{i}=\frac{ \sigma^{l}_{i}T^{l}}{T_{join}}.
\label{eq:newsigmas}
\end{equation}
The centers and widths computed in~\eqref{eq:centres} and \eqref{eq:newsigmas} respectively overlap at the transition points allowing for smooth transitions between consecutive \acp{dmp}.
The weights of the joined \ac{dmp} are obtained by stacking the $N$ weights of the $L$ \acp{dmp}. Therefore, the joined \ac{dmp} has $N*L$ kernels and $N*L$ weights. The phase variable \eqref{eq:sigmoidal_decay} is modified to run for the duration~$T_{join}$ of the joint motion. 

\cite{saveriano2019merging} extended the basis functions overlay approach to quaternion \acp{dmp}. Assuming that a sequence of $L$ quaternion \acp{dmp} is given. The angular acceleration in~\eqref{eq:quat:dw} is reformulated for each \ac{dmp} as 
\begin{equation}
\tau \bm{\Dot{\eta}}^l = \alpha_z(\beta_z 2 \, \LogQ(\textcolor{darkgreen}{\tilde{\bm{g}}^l_q}*\overline{\q}^l)-\bm{\eta}^l) + \textcolor{darkgreen}{\bm{f}^l_q(s)}
\label{eq:quat:dw_overlay}
\end{equation}
where $l$ indicates the $l$-th quaternion \ac{dmp} and $\bm{f}^l_q(s)$ is defined as in~\eqref{eq:fx_discrete_overlay}. The term $\tilde{\bm{g}}^l_q$ is the \textit{quaternion delayed goal function} and it ranges from $\q^l(0)$ to $\bm{g}^l_{q}$ in $T^l\,$ seconds (see \figref{fig:goal_functions}~(right)). To generate this moving target while preserving the geometry of $\bm{\mathcal{S}}^3$ it is needed that $\tilde{\bm{g}}^l_q$  moves along the geodesic connecting $\q^l(0)$ to $\bm{g}^l_{q}$. Therefore,  $\tilde{\bm{g}}^l_q$ is defined as
\begin{equation}
    \tilde{\bm{g}}^l_q(t+\delta t) = \ExpQ\left(\frac{\tau\tilde{\bm{\omega}}^l(t)}{2}\right)\ast\tilde{\bm{g}}^l_q(t)
\end{equation}
where
\begin{equation} 
\tilde{\bm{\omega}}^l(t) = \begin{cases}
\frac{2}{T^l}\LogQ\left(\bm{g}^l_{q}  \ast \overline{\q}^l(0)\right) & \sum\limits_{ \kappa=1 }^{ l-1 }{ T^{ \kappa } \leq t \leq \sum\limits_{ k=1 }^{ l }{ T^{ \kappa } }}\\  [0,\,0,\,0]\trsp, & \text{ otherwise } \end{cases}. \label{eq:qdelayedgoalfunction}
\end{equation}
The angular velocity in~\eqref{eq:qdelayedgoalfunction} is computed for each $l$. The term $2\LogQ\left(\bm{g}^l_{q}  \ast \overline{\q}^l(0)\right)$ represent the angular velocity that rotates $\q^l(0)$ into $\bm{g}^l_{q}$ in a unit time. Note, that the mappings $\LogQ(\cdot)$ and  $\ExpQ(\cdot)$ are defined in~\eqref{eq:quat:log} and~\eqref{eq:quat:exp} respectively. The delayed goal $\tilde{\bm{g}}^l_q$ crosses all the via-goals $\bm{g}^{l}_q,\,l=1,\ldots,L-1$ and then reaches the goal $\bm{g}^{L}_q$.

Results in \Figref{fig:Joining_overlay} are obtained when velocity threshold is applied to merge $2$ \acp{dmp} separately trained to fit the minimum jerk trajectories (black dashed lines). \figsref{fig:Joining_overlay}a--\ref{fig:Joining_overlay}e  show the position and \figsref{fig:Joining_overlay}f--\ref{fig:Joining_overlay}j  the orientation (unit quaternion) parts of the motion. This approach does not require a switching rule and automatically generates a smooth trajectory---with continuous velocity as shown in \figsref{fig:Joining_overlay}c~and~\ref{fig:Joining_overlay}h---that passes close to the via-point which favors the overall reproduction accuracy (\figref{fig:Joining_overlay}e~and~\ref{fig:Joining_overlay}j). However, the distance from the via-point depends on the weights of the joined primitives and cannot be separately decided. The trajectory generated with this approach tend to last longer than the demonstrations. This is due to the sigmoidal phase that vanishes after $T+\delta_s\,$s (\figref{fig:phase_dmp}). Depending on the application, the time difference may cause failures and has to be taken into account.

\subsection{Online adaptation}
\label{subsec:onlineUpdate}

The standard periodic \ac{dmp} learning approach approximates the shape $f_d(t)$ of the input trajectory $y_d$ in (\ref{eq:goal}) by changing the weights of the Gaussian kernel functions~\citep{Ijspeert2013Dynamical}. Updating of the weights is performed in such a way that the difference between the reference trajectory and the \ac{dmp} is reduced at every control step and gradually throughout the periodic repetitions. However, the \ac{dmp} can also be reshaped by some external feedback function to achieve different functionalities for different applications, for instance, tasks that require trail-and-error approach \citep{kober2008learning}, obstacle avoidance \citep{park2008movement, hoffmann2009biologically, tan2011potential}, coaching \citep{Petric2014Onlineapproach,gams2016adaptation} for robots, and adaptation of assistive exoskeleton behavior \citep{peternel2016adaptive}. Alternatively, the frequency of the existing periodic \acp{dmp} can be modulated online \citep{Gams2009online,Petric2011online}.








\subsubsection{Robot obstacle avoidance and coaching}
\label{subsubsec:obstacleAvoidance}
In \citep{park2008movement, hoffmann2009biologically, tan2011potential} the detected obstacle was fitted with a potential field function to change the shape of the \ac{dmp} to avoid it. More in details, \cite{tan2011potential} used the potential field to compute a time-varying goal and modified the resulting \ac{dmp} trajectory, while \citep{park2008movement, hoffmann2009biologically} added and extra forcing term to the \ac{dmp}. Similarly in \citep{gams2016adaptation} the human arm was fitted with a potential field function, which was used to reshape the \ac{dmp} to perform coaching. The potential field was coupled to the position of the human hand to make pointing gestures and indicate the direction in which the robot arm position trajectory should change:
\begin{equation}
\dot z = \Omega \left( {\alpha \left( {\beta  \left( - y\right) - z} \right) + C_\mathcal{O} + f} \right).
\label{eq:avoidance}
\end{equation}
The added coupling term $C_\mathcal{O}$ is the obstacle avoidance term that contains the potential field and is given in a simplified form for the sake of explanation as:
\begin{equation}
C_\mathcal{O} = d_s(||\mathcal{O}-y||) \exp(-\zeta(\mathcal{O}-y)),
\label{eq:avoidance2}
\end{equation}
where $\mathcal{O}$ is the obstacle (or human pointing gesture) and $y$ is the robot position. Exponential and $\zeta$ functions determine the potential field, while function $d_s$ controls the distance at which the perturbation field should start affecting the \ac{dmp}. For the full formulation of $C_\mathcal{O}$ and its parameters, see \citep{gams2016adaptation}. In \citep{Rai2017Learning} the method was extended to include generalization of the obstacle avoidance formulation in \eqref{eq:avoidance}. 

Alternatively, the faulty segment of collision \ac{dmp} trajectory can also be directly adjusted online by the human demonstrator \citep{Karlsson2017Autonomous}. On the other hand, the method in \citep{Kim2015Adaptability} considers obstacle avoidance as a constraint of an optimization problem, which modifies the  \ac{dmp} trajectory to prevent collisions.




\subsubsection{Robot adaptation based on force feedback}
\label{subsubsec:Forcecoupling}

Similarly as for obstacle avoidance, task dynamics can also be incorporated into \ac{dmp} as coupling terms.
In \citep{Gams2014} task dynamics were coupled on the acceleration and velocity level of the \ac{dmp}. The presented method was utilized for interaction tasks, where the human changed the behavior of the robot based on the exerted dynamics on the manipulator. 
\begin{align}
\tau \dz &= \alpha_z(\beta_z(g-y)-z)+ \dot{C}_f + f(x), \label{eq:force_feedback1}\\
\tau \dy &= z + C_f. \label{eq:force_feedback2}
\end{align}
whereas the force coupling term $C_f = \varsigma F$ is defined as a virtual or measured force $F$ and $\varsigma$ is a scaling factor, which essentially changes the dynamic behavior of the \ac{dmp}, enabling the motion primitive to instantly react to the coupled force.  Later, \cite{Zhou2016Learning} introduced a PD controller based coupling term formulation $C_{PD}= \varsigma (K^\mathcal{P}(F_d-F^e)-D^\mathcal{V}\dot{F}^e)$ coupled to the velocity part of the \ac{dmp} \eqref{eq:force_feedback2}. In the formulation $F_d$ represents the desired force, $F^e$ is the measured force, $\varsigma$ is a scaling factor and $K^\mathcal{P}$ and $D^\mathcal{V}$ are the proportional and derivative gains of the \ac{pd} controller. The coupling term formulation allows for controlled adaptation of robot motion to changes in the environment.

In \citep{kramberger2018passivity} this approach was extended, with a force feedback loop coupled to the velocity \eqref{eq:dmp_discrete2} and the goal $g$ of the \ac{dmp}. The outcome of this approach is a similar behavior as an admittance controller \citep{villani2008force}, with an difference that the  execution is directly on the trajectory generation level.
\begin{align}
\tau \dz &= \alpha_z(\beta_z((g+C_a)-y)-z)+f(x), \label{eq:force_feedback3}\\
\tau \dy &= z + \dot{C}_a. \label{eq:force_feedback4}
\end{align}
Here $ \dot{C}_a = \varsigma(F_d - F^e)$ is the first time-derivative of the admittance coupling term, which changes the velocity and consequently the integrated coupling term, the position output of the \ac{dmp}. The described approach can be used for Cartesian space motion, where the forces have to be substituted for desired and measured torques. This approach can be implemented in robot tasks involving contact with the environment as well as contact with humans.

\subsubsection{Exoskeleton joint torque adaptation}
\label{subsubsec:exoskeleton}

In \citep{peternel2016adaptive}, human effort was used to provide the information about the direction in which the assistive exoskeleton joint torque \ac{dmp} should change in order to minimize it. The human was included into the robot control loop by replacing the error calculation in \eqref{eq:reg3} with the human effort feedback term $U(E)$:
\begin{equation}
w_{i}(t_{\jmath+1}) = w_{i}(t_{\jmath})+\Psi_{i}P_{i}(t_{\jmath+1})U(E)\label{eq:reg1mod},
\end{equation}
where $E(t)$ is the current effort measured by human muscle activity through \ac{emg} signals\footnote{Note that other feedback that measures human effort can be used instead of \ac{emg}, such as joint torque or limb forces.}. Equations~\eqref{eq:dmp_periodic1}-\eqref{eq:psi_periodic} and \eqref{eq:reg2} are used in the original form. Equations~\eqref{eq:goal}-\eqref{eq:reg3} are not used, since \eqref{eq:reg1mod} is used to modulate the weights in \eqref{eq:fx_periodic} instead. 

The effort feedback term $U(E)$ closes the loop and acts as a feedback for adapting the weights of Gaussian kernels that define the shape of the trajectory. A positive $U(E)$ increases, while a negative $U(E)$ decreases the values of weights at a given section of the periodic \ac{dmp} that encodes joint torque. If the shape of the \ac{dmp} does not provide enough assistive power, the human has to exert effort (\ie muscle activity) to produce the rest of the power required to achieve the desired task under given dynamics. In turn, muscle activity feedback then increases the magnitude of the \ac{dmp} until the human effort term $U(E)$ is minimised. Note that each joint has its own torque \ac{dmp} and $U(E)$ term \citep{peternel2016adaptive}. After that point, the \acp{dmp} do not change unless the task, dynamics or conditions change. If they change, the human has to compensate for the change by an additional muscle activity, which in turn adapts the \acp{dmp} to the new required joint torques.

\subsubsection{Trajectory adaptation based on reference velocity}
\label{subsec:velocityTerm}

In many \ac{lfd} scenarios it is desired to modify both the spatial motion and the speed of the learned motion at any stage of the execution. Speed-scaled dynamic motion primitives first presented in \cite{nemec2013velocity} are applied for the underlying task representation. The original \ac{dmp} formulation from \eqref{eq:dmp_discrete1} and \eqref{eq:dmp_discrete2} were extended by adding a temporal scaling factor $\upsilon$ on the velocity level of the \ac{dmp}
\begin{align}
\upsilon(x)\tau \dz &= \alpha_z(\beta_z(g-y)-z)+f(x), \label{eq:velocity_scalling1}\\
\upsilon(x)\tau \dy &= z. \label{eq:velocity_scalling2}
\end{align}
Form \eqref{eq:velocity_scalling1} and \eqref{eq:velocity_scalling2}, it is evident that the velocity term is a function of phase, and therefore encoded with a set of \acp{rbf} similarly as in \eqref{eq:fx_discrete}. This method allows for modification of the spacial motion as well as the speed of the execution at any stage of the trajectory execution. The authors demonstrated the proposed method in a learning scenario, where after every learning cycle (using \ac{ilc}) a new velocity profile was encoded based on the wrench feedback, and thus converged to an optimal velocity for the specific task. \cite{Vuga2016Speed} extended the approach 
by incorporating a compact representation for non-uniformly accelerated motion as well as simple modulation of the movement parameters.

Later on, in \cite{nemec2018anefficient} the authors extended the previous approach to also incorporate velocity scaling of the encoded orientation trajectories represented with unit quaternions. The outcome of the presented work is a unified approach to velocity scaling for tasks executed in Cartesian space. Furthermore, a reformulation of the velocity approach called AL-\acp{dmp} was presented by \cite{Gaspar2018}. In this work they present a method, where the spatial and temporal components of the motion are separated, by means of the arc-lenght based on the time parameterized trajectory. Arc-lenght, based on the differential geometry of curves, is related to the speed of the movement, given as the time derivative of the demonstrated trajectory. The approach is well suited when multiple demonstrations are compared for extraction of relevant information for learning. 
\cite{weitschat2018safe} add an extra forcing term to keep the velocity within a certain predefined limit. The aim of this work is to guaranty a safe execution of the robot task when interacting with humans, as well as providing a framework for safe interaction in a changing environment where the robot position and velocity have to change over time. For a full formulation of the coupling term see \citep{weitschat2018safe}. Additionally, \cite{Dahlin2020438} in their work proposed a temporal coupling based on a repulsive potential, keeping the \ac{dmp} velocity within the predefined velocity limits while ensuring the path shape invariance.

 \subsection{Alternative formulations}
\ac{lfd} is a wide research area and many different approaches have been developed to reproduce human demonstrations \citep{Billard16learning}. As already mentioned, the aim of this tutorial survey is to provide a comprehensive overview of \acp{dmp} research and we intentionally skip the rich literature in the field of \ac{lfd}. However, we found some representations that are closely related to the \ac{dmp} formulation. This section briefly reviews them.   

\cite{calinon2009handling} computed an acceleration command for the robot in a PD-like form
\begin{equation*}
    \bm{\ddot{y}} = \Kp(\bm{y}_d - \bm{y}) + \Dv(\bm{\dot{y}}_d - \bm{\dot{y}}),
\end{equation*}
where $\Kp$ is a stiffness and $\Dv$ a damping gain, $\bm{y}$ is the measured state of the robot and $\bm{\dot{y}}$ its time derivative (velocity), $\bm{y}_d$ and $\bm{\dot{y}}_d$ are desired position and velocity retrieved with \ac{gmr}. Authors then shown that the acceleration command $\bm{\ddot{y}}$ can be seen as a mixture of linear dynamics, each converging to a certain attractor. Despite later work like \citep{kormushev2010robot} referred to this representation as ``a  modified  version'' of \acp{dmp} there are significant differences with the \ac{dmp} formulation properly highlighted by \citep{calinon2012statistical}.  

\cite{Herzog2016Optimal} computed an acceleration command for the robot from the linear system
\begin{equation*}
    \bm{\ddot{y}} = \bm{u} = \Kp(\bm{y}_d-\bm{y}),
\end{equation*}
where $\bm{y}$ is the measured state of the robot, $\bm{y}_d$ is a human demonstration, and $\Kp$ is a control gain computed using the linear-quadratic  regulator  method. Then, a compact representation of the control input trajectory $\bm{u}$ is computed by means of Chebyshev polynomials. This representation does not require a vanishing phase variable to ensure convergence, but the generalization to different start/goal position requires the application of the linear-quadratic regulator method to find a new sequence of control inputs.

Regarding periodic motions, \citep{ajallooeian2013general} proposed a dynamical system-based framework to learn rhythmic movements with an arbitrary shape and basin of attraction. They exploit phase-based scaling functions to represent the mapping between a known, base limit cycle and a desired periodic orbit. The basic limit cycle can be, for example, the one generated by a periodic \acp{dmp}, which makes the approach of \citep{ajallooeian2013general} a more general formulation of periodic primitives.

%% file: DMP_System_Integration.tex
\section{\texorpdfstring{\acp{dmp}}{} integration in complex frameworks}
\label{sec:integration}

This section reviews approaches where \acp{dmp} have been integrated into bigger executive frameworks. We categorize these approaches into five main research areas, namely \textit{grasping and manipulation}, \textit{impedance learning}, \textit{reinforcement learning}, \textit{deep learning}, and \textit{incremental and life-long learning}

\subsection{Manipulation tasks}\label{subsec:manipilation}
Successfully grasping an object is the first step towards robotic manipulation. Performing a grasping requires a (visual) perception of the environment to locate the object to grasp and decide the grasping points based on its geometry. In this setting, even small uncertainties may cause the object to drop and the grasp to fail. To improve the robustness of vision driven grasping, \cite{kroemer2010grasping} augmented \acp{dmp} with a potential field based on visual descriptors that adapts hand and finger trajectories to the object’s local geometry. This grasping strategy was integrated in a hierarchical control architecture where the upper level decides where to grasp the object and the lower level locally adapted the motion to robustly grasp the object \citep{kroemer2010combining}.
\cite{stein2014convexity} proposed a point cloud segmentation approach based on convexity and concavity of surfaces. The approach is particularly suited to recognize object handles and enables a robot to automatically grasp object.

The ability of grasping and using tools is also desirable to perform daily-life manipulation. In this respect, \citep{guerin2014adjutant} proposed the so-called \textit{tool movement primitives} that transform the demonstrations in a tool affordance frame. The result is a motion that generalize to different tool poses and to tools that share the same affordance(s). \cite{Li2015Teaching} considered tool usage with low-cost, non-dexterous grippers and propose a framework to learn bi-manual strategies for tool usage and compensate for the lack of dexterity. Bi-manual robotic manipulation is a challenging task that requires precise coordination between the hand movements and adherence to the spatial constrains. \cite{Thota2016Learning} developed a \ac{dmp}-based control framework for bi-manual manipulation that ensures time synchronization of the two hands while being robust to spatial perturbations and goal changes.

\begin{figure}[t]
    \centering
    \includegraphics[width=\columnwidth]{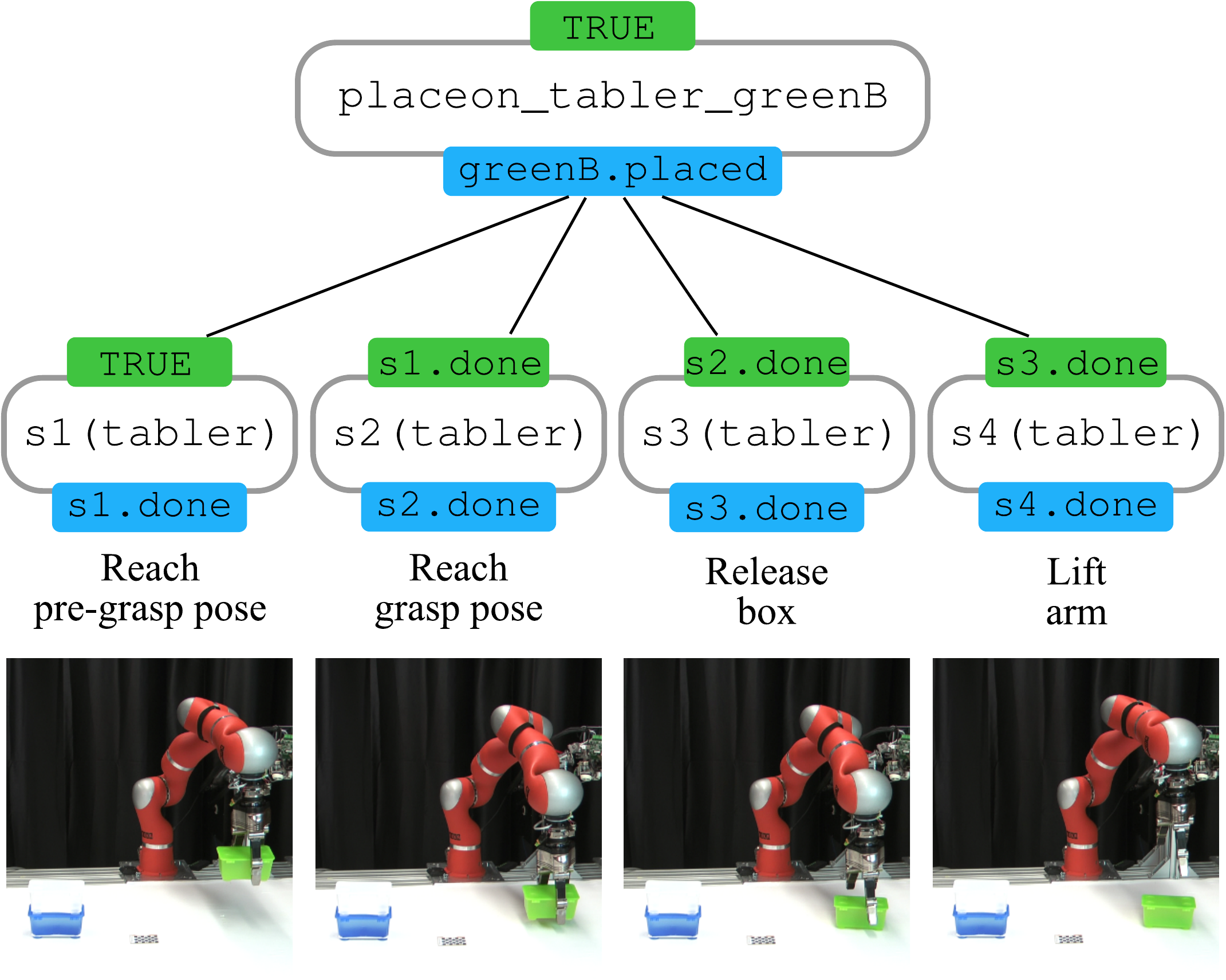}
    \caption{An example of hierarchical task decomposition and motion primitives sequencing from \citep{agostini2020manipulation}.}
    \label{fig:agostini_2020}
\end{figure}

Beyond the object grasping, everyday manipulation requires a precise execution of complex movements. Often such a complex movements are hard to encode into a single motion primitive, but they can be conveniently split into simpler motions (\eg reach and grasp) that can be properly sequenced and executed (\figref{fig:agostini_2020}).

The possibility of exploiting \acp{dmp} as the building blocks of complex tasks was investigated in \citep{Ramirez-Amaro2015understanding, caccavale2018imitation,caccavale2019kinesthetic}. In these works, a human teacher demonstrated a relatively complex task consisting of several actions performed on different objects. The demonstration was then automatically segmented into $M$ basic motions used to fit $M$ \acp{dmp}. While \cite{Ramirez-Amaro2015understanding} exploit semantic rules (\eg, \textit{reach an object with a knife means cut}) to infer high-level human activities, \citeauthor{caccavale2019kinesthetic} built a hierarchical structure to schedule the execution of the complex task by selecting the proper \ac{dmp} for the current executive context. They used kinesthetic teaching and verbal cues (open/close gripper commands) to provide task demonstrations. \cite{Lemme2014Selfsupervised} organize segmented task demonstrations into a motion primitives library learned from self-generated trajectory patches. They also introduced a mechanism to remove unused skills and update the library.  
 Kinesthetic teaching and haptic feedback were also used by \cite{eiband2019learning} to segment and recognize basic motions or \textit{skills}, and to build a tree describing geometric relationships---like reference frames and goal poses---between consecutive skills. At run time, the robot performed haptic exploration to locate objects in the scene and update the skill tree. The transformations in the skill tree were then used to define initial and goal pose of the \acp{dmp} and execute the task. Finally, \cite{Wu2018} integrated \acp{dmp} into a dialogue system with speech and ontology to learn or re-learn a task using natural interaction modalities.
 
 Collecting demonstrations becomes an issue of kinesthetic teaching or marker-based motion trackers cannot be used. The latter requires an expensive sensor infrastructure that is hard to build in real world scenarios like factory floors. Kinesthetic teaching needs torque controlled/collaborative robots that are still uncommon in industrial scenarios. To remedy this issue \citep{Mao2015Learning} exploited a low-cost RGB-D camera and track the human hand using the markerless approach proposed by \citep{Oikonomidis}. Collected data were then segmented into basic motions and used to fit \acp{dmp}.
 
 Described approaches assumes that human teachers always provide consistent and noiseless task demonstrations. \cite{Ghalamzan2015Anincremental} encoded noisy demonstrations into a \ac{gmm} and computed a noise free trajectory using \ac{gmr}. The noise free trajectory was then used to fit a \ac{dmp} that generalized to different start, goal, and obstacle configurations. \cite{niekum2012learning, niekum2015learning} designed a framework that learns from from unstructured demonstrations by segmenting the task demonstrations, recognizing similar skills, and generalizing the task execution. Interestingly, a user study on $10$ volunteers conducted by \citep{Gutzeit2018} showed that existing strategies for segmentation and learning are sufficiently robust to enable automatic transfer of manipulation skills from humans to robots in a reasonable time. Finally, some work \citep{denisa2013discovering, denisa2013newmotor,Denisa2015Synthesis} exploited transition graphs and trees to embed parts of a trajectory and search algorithms to discover sequence of partial parts and generate motions that have not been demonstrated.  

Approaches that rely on a hierarchical, tree-like structure to represent the task that has limited task generalization capabilities. \cite{lee2013skill} used probabilistic inference and object affordances to infer the adequate skill that can handle  uncertainties in the executive context. \cite{beetz2010generality} learned stereotypical task solutions from observation and used task planning and symbolic reasoning to execute novel mobile manipulation tasks. A generative learning framework was proposed by \citep{Worgotter2015Structural} to augment the robot's knowledge-base with missing information at different level of the cognitive architecture, including symbolic planning as well as object and action properties. \citep{paxton2016dowhat} used task and motion planning to generalize the execution of complex assembly tasks and proposed an learning by demonstration approach to ground symbolic actions. \citep{agostini2020manipulation} performed task and motion planning by combining an object-centric description of geometric relations between objects in the scene, a symbol to motion hierarchical decomposition depending on tree consecutive actions in the plan, and the \ac{lfd} approach developed in \citep{caccavale2019kinesthetic} (\figref{fig:agostini_2020}). A manipulation task was described at three different levels by \citep{aein2013toward}. The top-level provides a symbolic descriptions of actions, objects, and their relationships. The mid-level uses a finite state machine to generate a sequence of action primitives grounded by the lower level. A common point among these approaches is that they use \ac{dmp} to execute the task on real robots.

\subsection{Variable impedance learning control}
\label{subsec:vilcDMP}

Impedance control can be used to achieve complaint motions, in which the controller resembles a virtual spring-damper system between the environment and robot end-effector \citep{hogan1985impedance}. Such approach permits smooth, safe, and energy-efficient interaction between robots and environments (possibly humans). A standard model for such interaction is defined as
\begin{align}
\bm{\mathfrak{M}}\bm{\ddot{y}}_t &= \Kpt(\bm{y}_g-\bm{y}_t)-\Dvt\bm{\dot{y}}_t+\fet, 
\label{eq:msd_trans} \\
\bm{\mathcal{I}}\bm{\dot{\omega}}_t &= \Kot(\LogR(\R_g\bm{R}_t\trsp))-\Dwt\bm{\omega}_t+\tauet,
\label{eq:msd_ori}
\end{align}
where \eqref{eq:msd_trans} and \eqref{eq:msd_ori} correspond to translational and rotational cases respectively, 
$\bm{\mathfrak{M}}, \Kpt, \text{and } \Dvt$ are the mass, stiffness and damping matrices, respectively, for translational motion, while $\bm{\mathcal{I}}, \Kot, \text{and } \Dwt$ are the moment of inertia, stiffness and damping matrices, respectively, for rotational motion.
$\bm{\hat{R}},\bm{R}_t\in\bm{\mathcal{SO}}(3)$ are rotation matrices and correspond to desired rotation goal and actual orientation profile of the end-effector, respectively. 
$\fet$ and $\tauet$ represent the external force and torque applied to the robot end-effector.

\begin{figure}[t]
	\centering
	\includegraphics[width=\columnwidth]{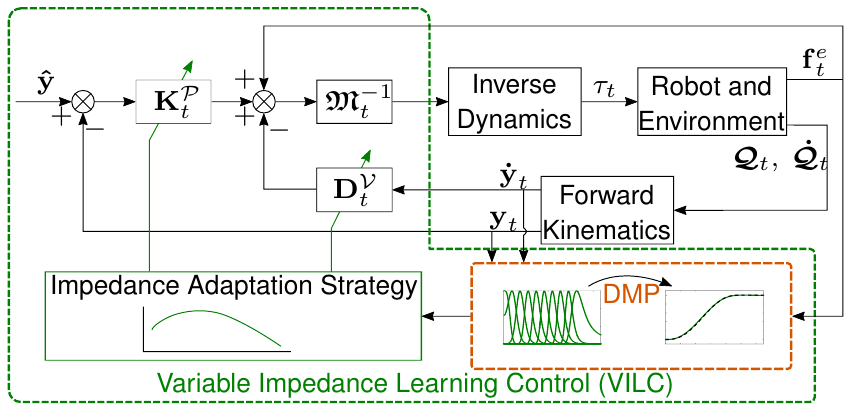}
	\caption{General control scheme of \ac{vic} and \ac{dmp}.}
	\label{fig:vilc_dmp}
\end{figure}

In fact, \ac{vic} plays an important role when a robot needs to interact with any environment in order to avoid high impact forces and damage for the environment or the robot (\ie change to low stiffness)\cite{ajoudani2012tele,abudakka2018force,peternel2018robotic}. On the other hand, it is important in rejecting unexpected and unpredictable perturbations from the environment to achieve a desired position tracking precision (\ie change to high stiffness) \cite{Yang2011}. In addition, it is also important in coordination of human-robot collaborative movements \cite{peternel2017human}. However, a robotic system still needs to learn how to adapt such \ac{vic} to unseen situations while avoiding hard-coding. Such paradigm of learning is called \acf{vilc}. Interested readers can refer to our recent survey on \ac{vilc} \citep{abudakka2020Variable}. 

In this review, we will mention some of the works that integrate \ac{dmp} with \ac{vic} in a \ac{vilc} framework. \Figref{fig:vilc_dmp} shows a simple generic example where \ac{dmp} is integrated in a \ac{vic} control scheme.

\cite{buchli2011learning} proposed one of the earliest approaches that integrates \ac{dmp} with \ac{pi2} algorithm \citep{theodorou2010generalized} to learn movements (position and velocity presented by \ac{dmp}) while optimizing impedance parameters. Later the authors exploited a diagonal stiffness matrix and expressed the variation (time derivative) of each diagonal entry as 
\begin{equation}
\dot{k}_{\vartheta_j,t} = \alpha_j \left(\bm{\curlywedge}_j\trsp(\bm{\vartheta}_j+\bm{\epsilon}_{j,t})- k_{\vartheta_j,t} \right), \quad j=1,\ldots,J ,
\label{eq:pi2_stiffness}
\end{equation}
where $j$ indicates the $j$-th joint, $k_{\vartheta_j,t}$ is the stiffness of joint $j$, $\bm{\epsilon}_{j,t}$ is a time-dependent exploration noise, each $\bm{\curlywedge}_j$ is a vector of $N$ Gaussian basis functions, and $\bm{\vartheta}_j$ are the learnable parameters for joint $j$. The stiffness parameterization in~\eqref{eq:pi2_stiffness} is also linear in the parameters and \ac{pi2} can be applied to find the optimal policy. Later, authors used \ac{pi2} to learn  \ac{vic} in deterministic and stochastic force fields \citep{stulp2012model}. \cite{nakanishi2011stiffness} proposed a method that optimizes a periodic motion a long with a time-varying joint stiffness. 


\citep{basa2015learning} introduced an extension to \ac{dmp} formulation by adding a second nonlinear function to cope with elastic robots as follow
\begin{equation}
\tau \dz = \alpha_z(\beta_z(g-y)-z) + f(x) + \textcolor{darkgreen}{f_2},
\label{eq:dmp_discrete1_f2}
\end{equation} 
where $\textcolor{darkgreen}{f_2}$ is defined as \eqref{eq:fx_discrete} but without the phase variable $x$. The main purpose of $\textcolor{darkgreen}{f_2}$ is to compensate the gravitational influence on the moved \ac{dof} at the end of the movement time and beyond. Differently, \cite{Haddadin2016Optimal} used optimal-control to execute near-optimal motion of elastic robots.

\cite{Nemec2016Bimanual} proposed a cooperative control scheme that enables dual arm robot to adapt its stiffness online along to the executed trajectory in order to provide accurate evolution. \citep{Umlauft2017Bayesian} used \ac{gp} along with \acp{dmp} (as proposed in \citep{Fanger2016Gaussian}) to predict the trajectories. During the execution, their admittance controller adapts both stiffness and damping online. 
The energy-tanks passivity-based control method has been integrated with \acp{dmp} to enforce passivity in order to stably adapt to contacts in unknown environments by adapting the stiffness online \citep{Shahriari2017,kramberger2018passivity, kastritsi2018progressive}.


Methods in \citep{peternel2014teaching,peternel2018robot,peternel2018robotic,yang2018dmps,yang2019learning,Bian2019extended} designed different multi-modal interfaces to let the human to explicitly teach an impedance behavior to the robot. Most of them combined \ac{emg}-based variable impedance skill transfer with \ac{dmp}-based motion sequence planning, inheriting the merits of these two aspects for robotic skill acquisition. \cite{hu2018evolution} used \ac{cmaes} to update the parameters of \acp{dmp} and variable impedance controller in order to reduce the impact in during the robot motion in noisy environments. \cite{Dometios2018Vision} integrated a \ac{ccdmp} with a vision-based motion planning method to adapt the reference path of a robot’s end-effector and allow the execution of washing actions.

\cite{Travers2016Shape, Travers2018Shape} proposed a shape-based compliance controller for the first time in locomotion, by implementing amplitude compliance on a snake robot moving in complex environment with obstacles. Their approaches allow a snake-like robots to blindly adapt to such complex unstructured terrains thanks to their proprioceptive gait compliance techniques

Recently, an adaptive admittance controller is proposed \citep{WANG2020Arobot} which integrates \ac{gmr} for the extraction of human motion characteristics, \ac{dmp} to encode a generalizable robot motion, and a \acs{rbfnn}-based controller for trajectory-tracking during the reproduction phase.

Novel \ac{lfd} approaches explicitly take into account that training data are possibly generated by certain Riemannian manifolds with associated metrics. 
\cite{abudakka2020Geometry} reformulated \acp{dmp} based on Riemannian metrics, such that the resulting formulation can operate with \ac{spd} data in the \ac{spd} manifold. Their formulation is capable to adapt to a new goal-\ac{spd}-point.

Recently, biomimetic controller has been integrated with \acp{dmp} \citep{ZENG2021Learning} in order to learn and adapt compliance skills.

\subsection{\texorpdfstring{\acf{rl}}{}}
\label{subsec:rl}

\begin{figure}[t]
	\centering
	\includegraphics[width=\columnwidth]{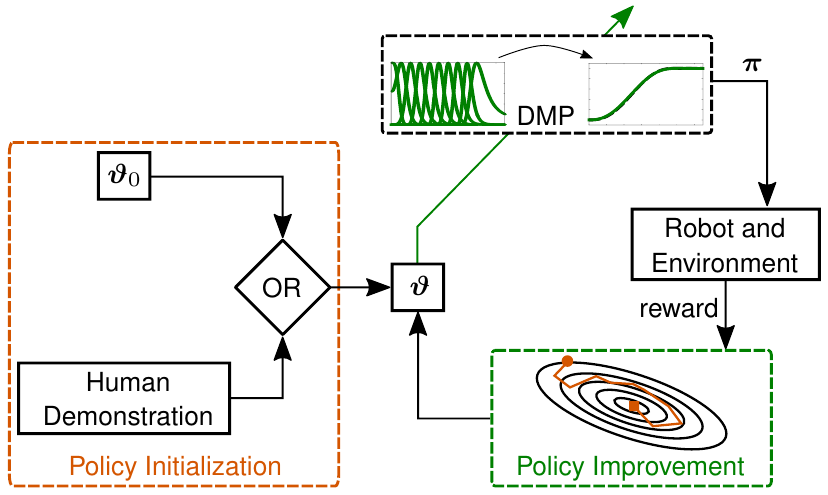}
	\caption{General block scheme of \ac{dmp}-based policy improvement.}
	\label{fig:rl_dmp}
\end{figure}

In \ac{rl}, an agent tries to improve its behavior via trial-and-error by exploring different strategies (\textit{actions}) and receiving a feedback (\textit{reward}) on the outcome of its actions. Actions $a$ are drawn from a \textit{policy} $\pi(s,a)$ that represent a mapping between \textit{states} $s$ and actions $a$. The goal of \ac{rl} is to find an optimal policy $\pi^{\star}$ that maximizes the cumulative expected reward, \ie the sum of expected rewards over a possibly infinite time interval. When the agent is a robot performing tasks in the real world the state and actions spaces are inherently continuous. Moreover, the robotic agent is affected by imperfect (\eg noisy) perception and inaccurate models (\eg contacts). Finally, performing a large amount of interactions with the real word (\textit{rollouts}) is expensive and possibly dangerous. As discussed by \citep{Kober2013reinforcement}, robotic specific challenges require specific solutions to make the \ac{rl} problem feasible.

\subsubsection{\texorpdfstring{\acp{dmp}}{} as control policies}
One possibility is to use parameterized policy and use \ac{rl} to search for an optimal, finite set of policy parameters. In this respect, \acp{dmp} have been widely used as policy parametererization. The general idea is shown in \Figref{fig:rl_dmp}. More in details, \citep{peters2008policy, peters2008reinforcement} showed that various policy gradient and actor-critic \ac{rl} approaches can be effectively applied to improve robotic skills parameterized as \acp{dmp}. Other research focused on developing policy search algorithms specifically for parameterized policies. Inspired by stochastic optimal control, \cite{theodorou2010generalized} proposed Policy Improvement with Path Integrals (\ac{pi2}) which is an application of path integral optimal control to \acp{dmp}. \ac{pi2} and \acp{dmp} have been successfully applied in several domains including \ac{vilc} \cite{buchli2011learning, buchli2010variable} and in-contact tasks \citep{Hazara2016Reinforcement}, grasping under state estimation uncertainties \citep{stulp2011learning}, bi-manual manipulation \citep{Zhao202018}, and robot-assisted endovascular  intervention \citep{Chi2018}.
\cite{kober2011policy} derived from expectation-maximization the so-called \ac{power}. \ac{power} and \acp{dmp} have been successfully applied to perform highly dynamic tasks including ball-in-a-cup \cite{kober2011policy} and pancake flipping \cite{kormushev2010robot}.


\subsubsection{Limit the search space}
Even with parameterized policies the number of rollouts needs to search for optimal policy parameters may become large, especially for robots with many \acp{dof}. Dimensionality reduction techniques can be exploited to perform policy search in a reduced space \citep{colome2014dimensionality}. The effectiveness of this approach was demonstrated in the challenging task of clothes (\ie soft tissues) manipulation \citep{colome2018dimensionality}. \ac{il} arises as an effective approach to policy initialization and to speed up policy search by reducing the number of rollouts \citep{kober2010imitation}. In this respect, \cite{kober2008learning, kober2010imitation2} augmented \acp{dmp} with a perceptual coupling term and propose to initialize the \ac{dmp} via human imitation and to refine the motor skill via \ac{rl}. \ac{il} can be eventually combined with dimensionality reduction \citep{tan2011computational} and several rollouts can be performed firstly in simulation \citep{cohen2014integrating} to further speed up the policy search. When multiple demonstrations are given, one can learn a mapping between policy parameters and query points (\eg, goal positions) and use the mapping to generalize to new situations (Section~\ref{subsubsec:task_paramters}). This strategy was used by \cite{nemec2011exploiting, nemec2012applying, nemec2013efficient} to provide a good initial policy for a new situation which is then further refined using \ac{rl}. Being the mapping estimated using example query points, the search space can be effectively constrained within query points making the policy search more efficient. \cite{Vuga2015Enhanced,Vuga2015Speed} combined this approach with a different \ac{dmp} formulation to optimize the velocity of execution. The approach was tested on diverse tasks including pouring water in a cup, where it prevented the water to split from the cup during the motion. \cite{Schroecker2016Directing} provided demonstrations in the form of soft via-points (Section~\ref{subsubsec:via-points}) which reduce the search space to the neighborhood of the taught via-points. Multiple demonstrations were used by \citep{reinhart2014efficient, reinhart2015efficient} to build a parameterized skill memory that connects low-dimensional skill parameterization to motion primitive parameters. This low-dimensional embedding is then leveraged for efficient policy search. Instead of learning a mapping from task to policy parameters, \cite{QueiBer2016Incremental} used data from the rollouts to incrementally learn a parametric skill (bootstrapping) and used it to generate a good initial policy for a new task.

\subsubsection{\texorpdfstring{\acp{dmp}}{} generalization and sequencing}
Instead of using generalization to provide a better initial policy, some researchers exploit \ac{rl} to improve and generalize the motion primitive. \citep{Andre2015Adapting} adapted \ac{dmp} policies to walk on sloped terrains. \cite{Muelling10} generalized to new situations using a mixture of \acp{dmp}. In their approach, \ac{rl} was used to estimate the shape parameters as well as to estimate the optimal responsibility of each \ac{dmp}. \citep{mulling2013learning} used episodic \ac{rl} to estimate meta-parameters like the temporal and spacial interception point of the ball and the racket typical of table tennis tasks. \cite{Lundell2017Generalizing} used parameterized kernel weights and \ac{rl} to search for optimal parameters, while \citep{Forte2015Exploration} augmented the given demonstration using \ac{rl}-based state space exploration to autonomously expand the robot's task knowledge. Metric \ac{rl} was exploited by \citep{Hangl2015Reactive} to smoothly switch between learned \ac{dmp} policies and execute a task in new situations.

\ac{rl} can be also applied to sequence multiple motion primitives and perform more complex task; a successful strategy when the robot has to perform, for instance, a manipulation task (Section~\ref{subsec:manipilation}). To sequence multiple primitives it is also of importance to learn the goal of each motion. \cite{tamosiunaite2011learning} used continuous value function approximation to optimize  the  goal  parameters  of a \ac{dmp} used to perform a pouring task. 
\cite{kober2011reinfrocement, kober2012reinforcement} learned a meta-parameter function that maps the current state to a set of meta-parameters including goal and duration of the movement.  Instead of separating shape and goal learning into different processes, \citep{stulp2011learning2, stulp2012reinforcement} extended \ac{pi2} to simultaneously learn shape and goal of a sequence of \acp{dmp}. 

\subsubsection{Skills transfer}
Learned skills can be potentially transferred across different tasks to speed up the learning process and increase robot autonomy. To this end, \cite{fabisch2014active} considered the case where the robot can actively choose which task to learn to make the best progress in learning. The process of actively selecting the task was considered as a non-stationary bandit problem for which suitable algorithmic solution exist while intrinsic motivation heuristics were exploited to reward the agent after the selection. 
\citep{Cho2019} defined the complexity of a motor skill based on temporal and spatial entropy of multiple demonstrations and used the measured complexity to generate an order for learning and transferring motor skills. Their experimental findings provided useful guidelines for skill learning and transfer. In short, humans have to demonstrate, when possible, the most complex task and then the robot is able to transfer the motor skills. Vice versa, if demonstrations are not given, it is more effective to start learning simple skills first and then transfer the simpler skills to more complex tasks.  


\subsubsection{Learning hierarchical skills}
\label{susbsub:secRL:LHS}
\ac{rl} often lacks scalability to high dimensional continuous state and action spaces. To remedy this issue, hierarchical \ac{rl} exploits a \textit{divide et impera} approach by decomposing a \ac{rl} problem into a hierarchy of sub-tasks in order to reduce the search space. Different levels in the hierarchy represent information at different time and/or spatial scale. 

\cite{stulp2011hierarchical} proposed to represent different options as \acp{dmp} to sequence. \ac{pi2} was extended to optimize shape and (sub-)goal of each \ac{dmp} at different levels of temporal abstraction. In particular, the shape was adjusted based on the cost up to the next primitive in the sequence, while the sub-goal considers the cost of the entire sequence of two \acp{dmp}. Layered direct policy search in \citep{End2017Layered} did not rely on a set of predefined sub-policies and/or sub-goals, but instead used information theoretic principles to  uncover a set of diverse sub-policies and sub-goals.

Reducing the number of rollouts required to discover optimal policies is also important in \ac{hrl}. As already mentioned, \ac{il} is a valuable option to find good initial policies. However, there are applications like manipulation with multi-fingered robotics hands for which it is hard or impossible to provide expert demonstrations. To make policy search more efficient, \cite{OjerDeAndres2018} used \ac{hrl} where the upper-level considers discrete action and state spaces to search for optimal finger gaiting and synchronization among the fingers. This information was passed to the lower-level where rhythmic \acp{dmp} and \ac{pi2} generated continuous commands for the fingers. Another possibility to increase data-efficiency is to use model-based approaches for \ac{rl}. \cite{Colome2015Afriction} exploited a friction model to improve a \ac{dmp} policy and manipulate soft tissues (a scarf). A model-based \ac{hrl} approach was proposed by \citep{Kupcsik2017Model} for data-efficient learning of upper-level policies that generalize well across different executive contexts. Finally, \citep{Li2018} proposed a hybrid hierarchical framework where the higher-level computes optimal plans in Cartesian space and converts them to desired joint targets using an efficient solver. The lower-level is then responsible to learn joint space trajectories under uncertainties using \ac{rl} and \acp{dmp}.

\subsection{Deep learning} \label{subsec:deep_learning}

A popular method of machine learning are \acp{nn}. Due to their non-parametric nature, they can effectively represent nonlinear mappings. A major drawback of \acp{nn} in the past was their computational complexity of learning. In recent years there is a renewed interest in \acp{nn}. New deep learning approaches were successfully \citep{lecun2015deep} applied in machine vision and language processing.

In recent years, deep learning has been applied also in robotics to learn task dynamics \citep{yang2016repeatable} and movement dimensionality reduction \citep{chen2015efficient}. The authors \citep{chen2015efficient, chen2016dynamic} introduced a framework called \ac{aedmp} which uses deep auto-encoders to find a movements represented in latent feature space. In this space \acp{dmp} can optimally be generalized to new tasks, as well as the architecture enables the \acp{dmp} to be trained as a unit. \cite{Pervez2017} in their work coupled the vison perception data for object calcification with task specific movement definitions represented with \acp{dmp}. The data was modeled with \acp{cnn}, where the images and the associated movements were directly processed by the deep \ac{nn}, thus preserving the associated \acp{dmp} properties and eliminating the need for extracting the task parameters during motion reproduction. Later on \cite{Kim2018Learning} combined deep \ac{rl} with \acp{dmp} to learn and generalize robotic skills from demonstration. The framework builds on a \ac{rl} approach to learn and optimize a new \ac{dmp} skill based of a demonstration. The \ac{rl} approach is backed up with a hierarchical search strategy, reducing the search space for the robot, which allows for more efficient learning of complex tasks. Furthermore, \cite{Pan2018} presented an deep learning approach form motion planning of high dimensional deformable robots in complex environments. The locomotion skills are encoded with \acp{dmp} and a \ac{nn} is trained for obstacle avoidance and navigation. The data is further optimized with deep Q-Learning showing that the learned planner can efficiently plan and navigate tasks for high dimensional robots in real time. 

\cite{Pahic2018} proposed a deep learning approach for perception-action couplings, demonstrating the coupling between the vision based images and associated movement trajectories. Later on they extended the approach  to incorporate \acp{cnn} and give a distinguishing property formulation for the approach \citep{Pahic2020}, which utilizes a loss function to measure the physical distance between the movement trajectories as opposed to measuring the distance between the \acp{dmp} parameters which have no physical meaning, leading to better performance of the algorithm. Recently, they extended the usage of \ac{gpr} to create a database needed to train autoencoder \acp{nn} for dimensionality reduction \citep{loncarevi2021Generalization}.


\subsection{Lifelong/Incremental learning}
\label{subsec:lifelong}

\begin{figure}[t]
	\centering
	\includegraphics[width=\columnwidth]{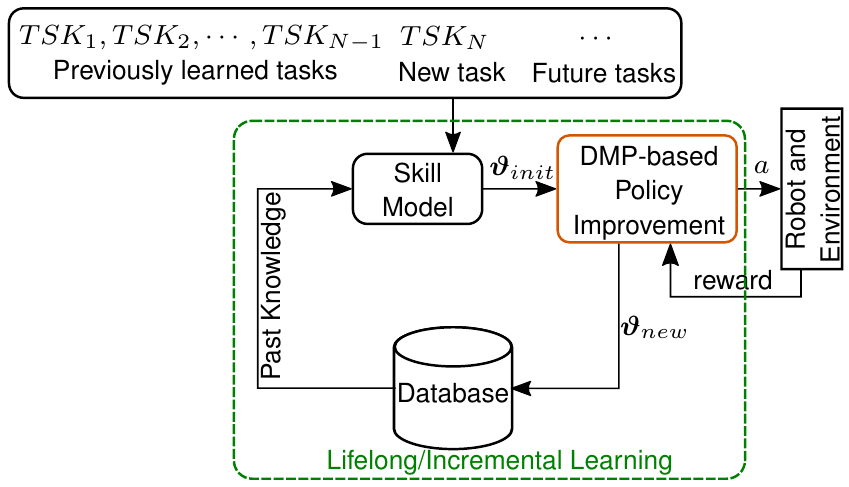}
	\caption{General framework of lifelong/incremental learning approach.}
	\label{fig:lifelong}
\end{figure}

Lifelong (incremental) learning is a framework which provides continuous learning of tasks arriving sequentially \citep{thrun1996learning, chen2018lifelong, fei2016learning}. 
The essential component of this framework is a database which maintains the knowledge acquired from previously learned tasks $TSK_1, TSK_2, \cdots, TSK_{N-1}$. 
Incremental learning starts from the task manager assigning a new task $TSK_{N}$ to a learning agent. 
In this case, the agent exploits the knowledge in the DB as prior data for enhancing the generalization performance of its model on the new task. 
After the new task $TSK_{N}$ is learned, database is updated with the knowledge obtained from learning $TSK_{N}$. 
In fact, the incremental learning framework provides an agent with three capabilities: (\emph{i}) continuous learning, (\emph{ii}) knowledge accumulation, and (\emph{iii}) re-using previous knowledge for future learning enhancements. Figure \ref{fig:lifelong} shows general structure of \ac{dmp} integrated in a lifelong framework.



\cite{churchill2014evolutionary} proposed a cognitive architecture capable of accumulating adaptations and skills over multiple tasks in a manner which allows recombination and re-use of task specific competences. \cite{Lemme2014Selfsupervised} segmented demonstrations based on geometric similarities, and subsequently created a motion primitives library. The library is updated by removing unused skills and including new ones. 
Multiple  demonstrations  are  used  by   \citep{reinhart2014efficient, reinhart2015efficient} to build a parameterized skill memory that connects  low-dimensional  skill  parameterization  to  motion primitive  parameters.  This  low-dimensional  embedding  is then leveraged for efficient policy search.
Piece-wise linear phase is used to improve incremental learning performance \citep{Samant2016Adaptive}. \cite{Duminy2017Strategic} designed a framework for learning which data collection strategy is most efficient for acquiring motor skills to achieve multiple outcomes, and generalize over its experience to achieve new outcomes for cumulative learning.

\begin{figure*}[t]
	\centering
	\includegraphics[width=\linewidth]{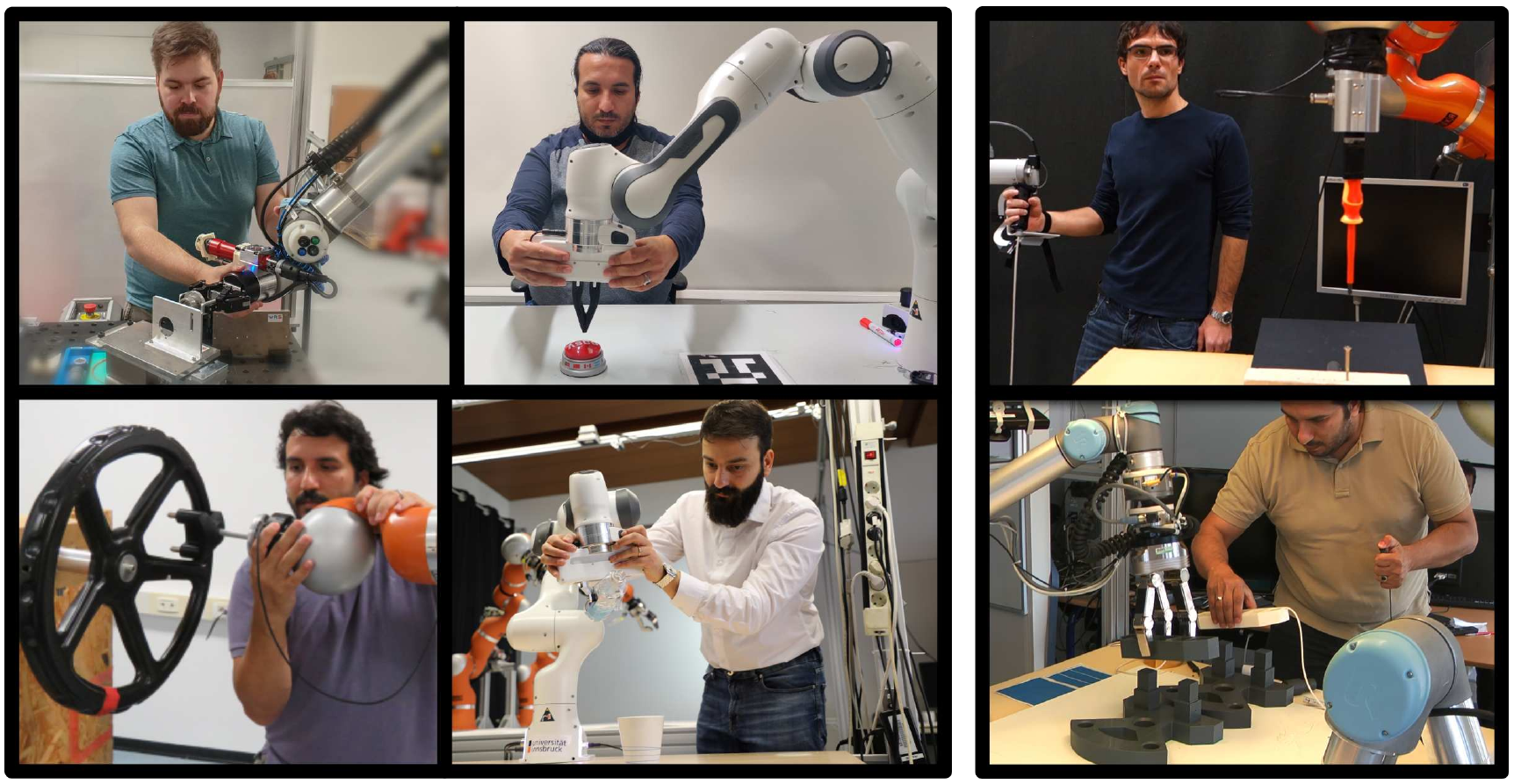}
	\caption{Human operators teach the robot how to perform different tasks. \emph{Left} scenarios use robots' gravity compensation mode to enable kinesthetic guiding, where a human operator guides the robot’s tool center point along the desired trajectory in such a way that the desired task is successfully executed \citep{Sloth2020499,AbuDakka2015Adaptation,abudakka2018force,caccavale2019kinesthetic}. \emph{Right} scenarios use teleoperation system to demonstrate appropriate robot movements either through haptic interface \citep{peternel2018robotic} or magnetic trackers \citep{AbuDakka2015Adaptation}.}
	\label{fig:lfd}
\end{figure*}

A generative learning framework is proposed to augment the robot's knowledge-base with missing information at different level of the cognitive architecture including symbolic planning as well as object and action properties \citep{Worgotter2015Structural}.

\cite{Wang2016Dynamic} proposed a modified formulation of \acp{dmp} called as \ac{dmp}+ which capable of efficiently modify learned trajectories by improving the usability of existing primitives and reducing user fatigue during \ac{il}. Later, \ac{dmp}+  had been integrated into a dialogue system with speech and ontology to learn or re-learn a task using natural interaction modalities \citep{Wu2018}.

In literature, it has been shown that incremental learning provides better generalization than the isolated learning approaches in terms of interpolation, extrapolation and the speed of learning \citep{hazara2017model}.
\cite{hazara2018speeding} improved their \ac{gpdmp} \citep{Lundell2017Generalizing} in order to construct, incrementally, a database of motion primitives, which aims to improve the generalization to new tasks. Furthermore,
it has been transferred incrementally from simulation to the
real world \citep{hazara2019transferring}. Moreover, authors endow incremental learning with a task manager, which capable of selecting a new task by maximizing future learning while considering the current task performance \citep{Hazara2019Active}.




%% file: DMP_Applications.tex
\section{\texorpdfstring{\acp{dmp}}{} in Application Scenarios}
\label{sec:applications}

We categorize the applications into several subsections based on different topics. We first separate the use of \acp{dmp} for robot interaction with the \textit{passive environment} (\eg tools, objects, surfaces, \etc) and for interaction with an agent that involve \textit{co-manipulation} (\eg human, another robot, \etc). Additionally, we examine several other major application areas, such as \textit{human body augmentation/rehabilitation} with exoskeletons, \textit{teleoperation}, \textit{motion analysis/recognition}, \textit{high \ac{dof} robots}, and \textit{autonomous driving and field robotics}.


\subsection{Robots in contact with passive environment}\label{subsec:passive_environment}
Most of the daily tasks that the robots perform involve some kind of physical interaction with the environment that requires control of forces or positions. Nevertheless, simultaneous control of force and position in the same axis is not possible \citep{stramigioli2001}\footnote{There is a duality in impedance-admittance, \ie, the force produce motion and motion produces force, therefore if one is the input, the other can only be the output of the control system \cite{peternel2017method}.}, and therefore the control approaches have to make a compromise between prioritizing position control or force control \citep{schindlbeck2015unified}. The key to such control is for the robot to learn appropriate force or position reference trajectories that can lead to the desired task performance in interaction with the environment.

\subsubsection{Demonstration of interaction tasks}
A common approach to teaching robot motion trajectories is kinesthetic guidance (\figref{fig:lfd}-\emph{Left}), where the human operator holds the robot arm and shows the appropriate movements to be encoded by \acp{dmp} \citep{kormushev2011imitation,AbuDakka2015Adaptation,joshi2017robotic,Papageorgiou2020,Papageorgiou2020Apassive}. Recently, the technology is protruding into high risk fields such as invasive surgery, where high-dimensional fine human-like manipulation skills are being demonstrated \citep{su2021} and executed with robots \citep{Su20202203,Ginesi201937}. In \citep{kormushev2011imitation}, the human held the robot arm and used kinesthetic guidance to teach the position and orientation trajectories necessary to perform ironing and door opening task. In the second stage the corresponding forces and torques were recorded with a haptic device in a teleoperation setup. For setups where the robot arm is equipped with multiple force/torque senors, the two demonstration steps with additional control policies can be combined into one \citep{Steinmetz2015Simultaneous, Montebelli2015Onhanding}. 

An alternative to learning force trajectories is to learn the impedance of the robot by learning the desired stiffness trajectories. The ability to change the impedance of the arm is crucial to simplify the physical interaction in unpredictable and unstructured environments \citep{hogan1984adaptive,burdet2001central}. In \citep{peternel2015human,peternel2018robotic} teleoperation was used with a push-button interface to command the robot impedance, which was learned by \acp{dmp} that enabled the robot to perform various collaborative assembly tasks. For example, the learned position and stiffness \acp{dmp} were used to insert a peg in a groove to bind the two parts \citep{peternel2015human}, or to screw a bolt \citep{peternel2018robotic}. A similar approach was used in \citep{yang2018dmps} to learn \acp{dmp} used for vegetable cutting task.

\begin{figure}[!t]
	\centering
	\includegraphics[width=\columnwidth]{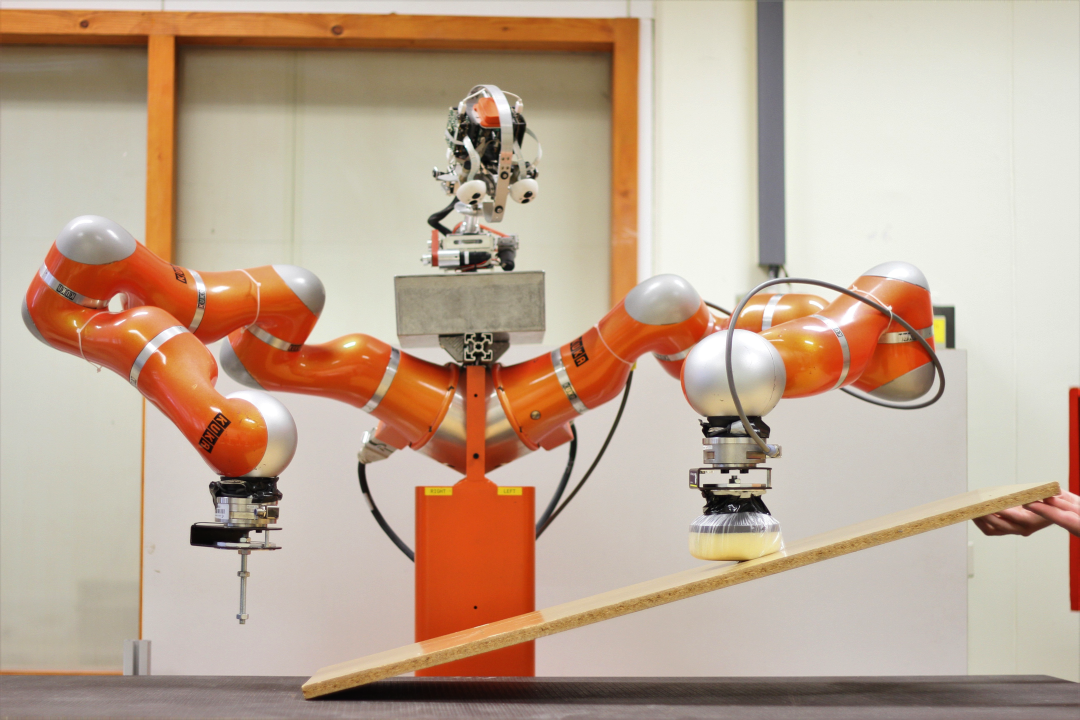}
	\caption{Using \acp{dmp} for adapting to changing surfaces (\eg wiping task) \citep{kramberger2018passivity}}
	\label{fig:wiping}
\end{figure}

While teleoperation based methods are very effective to teach the robot \acp{dmp} for interaction tasks, it usually involves a complex and expensive system. The method in \citep{abudakka2018force} enabled the robot to learn stiffness profiles through measurement of interaction force with the environment to perform valve turning task. The method in \cite{peternel2017method} used human demonstration and \ac{emg} to learn stiffness \acp{dmp} from human muscle activity measurements in order to perform sawing and wiping (\figref{fig:wiping}) tasks.

Nevertheless, adaptation of a single trajectory is unlikely to generate an appropriate solution for more general cases, where the task execution needs to change significantly. After learning the initial \ac{dmp} motion trajectories through kinesthetic guidance, the robot can then adapt them based on the measured force of interaction while performing the task. \cite{Pastor2011Online} introduced a method for real-time adaptation of demonstrated \acp{dmp} trajectories depending on the measured sensory data. They developed an adaptive regulator for trajectory adaptation based on estimated and actual force data. Recently, \cite{Prakash2020}, extended the real-time adaptation approach incorporating a fuzzy fractional order sliding mode controller in order to efficiently and stably adapt the demonstrated \ac{dmp} trajectory to fast movements, such as a ping pong swing.

\cite{Sutanto2018} presented a data-driven framework for learning a feedback model from demonstrations. They used an \ac{rbfnn} to represent the feedback model for the movement primitive. Similarly to this research, \cite{gams2010line} proposed a method for adaptation of demonstrated movements depending on the desired force, with which the robot should act on the environment. Thus, they ensured the adaptation of the learned movements to different surfaces. This approach  was later expanded \citep{Pastor2011Online} to provide the statistically most likely force-torque profile \citep{pastor2012towards} and furthermore, force-torque data was used for training a classifier \citep{Straizys2020} in order to modulate the demonstrated trajectory for the use with delicate tasks such as tissue or fruit cutting.

Moving onward form policy learning, \cite{do2014learn} presented an adaptation framework, where not only the desired adaptation force or trajectory, but the entire skill can be learned. They demonstrated the method with a wiping task under different environmental conditions. 

\subsubsection{Assembly tasks}
\begin{figure}[!t]
	\centering
	\includegraphics[width=\columnwidth]{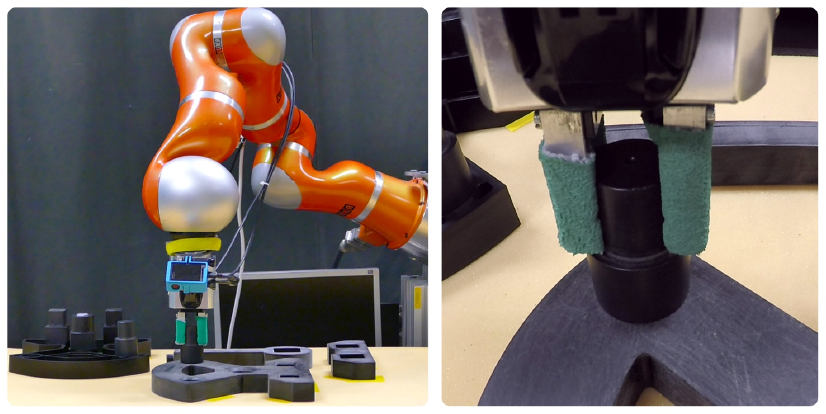}
	\caption{An example of using \acp{dmp} in assembly tasks (\eg peg-in-the-hole) \citep{kramberger2016learning}}.
	\label{fig:pih}
\end{figure}

Assembly presents one of the more challenging tasks to automate, where not only position trajectories but also task dynamics have to be taken into account. To deal with this challenge, various methods were proposed. \cite{AbuDakka2015Adaptation} proposed a method that can learn the orientation aspect of the complex physical interaction, like the peg-in-the-hole assembly tasks (\figref{fig:pih}). The proposed method was integrated in an industrial assembly framework where the key challenge was to adapt to uncertainties presented by the assembly task \citep{kruger2014technologies, abudakka2014solving}.

Complex assembly tasks that are subject to change cannot be demonstrated and executed on the fly therefore, adaptation methods are required for ensuring a successful execution. \cite{Nemec20206521} used exception strategies for dealing with complex assembly cases . \cite{Sloth2020499} presented an exception strategy framework, combining discrete and periodic \ac{dmp}, coupled with force control to learn an assembly task under tight tolerances. \cite{Gaspar2020346} presented several industrial assembly challenges and focused on fast and efficient setup of industrial tasks with the emphasis on \ac{lfd}. \cite{Angelov20202658} incorporated several different control policies by taking into account the dynamics and sequencing of the task. 

In some cases, active exploration and autonomous database expansion can be used for learning assembly policies automatically. In \citep{Petric2015Bioinspired} the proposed algorithm can build and combine \ac{cmp} motion knowledge from a database in an autonomous manner.

Complementary to assembly tasks, disassembly is also challenging by solely using the demonstrated trajectories. As described in \citep{Ijspeert2013Dynamical}, \acp{dmp} have a unique point attractor in the specified goal parameter of the movement, essentially repelling the idea of reversibility. Therefore, \cite{nemec2018anefficient} proposed a framework, where the disassembly challenge was tackled by learning two separate \acp{dmp} from a single demonstrated motion; one forwards and one backwards. \cite{san2019towards} took the idea further and reformulated the \acp{dmp} phase system with a logistic differential equation to obtain two stable point attractors. This approach provided a reversibility formulation of the dynamical system and demonstrated the effectiveness of the algorithm on a peg-in-hole assembly task.

\subsubsection{Learning methods for contact adaptation}
Desired force-torque profiles can be tracked using \ac{ilc} \citep{Gams2014,Gams2015Accelerating}. In repetitive robotic tasks, iterative learning has been gaining increased popularity \citep{bristow2006survey} due to its effectiveness and robustness. However, in order to achieve effective results, a careful tuning of learning parameters is required. \cite{norrlof2002adaptive} and \cite{tayebi2004adaptive} presented an adaptive learning approach for automated tuning of learning parameters.

Another approach is to use \ac{rl} to adapt \acp{dmp}. For example, in \citep{buchli2010variable,buchli2011learning} stiffness parameters were adjusted during the task execution by \ac{rl}.

Alternatives to feedback-based adaptation of \acp{dmp} and \ac{rl} are scalability and generalization approaches. \cite{Matsubara2011} proposed an algorithm for the generation of new control policies from existing knowledge, thereby achieving an extended scalability of \acp{dmp}, while mixture of motor primitives were used for generation of table tennis swings \citep{Muelling10}. On the other hand, generalization of \acp{dmp} was combined with model predictive control by \cite{Krug2015} or applied to \ac{dmp} coupling terms by \citep{Gams15}, which were learned and later added to a demonstrated trajectory to generate new joint space trajectories.

\cite{Stulp13_hum} proposed to learn a function approximator with one regression in the full space of phase and tasks parameters, bypassing the need for two consecutive regressions. \cite{forte2012line} performed a comparison study of \ac{lwr} and \ac{gpr} for trajectory generalization. This work shows that higher accuracy can be achieved with \ac{lwr} trajectory approximation. \cite{Koropouli2015} presented a generalization approach for force control policies. By learning both the policy and the policy difference data using \ac{lwr}, they could estimate the policy at new inputs through superposition of the training data.

\citep{Denisa2016} used \ac{gpr} based generalization over combined joint position trajectories and torque commands in the framework of \acp{cmp}. To showcase the versatility of the approach, \citep{Petric2018} applied it for robot based assembly tasks. Finally, \cite{kramberger2017generalization} extended the approach to account for variations of the desired tasks, \eg assembly of similar objects. This enables the robot movements to be automatically generated with the use of \ac{lwr} from a demonstrated database of successful task executions, which include kinematic and dynamic demonstrated trajectories encoded with \acp{dmp}. The newly obtained data is used to account for the changes in the work-space. Nevertheless, a major problem in statistical learning is how to efficiently deal with singularity free representations of orientation trajectories. To resolve this issue, \cite{kramberger2016generalization} proposed a formulation for Cartesian space \acp{dmp} where orientations are represented with unit quaternion.

\subsection{Human--robot co-manipulation}

\begin{figure}[!t]
    \centering
    \includegraphics[width=\columnwidth]{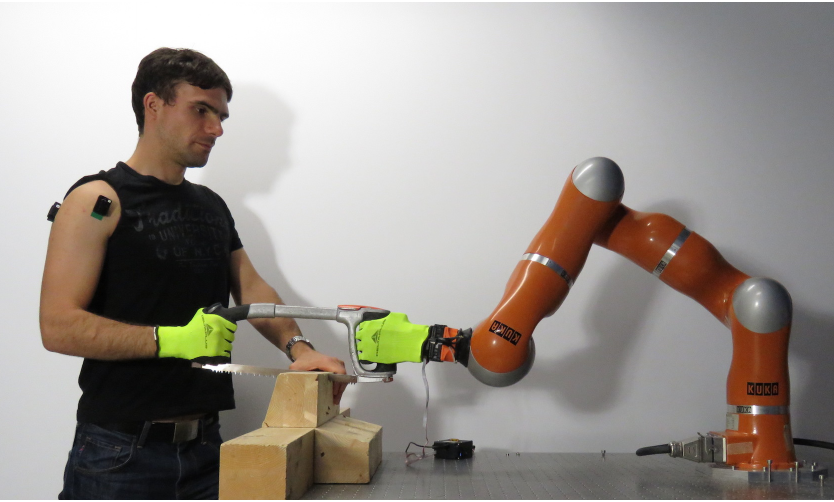}
    \caption{An example of using \acp{dmp} for collaborative human--robot sawing from \citep{peternel2018robot}.}
    \label{fig:sawing}
\end{figure}

While control of robot interaction with the passive environment can solve majority of the tasks, in some cases the robot needs to interact with an active agent (\eg human, another robot, etc.). Human-robot collaboration is becoming one of the key fields in robotics \citep{ajoudani2018progress}. To perform a successful physical human-robot collaboration, the robot must be able to control complex movements in coordination with the human partner. In this direction, the ability to modulate the impedance is important to coordinate the physical interaction during human-robot co-manipulation of tools \citep{peternel2017human}. \acp{dmp} offer an elegant solution to encode such coordinated dynamic movements.

In \citep{peternel2014teaching} the collaborative robot was thought online through teleoperation how to perform collaborative sawing with a human co-worker. The impedance was commanded to the robot through muscle activity measurement using \ac{emg}. \acp{dmp} were used to encode coordinated phase-dependent motion and impedance as demonstrated by the human teleoperator. Teaching though teleoperation is an effective way to convey the physical interaction skill to the collaborative robot, however the setup can be expensive and is not widely available.

An intuitive alternative to teleperation is for the robot to learn the skill directly though physical interaction with the human partner while they are collaborating. Numerous methods have focused on learning the synchronized motion between collaborative partners \citep{kulvicius2013interaction,prada2013dynamic,Gams2014,umlauft2014dynamic,zhou2016coordinate,peternel2018robot,sidiropoulos2019human,ugur2020compliant}. For example, in \citep{kulvicius2013interaction} the interactive movements were encoded with \acp{dmp} and adapted them based on the measured force arising from the disagreements between agents during co-manipulation. Similarly, in \citep{Gams2014} the collaborative movements were encoded with \acp{dmp} and adapted using force feedback and \ac{ilc}. The approach in \citep{zhou2016coordinate} combines two \acp{dmp} to encode the movements of each partner's arm, which are coupled in a leader-follower manner.

Besides adapting the collaborative movements, in \citep{peternel2016adaptation,peternel2018robot} the robot used \acp{dmp} to also learn the impedance online directly from the co-manipulation with the human (\figref{fig:sawing}). The robot started with a basic skill set that enabled it to collaborate with the human in a pure follower role. Thorough the collaborative task execution the robot then learned the motion and impedance trajectories online and encoded them with \acp{dmp}. When the human became fatigued, the robot used the learned advanced skill to take over majority of the task execution.

The method in \citep{BenAmor2014} proposed an upgraded version of standard \acp{dmp} called \textit{Interaction Primitives} that can account for a probabilistic nature of collaborative movements. Rather than having a single value of weights, the \ac{dmp} includes weight distributions. This distribution enabled the robot to learn the inherent correlations of cooperative actions and infer the behavior of the human partner during the cooperation. \citep{Cui2016Environment, cui2019environment} used visual information to extract context related parameters that augment the interaction primitives to increase the robustness during the task execution. 

There are also other types of co-manipulation scenarios, such within-hand bi-manipulation or human-robot object handover. For example, in \citep{koene2014experimental,Gao2019} \acp{dmp} were used to perform bi-manipulation, while in \citep{prada2014implementation,Solak20198246,Lafleche2019,Abdelrahman2020} \acp{dmp} were used for human-robot object handover.

When the environment is hazardous for the human workers or when there are too many robots compared to the number of human workers, the obvious solution is to make robot collaborate between themselves. The method in \citep{peternel2017robots} used \acp{dmp} to make novice robots learn from the expert robot through co-manipulation. Initially the novice robot remained compliant to let the expert robot lead the task execution. In the first stage, the novice robot learned the reference motion through \acp{dmp}. In the second stage, it became stiff to perform the newly learned motion, while the expert robot initiated stiff/compliant phases expected in the collaborative task execution. Finally, the novice robot then learned in which phases of the task to increase or decrease the impedance and encoded this impedance behavior with \acp{dmp}.

\subsection{Human assistance, augmentation, and rehabilitation}

\begin{figure}[t]
	\centering
	\includegraphics[width=.9\columnwidth]{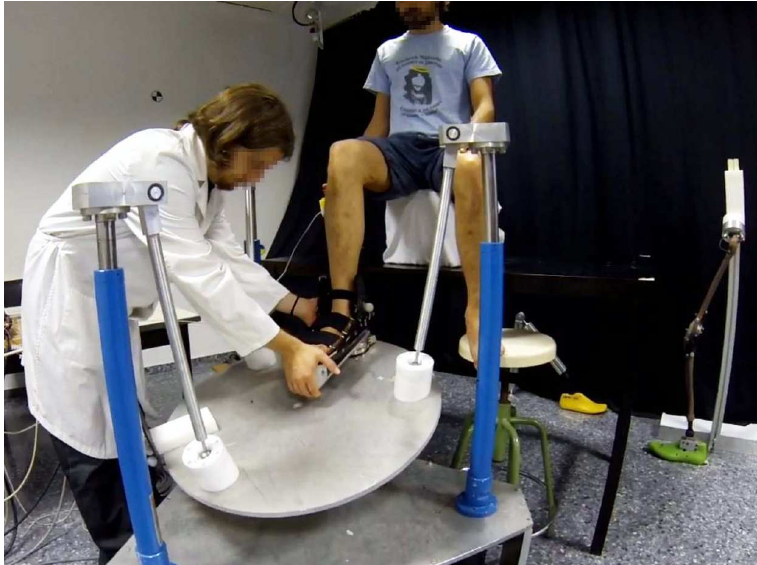}
	\caption{An example of using \acp{dmp} for teaching passive exercises for ankle rehabilitation \citep{abudakka2015Trajectory, abudakka2020passive}.}
	\label{fig:ankle}
\end{figure}

The most common type of co-manipulation is the classic human-robot collaboration, where a human and a robotic agent are physically performing industrial or daily tasks. Another type of co-manipulation occurs when a human is wearing an exoskeleton. In most cases, the exoskeleton simply amplifies the current human motion \citep{kong2006design,fleischer2008human}. However, in some cases we want the exoskeleton to execute pre-defined trajectories in order to perform a physical therapy on patients, or to completely offload a repetitive motion of healthy human workers.

\begin{figure}[!t]
    \centering
    \includegraphics[width=\columnwidth]{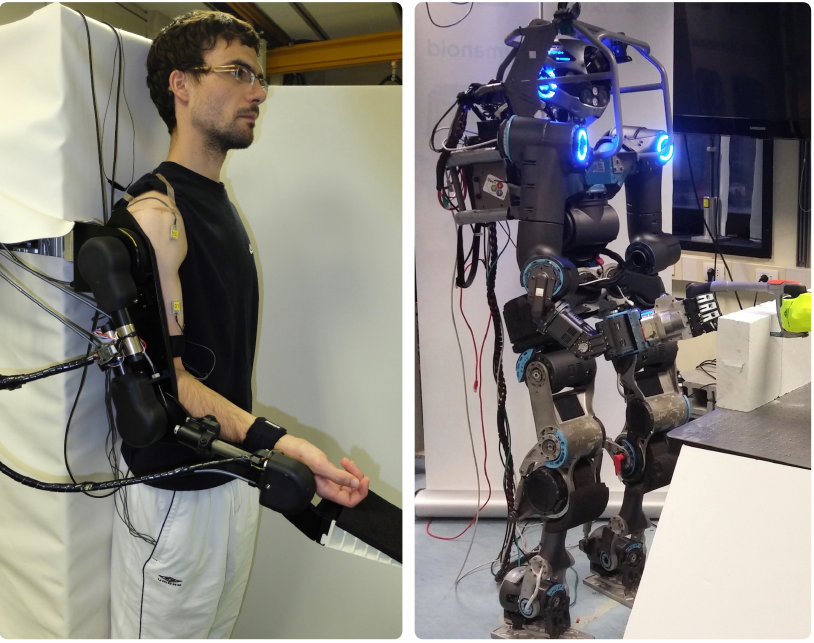}
    \caption{The left photo shows arm exoskeleton application from \citep{peternel2016adaptive}. The right photo shows high-DoF humanoid robot Walk-man \citep{tsagarakis2017walk} performing sawing in \citep{peternel2017robots}.}
    \label{fig:exoskeleton_humanoid}
\end{figure}

The methods in \citep{Lauretti2017Learning,lauretti2018learning} obtained \acp{dmp} in offline by learning by demonstration, which were then used by an arm exoskeleton to support human movements. In \citep{peternel2016adaptive} the control method employed \acp{dmp} to interactively adapt the joint torques required to perform the arm exoskeleton movements and compensate all the underlying dynamics (\figref{fig:exoskeleton_humanoid}-\emph{Left}). The phase-dependent toque trajectory updated online in order to minimise the muscle activity feedback measured by \ac{emg}. In \citep{Petric2016Cooperative} the robot encoded the assistive motion with \acp{dmp} and then adapted it by taking into account aspects of human motor control through the Fitts’ law. 

Gait related rehabilitation with exoskeletons is a very common application of \acp{dmp} and there are numerous examples \citep{abudakka2015Trajectory,huang2016hierarchical,huang2016learning,Hwang20191905,Yuan20203830,Amatya2020}. In \citep{abudakka2015Trajectory, abudakka2020passive} a parallel robot was used for ankle rehabilitation, where the movements were generated by \acp{dmp} (\figref{fig:ankle}). In \citep{huang2016hierarchical} \acp{dmp} were used to learn the gait motion trajectories for a lower-body exoskeleton. This approach was then extended with a \ac{rl} method to adapt a force coupling term (similar to earlier approaches presented in \secref{subsubsec:Forcecoupling}) to enable online adaption of motion trajectories \citep{huang2016learning}.

Besides normal gait, \acp{dmp} were also applied for stair-ascend \citep{Xu2020} and sit-to-stand \citep{kamali2016trajectory} assistive movements of lower-body exoskeletons. In \citep{Joshi20191156}, a robotic arm was used to assist humans with putting the cloths on their body, where the movements were generated by \acp{dmp}.

Besides assistive body movement and rehabilitation, \acp{dmp} were also applied for relaxation purposes. For example, in \citep{Li2020} a robotic arm provided massage movement through \acp{dmp}.

\subsection{Teleoperation}

Teleoperation is one of the major fields of robotics and enables a human to have a direct and real-time control over a (remote) robot. Typically the control is done through interfaces that can capture the human commands to be sent to the robot and that can provide haptic feedback from the robot. While teleoperation focuses on giving the human operator a full or shared control over the robot, \acp{dmp} are used to encode autonomous robot behaviors. Therefore, here we mostly examine cases where teleoperation is used to teach the robot new autonomous behavior encoded by \acp{dmp}.

In \citep{kormushev2011imitation} a combination of kinesthetic teaching and teleoperation was employed to form the \ac{dmp}-based robot skill for ironing. After the motion trajectories were learned through kinesthetic guidance, the corresponding forces were recorded by using haptic device and a teleoperation system. In \citep{peternel2014teaching} teleoperation was used to teach the robot how to physically collaborate with another human. Since there was no haptic feedback, the teleoepration setup was unilateral, but the human was able teach also the impedance of the robot in addition to motion. The former was commanded by muscle activity measurement through \ac{emg}, while the latter was was commanded by the movement of the human operator's arm as measured by an optical motion capture system.

In \citep{peternel2018robotic} the human operator thought the robot through teleoperation how to perform autonomous assembly actions (\figref{fig:lfd}-\emph{Right}). \acp{dmp} were used to encode the commanded impedance and motion, however a more practical push-button based impedance command interface was employed. More importantly, the teleoperation setup was bilateral and the haptic interface provided the human operator the feedback about the forces the robot felt. Similarly, teleoperation approaches were used in \citep{yang2018dmps,Lentini2020}.

Real robot is not always necessary to acquire new skills. In \citep{Beik-Mohammadi2020} the robot and the environment were simulated and the human operator used a virtual reality system. A combination of \acp{dmp} and \ac{rl} was used to form an adaptive skill. The scenario proposed in \citep{AbuDakka2015Adaptation} was teleoperation in its basis, however the human demonstrator did not just pretend that he/she is embodied in the robot, but the robot task environment was cloned at the human side (\figref{fig:lfd}-\emph{Right}). This removed the need for force feedback and haptic device, since the human felt the real environment on his/her side, while the motion was captured by non-contact based sensory system (\ie magnetic trackers) and then mirrored on the robot.

Multiple demonstrations through teleoperation can be inconsistent, especially if done in a multi-agent shared-control setting. The method proposed in \citep{Pervez20192055} can synchronize inconsistent demonstration through shared-control teleoperation and encode them with \acp{dmp}.

\subsection{High \texorpdfstring{\ac{dof}}{} robots}
\acp{dmp} provide an elegant and fast way to deal with systems with high-dimensional space by sharing one canonical system \eqref{eq:dmp_discrete3} among all \acp{dof} and maintain only a separate set of transformation systems. By high-dimensional space we are referring to systems with 10 or more \acp{dof} (\ie Walk-man humanoid robot in \Figref{fig:exoskeleton_humanoid}-\emph{Right}). In this section, we will quickly mention some of the potential works with high number of \acp{dof}.

\cite{Ijspeert2002Learning, ijspeert2002learningAttractor} used \acp{dmp} in an \ac{il} framework to learn tennis forehand, a tennis backhand, and rhythmic drumming using 30-\acp{dof} humanoid robot. \cite{Pastor2009} used \acp{dmp} to encode a 10-\acp{dof} exoskeleton robot arm. \cite{Luo2015Learning} integrated \acp{dmp} with  stochastic policy gradient \ac{rl} and \ac{gpr} in order to design an  online adaptive push recovery control strategy. The approach had been applied to PKU-HR5 humanoid robot with 20-\acp{dof}. \cite{Andre2015Adapting, Andre2016Skill} implemented a predictive model of sensor traces that enables early
failure detection for humanoids based on an associative skill memory  to periodic movements and \acp{dmp}. They applied their algorithm on DARwIn-OP with 20-\acp{dof} in simulation. \cite{pfeiffer2015gesture} represented gestures by applying \acp{dmp} on REEM robotic platform with 23-\acp{dof}. \cite{Nah2020} proposed an approach to optimize \ac{dmp} parameters in order to deal with the complexity of of high \ac{dof} system like a whip. They tested their approach in simulation for 10-, 15-, 20-, and 25-\acp{dof} systems. In order to reduce the number of required rollouts for adaptation to new task conditions, \cite{Queiber2018bootstrapping} used \ac{cmaes} to optimize \acp{dmp} parameters. In addition, they introduced a hybrid optimization method that combines a fast coarse optimization on a manifold of policy parameters with a fine grained parameter search in the unrestricted space of actions. The approach was successfully illustrated in simulation using a 10-\acp{dof} robot arm. \cite{Liu202054652} proposed \ac{dmp}-based trajectory generation to enable a full-body humanoid robot with 10-\acp{dof} (for the two legs) to realize adaptive walking.

\cite{Travers2016Shape, Travers2018Shape} proposed a framework that integrates \ac{dmp} with Gaussian-shaped spatial activation windows in order to plan the motion for high \ac{dof} robotic systems (\eg snake-like robot) in complex environment (with obstacles) by linking low-level controllers to high-level planners.

\subsection{Motion analysis and recognition}
\acp{dmp} tend to fit topologically similar trajectories with similar shape parameters $w_i$ \citep{Ijspeert2013Dynamical}. This behavior, due to the temporal and spatial invariance of \acp{dmp}, makes the shape parameters a useful descriptor to recognize similar motions. Indeed, \citep{strachan2004dynamic} have shown that the shape parameters computed for $5$ repetitions of $4$ classes of discrete hand gestures---measured with a $3$ \acp{dof} accelerometer---are linearly separable, \ie easy to classify. \cite{lantz2004rhytmic} draw similar conclusions for $10$ classes of periodic hand gestures. \cite{xu2005internal} used the correlation between the parameter vectors of two \acp{dmp} to measure the similarity between the original motion and recognize gait patterns. Similarly, \citep{Ijspeert2013Dynamical} used the correlation between parameter vectors to recognize the $26$ letters of the Graffiti alphabet.

The shape parameters $w_i$ are also suitable to fit more sophisticated classifiers like support vector machines. This strategy was used to successfully classify gestures observed with a monocular \cite{liu2014visual} or a binocular \citep{Wang2015Astudy} camera. Instead of considering a fixed number of basis function (number of shape parameters), \citep{Zhang2017Robust} used fast dynamic time warping \citep{salvador2007toward} to align parameter vectors of different length and then used $K$-nearest neighbors to classify different motions. 

Motion recognition can also be used to determine whether the robot is correctly executing a task by comparing sensed data with a movement template. In this respect, \citep{Andre2016Skill} used an associative skill memory, like the one in \citep{Pastor2011skill}, as a predictive model of sensor traces that enables early failure detection. In this work, \acp{dmp} were used to compactly encode the associative skill memory and speed up the failure detection.
Described approaches demonstrate that \acp{dmp} are a valuable option for gesture recognition especially for systems with limited computational power.

Humans tend to perform the same task in slightly different manners. Sometimes differences in the execution style contain useful information to adapt the motion to difference executive context. This is the case, for instance, of a reaching motion with and without an obstacle on the way. To capture the execution style \citep{matsubara2010learning} augmented the forcing term of the \ac{dmp} with a style parameter learned from multiple demonstrations. At run time, different style parameters can be used to smoothly interpolate between demonstrated behaviors. \cite{zhao2014generating} employed movements with different styles, but also learned a smooth mapping between style parameters and goal to improve the generalization.

When humans provide seamless demonstrations, \acp{dmp} can be used for online segmentation and recognition. To this end, \citep{Meier2011movement} assumed that a library of \acp{dmp} is given and used it to recognize motion segments during a task demonstration. Instead of using exemplar templates for each class of primitives, \citep{chang2013motion} segmented a video stream using motion to non-motion transitions, fitted \acp{dmp} on segmented data, and performed clustering to group similar motion segments in an unsupervised fashion. \cite{Song2020325} performed unsupervised trajectory segmentation using the concept of key points, \ie shared features across different task demonstrations. \citep{mandery2016using} segmented whole-body motions by detecting contacts with the environment and used them to build a  probabilistic language  model where words represent the poses and sentences sequences of poses. The learned language model was used to plan whole-body motion trajectories executed by joining multiple \acp{dmp} (see \secref{subsec:joiningDMPs}).

\acp{dmp} have been developed as a computational model of the neurobiological motor primitives \citep{schaal2007dynamics}. Experimental findings from neurophisiology related to the spinal force fields in frog have inspired the modification of \acp{dmp} formulation in \citep{hoffmann2009biologically}. As discussed in Section~\ref{subsubsec:start_goal}, this multidimensional representation overcomes limitations of classical \acp{dmp} like trajectory overshooting and dependence of the trajectory from the reference frame used to describe the motion. \cite{hoffmann2009biologically} also derived a collision avoidance strategy for \acp{dmp}, inspired by the way human avoid collisions during arm motion. 
\citep{dewolf2016spiking} investigated the human ability to cope with to changes in the arm dynamics and kinematic structure during motion control. They proposed a spiking neuron model of the motor control system that uses \acp{dmp} to implement the preparation and planning functionalities of the premotor cortex.
The effects of changes in the robot's dynamic parameters on the tracking performance of a \ac{dmp} trajectory were studied in  \citep{kuppuswamy2011impact}. Their findings suggests that the change in the body parameters should be explicitly considered in the \ac{dmp} learning process. 
\cite{Hotson2016High} augmented a brain-machine interface that captures neural signals with a \ac{dmp} model of the endpoint trajectories executed by a non-human primate. The system was used to decode real trajectories form a primate manipulating four different objects.

\subsection{Autonomous driving and field robotics}
\acp{dmp} can be utilized in various autonomous non stationary fields of robotics. \cite{perk2006motion} utilized \acp{dmp} for defining flight paths and obstacle avoidance for \acp{uav}, where the trajectories were generated based on the joystick movements controlling the throttle of the \ac{uav} motors. Later, \citep{fang2014control} extended the approach to encode user demonstrated \ac{uav} data, extracting and encoding the rhythmic and linear segments of the flight trajectory, and combining them into a flight control skill. Furthermore, \citep{tomic2014learning} formulated the \ac{uav} movements as a optimal control problem. The output of the optimal control solver was encoded with \acp{dmp}, enabling them to generalize and apply in-flight modifications to the \ac{uav} flight trajectories in real-time. Similarly, \cite{Lee2018,Kim2018Cooperation} presented a framework for \ac{uav} cooperative areal manipulation tasks, based on an adaptive controller which adapts the movement of the \ac{uav} in relation to the mass and inertial properties of the payload. In addition, \acp{dmp} were incorporated in the control scheme to modify the flight trajectories and avoid obstacles on the fly. The approach was later extended to incorporate path optimization, where \acp{dmp} play a significant tole for real time obstacle avoidance \citep{Lee2020135406}.    

As mentioned before, \acp{dmp} represent a versatile movement representation, which can be  implemented in various tasks and scenarios. One of the recent applications in this field are also \acp{auv}. \cite{Carrera2015Cognitive} integrated the \acp{dmp} in a learning by demonstration scenario for an \ac{auv}. The demonstrated data consisted of the manipulator and vehicle sensory outputs, which were efficiently used to demonstrate an underwater valve turning task.    

\acp{dmp} are also represented in the autonomous driving domain. In the recent work of \citep{Wang2018,wang2019motion}, the authors propose a framework which decomposes the complex driving data into a more elementary composition of driving skills represented as motion primitives. In the proposed framework, \acp{dmp} are utilized to represent the driver's trajectory with acceptable accuracy and can be generalized to different situations.

%% file: DMP_Discussion.tex
\section{Discussion}\label{sec:discussion}
This section provides guidelines to choose, among the several discussed in this work, the most appropriate approach for a given application. A useful criterion to decide whether to use a particular approach is the availability of code that greatly simplifies the implementation. We have searched for open-source \ac{dmp} implementations and listed them in a Git repository (see \secref{sec:code}). To further contribute the community, we have also released the implementations listed in \tabref{tab:source_code}. This section ends with a discussion on the limitations inherent to the \ac{dmp} formulation, the open issues, and the possible research directions. These are summarized in \tabref{tab:conclusion}.

\subsection{Guidelines for different applications}
Previous sections present different \ac{dmp} formulations and extensions together with possible application scenarios. As usual, there is not a single formulation that serves all the scopes and purposes, and the suitable approach to use depends on the goal to achieve and conditions of application. For this reason, we present some guidelines to guide the user in the process of selecting the formulation to use. 

\subsubsection{Discrete vs periodic}
For a task with distinct starting and ending points, discrete \acp{dmp} are a logical option to encode the movement trajectories between them. Examples of these tasks include: reaching and  pick-and-place, \citep{stulp2009compact2,forte2012line,Denisa2016,caccavale2019kinesthetic}, specific actions of assembly \citep{kruger2014technologies, abudakka2014solving,Gaspar2020346,Nemec20206521,Angelov20202658} and cutting \citep{yang2018dmps,Straizys2020}.

When the starting and ending points coincide, periodic \acp{dmp} are the logical option, since the encoded movements can be repeated over and over again. Good examples of their application are repetitive tasks such as locomotion \citep{ruckert2013learned,Wensing2017Sparse}, human body augmentation/rehabilitation \citep{peternel2016adaptive}, wiping a surface \citep{gams2016adaptation,peternel2017method,kramberger2018passivity} and sawing \citep{peternel2018robot}. Nevertheless, even typically non-repetitive tasks that are executed just once every now and then can still be encoded with periodic \acp{dmp} when the starting and ending points coincide \citep{peternel2018robotic}.

There are cases where it is not possible to clearly distinguish if the motion is periodic or discrete. For instance, \citep{ernesti2012encoding} have shown that the first step in a gait of a humanoid robot is a transients towards a periodic motion. Their representation is a good candidate to encode transients converging to a limit cycle trajectories. Finally, in some cases like in complex assembly, the task requires a combination of discrete and periodic \acp{dmp} \citep{Sloth2020499}.

\subsubsection{Space representation}
The original formulation of \acp{dmp} were and are still successfully applied to multidimensional independent data with each \ac{dof} $\in \mathbb{R}$ (\secref{subsec:discrete:classical} and \ref{subsec:periodic:classical}). These data can be joint or Cartesian positions, forces, torques, \etc, where every \ac{dof} of the data can be evolved independently form the rest. However, such formulation is not enough to successfully encode data with specific geometry constraints without pre- and/or post-processing the data. Examples of such data are: \emph{i}) orientation, where data are tight up by additional constraints (\ie the orthogonality in case of rotation matrix representation or the unit norm of the quaternion representation); \emph{ii}) full stiffness/damping matrices and manipulability matrices are encapsulated in an \ac{spd} matrices.

In many early works, orientation trajectories were learned and adapted without considering its geometry constraints \citep{Pastor2009}, leading to improper orientation and hence requiring an additional re-normalization. In a different example, \cite{Umlauft2017Bayesian} used eigendecomposition for impedance adaptation.  

In order to comply with such geometry constraints, researchers provided new formulation of \acp{dmp} that ensures proper unit quaternions or rotation matrices over the course of orientation adaptation \cite{AbuDakka2015Adaptation,Ude2014Orientation,saveriano2019merging,koutras2020correct}, and proper \ac{spd} matrices over the course of the adaptation of \ac{spd} profiles (\eg stiffness or manipulability ellipsoids) \citep{abudakka2020Geometry}. We believe that using these geometry-aware \acp{dmp} is preferable to encode data with underlying geometry constraints.

\subsubsection{Weights learning method}
\acp{dmp} represent motion trajectories as stable dynamical systems with learnable weights that define the shape of the motion. In the \ac{lfd} paradigm, \ac{dmp} weights are usually learned in a supervised manner using human demonstrations. The procedure used to transform human demonstrations into training data for the \ac{dmp} forcing term is highlighted in \secref{par:learning_forcing}. Given the training data, different techniques can be used to fit the weights.

\ac{lwr} is widely used when the forcing term is a combination of \acp{rbf} as in \eqref{eq:fx_discrete}. If multiple demonstrations are given, one can exploit \ac{gmm}/\ac{gmr} as in \citep{Pervez2017Novel} or \ac{gpr} as in \citep{Fanger2016Gaussian} to represent the forcing term and use expectation--maximization to fit the (hyper-)parameters. Deep \acp{nn}, typically trained via back-propagation, seem an appealing possibility to map input images into forcing terms \citep{Pervez2017}, mimicking the human perception-action loop. Although appealing, the possibility of exploiting deep learning techniques as motion primitives requires further investigations.

\begin{table*}[t]
\small\sf\centering
\caption{Open-source implementations of \ac{dmp}-based approaches that we have released to the community. The source code for each appraoch is available at \url{https://gitlab.com/dmp-codes-collection}} 
\begin{tabular}{m{0.15\textwidth}m{0.2\textwidth}m{0.1\textwidth}m{0.45\textwidth}}
\toprule
\textbf{Approach} & \textbf{Author} & \textbf{Language} & \textbf{Description} \\
\midrule
Discrete \ac{dmp} & \emph{Fares J. Abu-Dakka} & \cpp{} & An implementation for discrete \ac{dmp} based on the work in \citep{ude2010task,AbuDakka2015Adaptation,Ude2014Orientation}.\\\\
Periodic \ac{dmp} & \emph{Luka Peternel} & \python{} & An implementation for periodic \ac{dmp} based on the work in \citep{peternel2016adaptive}.\\\\
Unit quaternion \ac{dmp} & \emph{Fares J. Abu-Dakka}  & \matlab{} and \cpp{} & An implementation for unit quaternion \ac{dmp} and goal switching based on the work in \citep{AbuDakka2015Adaptation,Ude2014Orientation}.\\\\
\ac{spd} \ac{dmp} & \emph{Fares J. Abu-Dakka}  & \matlab{} & An implementation for \ac{spd} \ac{dmp} and goal switching based on the work in \citep{abudakka2020Geometry}.\\\\
Joining \acp{dmp} &  \emph{Matteo Saveriano}  & \matlab{} & An implementation for joining multiple \acp{dmp} based on the work in \citep{saveriano2019merging}.\\\\
Coupling-force \acp{dmp} &  \emph{Aljaz Kramberger}  & \matlab{} & An implementation for discrete \acp{dmp} and force coupling terms based on the work in \citep{kramberger2018passivity}.\\
\bottomrule
\end{tabular}
\label{tab:source_code}
\end{table*}

In real applications, there can be a misplacement between the \ac{dmp} trajectory and the robot motion. Typical examples include assembly or other tasks that require physical interaction with the environment (see \secref{subsec:passive_environment}). In this situations, the \ac{dmp} motion can be incrementally adjusted to improve the robot performance. \ac{ilc} arises as an interesting approach to iteratively update the \ac{dmp} weights as it ensures a rapid convergence to the desired performance \citep{Gams2014, AbuDakka2015Adaptation, kramberger2018passivity}. However, \ac{ilc} assumes that a target behavior to reproduce is given. When the target behavior cannot be easily specified and the robot performance is not satisfactory, \ac{rl} solutions have to be adopted. As detailed in \secref{subsec:rl}, \acp{dmp} are effective control policies and, combined with policy search algorithms like \ac{pi2} or \ac{power}, are able to solve complex and highly dynamic tasks.

\subsubsection{Online adaptation}

Performing robotic tasks in the real world requires adaptation capabilities. When adaptation of \acp{dmp} based on some feedback is required, one of the extension methods should be applied. For example, to change the existing movement based a detected obstacle, the method in \citep{park2008movement, hoffmann2009biologically, tan2011potential,gams2016adaptation} can be used (see \secref{subsubsec:obstacleAvoidance}). If it is necessary to adaptively learn the movement dynamics based on real-time effort feedback, the method in \citep{peternel2016adaptive} can be employed (see \secref{subsec:onlineUpdate}). 

Furthermore, for industrial tasks, such as assembly or polishing, adaptation strategies combining force control with demonstrated trajectories can be applied \citep{AbuDakka2015Adaptation,kramberger2016generalization,gams2010line}, ensuring the system will follow the predefined trajectory and adapt to the environmental uncertainties. For online adaptation \acp{dmp} can be used as a trajectory generator, which output represents an input to the force control algorithm, on the other hand, force feedback can directly be incorporated as a coupling term in the \acp{dmp} formulation (see \secref{subsubsec:Forcecoupling}), eliminating the need for an additional force controller. Similar approach can also be utilized for velocity based adaptation of the movements (see \secref{subsec:velocityTerm}).  

\subsubsection{Impedance vs force}

In physical interaction tasks, \acp{dmp} can be used to either learn force or impedance \citep{peternel2017method}. If the task requires position control, then the impedance should be learned with \acp{dmp} in combination with the reference position. If the task requires to control a specific force, \eg pushing on a surface during the wiping and drilling, either force or impedance is feasible. However, if safety is the most critical aspect, the \acp{dmp} should be used to learn impedance control so that the robot can be made soft. 

Furthermore, to overcome any undesirable movements, the control policy can be augmented with a tank-based passivity approach \citep{Shahriari2017}. This approach monitors the energy flow between the modeled sub-systems, \eg \acp{dmp} trajectory generation, impedance control, environment. In an event of an energy violation, the system will first try to passively compensate for the violation and subsequently if the violation cannot be compensated e.g. the energy tank is depleted, stop the system. 
In cases, where the task characteristics are not fully known, a learning policy can be added on top of the passivity approach \citep{kramberger2018passivity} in-order to learn the overall energy requirements for the task.

\subsection{Resources and codes}\label{sec:code}

Availability of code and datasets is useful to speed-up the setup of novel applications without the need of re-implementing a promising approach from scratch. We have searched for available \ac{dmp} implementation and found out that several researchers published their \ac{dmp} codes in various open-source repositories. We decided to list the available implementations on the Git repository that accompanies this paper (\url{https://gitlab.com/dmp-codes-collection/third-party-dmp}). For each implementation, we mention the type of \ac{dmp}, the author, the url to download the code, and the used programming language. We also provide a short description of the key features.

Apart from listing existing approaches, the Git repository that accompanies this paper contains implementation that we decided to release to the community. The list of provided implementations is given in \tabref{tab:source_code}.

\subsection{Limitations and open issues}
As any motion primitive representation, \acp{dmp} have strengths but also inherent limitations. The advantages of the \acp{dmp} have been widely discussed in previous sections. Here, we present the main limitations of the \acp{dmp} and discuss open issues that require further investigation. A summary of these limitations is presented in \tabref{tab:comparison}.   

\subsubsection{Implicit time dependency}
The phase variable used to suppress the non-linear forcing term and ensure convergence to a given goal introduces an implicit time dependency in the \ac{dmp} formulation. 
The reason for representing the time dependency implicitly as a dynamical system is that such a phase variable can be conveniently manipulated. For example, in \secref{par:phase_stopping}, we have seen how to manipulate the phase variable to slow-down (or even stop) the execution.
A drawback of the time dependency is that the shape of the \ac{dmp} motion is significantly affected by the time evolution of the phase variable. If the phase vanishes too early, the last part of the trajectory is executed with a linear dynamics converging to the goal. If the phase lasts too long, the trajectory may overshoot and fail to reach the goal within the desired time. In both cases, the \ac{dmp} motion may significantly deviate from the demonstration. A properly designed phase stopping mechanism can remedy the issue, but the proper phase stopping to adopt depends on the specific application. 

In order to overcome this limitation, several authors focused on learning stable and time-independent (or autonomous) dynamical systems from demonstrations. A globally stable and autonomous system generates a vector field that converges to the given goal from any initial state. Without the need of a phase variable, the generated motion depends only of the current state of the system.  Notable approaches to learn stable and autonomous systems exploit Lyapunov theory \citep{Khansari2011learning, Khansari2014learning}, contraction theory \citep{ravichandar2015learning, Blocher2017learning}, diffeomorphic transformations \citep{Neumann2015learning, Perrin16fast}, and passivity considerations \citep{kronander2015passive}. These approaches have been effectively used to learn complex movements from demonstrations. 

In general, autonomous systems have the potential to represent much more complex movements than \acp{dmp}. For example, autonomous systems can encode different motions in different regions of the state-space. In this respect, \acp{dmp} can only generate a stereotypical trajectory connecting the start to the goal, regardless where the initial state is placed in the state-space. However, the stereotypical motion generation is also an advantage of \acp{dmp} since it makes easier to predict the generated motion in regions of the state-space poorly covered by training data. On the contrary, it is hard to predict how an autonomous system generalizes where only few or no training data are available. \acp{dmp} are know to scale well in high-dimensional spaces since the learned forcing term always depends on a shared, scalar phase variable. Autonomous systems perform learning directly on the high-dimensional state-space, which poses numerical challenges and requires much more training data.  In synthesis, each representation has its own advantages and disadvantages and the choice between time-dependant and autonomous motion primitives depends on the specific application.

\subsubsection{Stochastic information}
Representing the demonstrated motion as a probability distribution has several advantages. For example, in a probabilistic framework the generalization to new a goal (or a via-point) is achieved using conditioning on the new goal (via-point), while the covariance computed from the probabiltiy distribution can represent couplings between different \acp{dof}  \citep{Paraschos2013}.
As a matter of fact, classical \acp{dmp} are deterministic and lack the stochastic information on the modelled motion.

\cite{BenAmor2014} proposed an approach to estimate the predictive distribution $\mathcal{P}(\w|y_{1:\mathfrak{T}})$ that relates the \ac{dmp} weights $\w$ and a partial trajectory $y_{1:\mathfrak{T}}$ observed for $\mathfrak{T}$ time instants. $\mathcal{P}(\w|y_{1:\mathfrak{T}})$ is used to estimate the most likely weights given a partial movement and to reconstruct the missing part of the trajectory. However, a full probabilistic characterization of \acp{dmp} is still missing.

The \ac{promp} framework \citep{Paraschos2013} proposed an alternative movement primitive representation that contains information about the variability across different demonstrations as well as different \acp{dof} in the form of a covariance matrix. This enables to explicitly encode the couplings between different directions and to increase the generalization by conditioning on a desired goal, via-point, or intermediate velocity. The covariance computed by \acp{promp} represent the variability and the correlation in the demonstrations. In other representations, like \ac{gpr}, the covariance is a measure of the model uncertainty due to the lack of training data. \acp{kmp} \citep{huang2019kernelized, silverio2019uncertainty} offer the possibility of modelling variability, correlation, and uncertainty in the same framework. However, \ac{kmp}'s computational cost can be elevated compared to \ac{dmp} in longer trajectories due to the computation of the inverse of the kernel matrix.

\subsubsection{Closed-loop implementation and issues}
A vast majority of methods employ \acp{dmp} only as a reference trajectory generator for the closed-loop controller, which then actually executes it. However, the \acp{dmp} can also be used as a part of the close-loop controller itself and only a few methods explored this concept. For example, in \cite{peternel2016adaptive} the \acp{dmp} are directly torque generators for exoskeleton actuators in the control loop, which is closed by a feedback from the human user's muscle activity. Nevertheless, in such scenario the closed-loop stability and passivity become crucial considerations that have to be addressed and resolved before the wide-spread application \citep{kramberger2018passivity}.

\subsubsection{Coping with high-dimensional inputs}
One of the main limitations of \ac{dmp} is that it encodes human and robot trajectories explicitly with the time (\ie 1--D input) which may lead to a synchronization issues  since human motions in the new evaluations could be significantly different (\eg faster/slower velocity) from the demonstrated ones. In order to avoid synchronization problem, \cite{BenAmor2014} designed a time-alignment strategy, while \citep{Pervez2017Novel} estimated the phase signal during the training using expectation-maximization~\citep{bishop2006pattern}.

As the \ac{dmp} models trajectories using basis functions, this works effectively when learning time-driven trajectories (\ie 1--D input). However, when demonstrations comprise high-dimensional inputs, specifying the center vectors and widths of basis functions becomes quite cumbersome. Specifically, as discussed in \citep{bishop2006pattern} the number of basis functions often increases exponentially when the dimension of inputs increases. To alleviate this limitation, some approaches investigated modern deep learning techniques. \cite{Pahic2020} used a deep \ac{nn} to synthesize \ac{dmp} weights from an input image. The classical \ac{dmp} formulation is then used to generate motion trajectories. \cite{Pervez2017} used a \ac{cnn}~\citep{lecun2015deep} to predict 2--D task parameters (\eg the position of a target) from an input image and a fully connected \ac{nn} to retrieve the forcing term from the 2--D parameters and the phase variable. The \ac{cnn} and the fully connected \ac{nn} are trained in two separate stages. The approach is promising, but the separate training  of the two networks increases the pre-processing and complicates the learning process.

Alternative approaches in literature, such as \ac{gmm}/\ac{gmr} \citep{Calinon2016}, \ac{tpgmm} \citep{Calinon2016}, \ac{kmp} \citep{huang2019kernelized, Huang2020Toward}, can be directly applied for learning demonstrations comprising high-dimensional inputs.

\begin{table}[!th]
\small\sf\centering
	\caption{A summary of \ac{dmp} features and limitations that have been solved (\cmark) or partially solved (\lmark).} 
	\resizebox{\columnwidth}{!}{%
			\begin{tabular}{m{0.2\columnwidth}m{0.6\columnwidth}m{0.1\columnwidth}}
			\toprule
			\textbf{Limitation}	& \textbf{Related work} & \textbf{Status} \\
			\toprule 
				Via-points  & \citep{ning2011accurate,ning2012novel,weitschat2018safe,saveriano2019merging,Zhou2019Learning} &  \multicolumn{1}{c}{\cmark}\\
				\midrule
				Start-point  & \citep{hoffmann2009biologically,Ijspeert2013Dynamical, weitschat2013dynamic, dragan2015movement} & \multicolumn{1}{c}{\cmark} \\
				\midrule
				Goal-point  & \citep{Ijspeert2013Dynamical,Ude2014Orientation,abudakka2020Geometry, dragan2015movement, weitschat2018safe} & \multicolumn{1}{c}{\cmark} \\
				\midrule
				Obstacle avoidance  & \citep{park2008movement, hoffmann2009biologically,tan2011potential,Kim2015Adaptability,Rai2017Learning} & \multicolumn{1}{c}{\cmark} \\
				\midrule
				Geometry-constrained data  & \citep{Pastor2009,AbuDakka2015Adaptation,Ude2014Orientation,saveriano2019merging,abudakka2020Geometry} & \multicolumn{1}{c}{\lmark \tablefootnote{The referred work extended the classical \ac{dmp} to different space like $\bm{\mathcal{SO}}(3)$ or $\spd$. Although formally similar, the extention to other Riemannian manifolds like the Grassmannian or the Hyperbolic  manifolds is non-trivial and still not fully addressed.}} \\\midrule
				Probabilistic  & \citep{BenAmor2014}  & \multicolumn{1}{c}{\lmark} \\
				\midrule
				Extrapolation  & \citep{pervez2018learning, Zhou2019Learning} & \multicolumn{1}{c}{\lmark} \\
				\midrule
				High-dim input  & \citep{Pervez2017Novel, Pahic2020} & \multicolumn{1}{c}{\lmark} \\
				\midrule
				Closed-loop  & \citep{peternel2016adaptive, kramberger2018passivity} & \multicolumn{1}{c}{\lmark} \\\midrule
				Multi-attractor  & \citep{nemec2018anefficient,san2019towards} & \multicolumn{1}{c}{\lmark} \\
				\bottomrule 
			\end{tabular}
	}
	\label{tab:conclusion}
\end{table}

\subsubsection{Multi-attractor systems}
The well known second-order dynamic properties of the \acp{dmp}, strive towards a single attractor system \citep{ijspeert2002learningAttractor}. The properties, \eg convergence and modulation of the motion, are well studied and implementations can be found in many research papers. Because of the second-order dynamics, the system becomes unstable if for example the motion is reversed during the execution. In the past years, two main approaches describing the reversibility problem have been introduced. In the first approach \citep{nemec2018anefficient}, reversibility is considered as leaning two separate primitives, one for each direction of the motion. The approach is promising, but does not reflect true reversibility, because it uses one attractor point for each primitive. 

On the other hand, \citep{san2019towards} introduced an alternative formulation with two stable attractor systems. The first attractor is defined at the starting point $y_0$ of the trajectory and the subsequent one at the goal $g$, the dynamical system between them guaranties a stable convergence depending on the selected attractor. The approach demonstrated true reversibility, while keeping all the \acp{dmp} properties. Nevertheless, all questions have not been resolved yet, the approach was evaluated on tasks and joint space position trajectories. A proper formulation for dealing with orientations e.g. quaternions in task space is still missing, 

%% file: root.bbl
\begin{thebibliography}{328}
\providecommand{\natexlab}[1]{#1}
\providecommand{\url}[1]{\texttt{#1}}
\providecommand{\urlprefix}{URL }
\expandafter\ifx\csname urlstyle\endcsname\relax
  \providecommand{\doi}[1]{DOI:\discretionary{}{}{}#1}\else
  \providecommand{\doi}{DOI:\discretionary{}{}{}\begingroup
  \urlstyle{rm}\Url}\fi

\bibitem[{Abdelrahman et~al.(2020)Abdelrahman, Mitrevski and
  Ploger}]{Abdelrahman2020}
Abdelrahman A, Mitrevski A and Ploger P (2020) Context-aware task execution
  using apprenticeship learning.
\newblock In: \emph{IEEE International Conference on Robotics and Automation}.
  pp. 1329--1335.

\bibitem[{Abu-Dakka et~al.(2014)Abu-Dakka, Nemec, Kramberger, Buch, Kr{\"u}ger
  and Ude}]{abudakka2014solving}
Abu-Dakka F, Nemec B, Kramberger A, Buch A, Kr{\"u}ger N and Ude A (2014)
  Solving peg-in-hole tasks by human demonstration and exception strategies.
\newblock \emph{Industrial Robot} 41(6): 575--584.

\bibitem[{Abu-Dakka and Kyrki(2020)}]{abudakka2020Geometry}
Abu-Dakka FJ and Kyrki V (2020) Geometry-aware dynamic movement primitives.
\newblock In: \emph{IEEE International Conference on Robotics and Automation}.
  Paris, France, pp. 4421--4426.

\bibitem[{Abu-Dakka et~al.(2015{\natexlab{a}})Abu-Dakka, Nemec, J{\o}rgensen,
  Savarimuthu, Kr{\"u}ger and Ude}]{AbuDakka2015Adaptation}
Abu-Dakka FJ, Nemec B, J{\o}rgensen JA, Savarimuthu TR, Kr{\"u}ger N and Ude A
  (2015{\natexlab{a}}) Adaptation of manipulation skills in physical contact
  with the environment to reference force profiles.
\newblock \emph{Autonomous Robots} 39(2): 199--217.

\bibitem[{Abu-Dakka et~al.(2018)Abu-Dakka, Rozo and
  Caldwell}]{abudakka2018force}
Abu-Dakka FJ, Rozo L and Caldwell DG (2018) Force-based variable impedance
  learning for robotic manipulation.
\newblock \emph{Robotics and Autonomous Systems} 109: 156--167.

\bibitem[{Abu-Dakka and Saveriano(2020)}]{abudakka2020Variable}
Abu-Dakka FJ and Saveriano M (2020) Variable impedance control and learning --
  a review.
\newblock \emph{Frontiers in Robotics and AI} 7: 177.

\bibitem[{Abu-Dakka et~al.(2015{\natexlab{b}})Abu-Dakka, Valera, Escalera,
  Vall{\'e}s, Mata and Abderrahim}]{abudakka2015Trajectory}
Abu-Dakka FJ, Valera A, Escalera J, Vall{\'e}s M, Mata V and Abderrahim M
  (2015{\natexlab{b}}) Trajectory adaptation and learning for ankle
  rehabilitation using a 3-prs parallel robot.
\newblock In: Liu H, Kubota N, Zhu X, Dillmann R and Zhou D (eds.)
  \emph{Intelligent Robotics and Applications}. Cham: Springer International
  Publishing, pp. 483--494.

\bibitem[{Abu-Dakka et~al.(2020)Abu-Dakka, Valera, Escalera, Abderrahim, Page
  and Mata}]{abudakka2020passive}
Abu-Dakka FJ, Valera A, Escalera JA, Abderrahim M, Page A and Mata V (2020)
  Passive exercise adaptation for ankle rehabilitation based on learning
  control framework.
\newblock \emph{Sensors} 20(21): 6215.

\bibitem[{Aein et~al.(2013)Aein, Aksoy, Tamosiunaite, Papon, Ude and
  W{\"o}rg{\"o}tter}]{aein2013toward}
Aein M, Aksoy E, Tamosiunaite M, Papon J, Ude A and W{\"o}rg{\"o}tter F (2013)
  Toward a library of manipulation actions based on semantic object-action
  relations.
\newblock In: \emph{IEEE/RSJ International Conference on Intelligent Robots and
  Systems}. pp. 4555--4562.

\bibitem[{Agostini et~al.(2020)Agostini, Saveriano, Lee and
  Piater}]{agostini2020manipulation}
Agostini A, Saveriano M, Lee D and Piater J (2020) Manipulation planning using
  object-centered predicates and hierarchical decomposition of contextual
  actions.
\newblock \emph{IEEE Robotics and Automation Letters} 5(4): 5629--5636.

\bibitem[{Ajallooeian et~al.(2013)Ajallooeian, Van Den~Kieboom, Mukovskiy,
  Giese and Ijspeert}]{ajallooeian2013general}
Ajallooeian M, Van Den~Kieboom J, Mukovskiy A, Giese M and Ijspeert A (2013) A
  general family of morphed nonlinear phase oscillators with arbitrary limit
  cycle shape.
\newblock \emph{Physica D: Nonlinear Phenomena} 263: 41--56.

\bibitem[{Ajoudani et~al.(2012)Ajoudani, Tsagarakis and
  Bicchi}]{ajoudani2012tele}
Ajoudani A, Tsagarakis N and Bicchi A (2012) Tele-impedance: Teleoperation with
  impedance regulation using a body--machine interface.
\newblock \emph{The International Journal of Robotics Research} 31(13):
  1642--1656.

\bibitem[{Ajoudani et~al.(2018)Ajoudani, Zanchettin, Ivaldi, Albu-Sch{\"a}ffer,
  Kosuge and Khatib}]{ajoudani2018progress}
Ajoudani A, Zanchettin AM, Ivaldi S, Albu-Sch{\"a}ffer A, Kosuge K and Khatib O
  (2018) Progress and prospects of the human--robot collaboration.
\newblock \emph{Autonomous Robots} 42(5): 957--975.

\bibitem[{Alizadeh et~al.(2016)Alizadeh, Malekzadeh and
  Barzegari}]{alizadeh2016learning}
Alizadeh T, Malekzadeh M and Barzegari S (2016) Learning from demonstration
  with partially observable task parameters using dynamic movement primitives
  and gaussian process regression.
\newblock In: \emph{IEEE/ASME International Conference on Advanced Intelligent
  Mechatronics}. pp. 889--894.

\bibitem[{Amatya et~al.(2020)Amatya, Rezayat~Sorkhabadi and Zhang}]{Amatya2020}
Amatya S, Rezayat~Sorkhabadi S and Zhang W (2020) Human learning and
  coordination in lower-limb physical interactions.
\newblock In: \emph{Proceedings of the American Control Conference}, volume
  2020-July. pp. 557--562.

\bibitem[{Andr\'{e} et~al.(2016)Andr\'{e}, Santos and Costa}]{Andre2016Skill}
Andr\'{e} J, Santos C and Costa L (2016) Skill memory in biped locomotion:
  Using perceptual information to predict task outcome.
\newblock \emph{Journal of Intelligent and Robotic Systems: Theory and
  Applications} 82(3-4): 379--397.

\bibitem[{Andr\'{e} et~al.(2015)Andr\'{e}, Teixeira, Santos and
  Costa}]{Andre2015Adapting}
Andr\'{e} J, Teixeira C, Santos C and Costa L (2015) Adapting biped locomotion
  to sloped environments: Combining reinforcement learning with dynamical
  systems.
\newblock \emph{Journal of Intelligent and Robotic Systems: Theory and
  Applications} 80(3-4): 625--640.

\bibitem[{Angelov et~al.(2020)Angelov, Hristov, Burke and
  Ramamoorthy}]{Angelov20202658}
Angelov D, Hristov Y, Burke M and Ramamoorthy S (2020) Composing diverse
  policies for temporally extended tasks.
\newblock \emph{IEEE Robotics and Automation Letters} 5(2): 2658--2665.

\bibitem[{Atkeson et~al.(1997)Atkeson, Moore and Schaal}]{atkeson1997locally}
Atkeson CG, Moore AW and Schaal S (1997) Locally weighted learning.
\newblock \emph{Artificial Intelligence Review} 11: 11--73.

\bibitem[{Basa and Schneider(2015)}]{basa2015learning}
Basa D and Schneider A (2015) Learning point-to-point movements on an elastic
  limb using dynamic movement primitives.
\newblock \emph{Robotics and Autonomous Systems} 66: 55--63.

\bibitem[{Beetz et~al.(2010)Beetz, Stulp, Esden-Tempski, Fedrizzi, Klank,
  Kresse, Maldonado and Ruiz}]{beetz2010generality}
Beetz M, Stulp F, Esden-Tempski P, Fedrizzi A, Klank U, Kresse I, Maldonado A
  and Ruiz F (2010) Generality and legibility in mobile manipulation: Learning
  skills for routine tasks.
\newblock \emph{Autonomous Robots} 28(1): 21--44.

\bibitem[{Beik-Mohammadi et~al.(2020)Beik-Mohammadi, Kerzel, Pleintinger,
  Hulin, Reisich, Schmidt, Pereira, Wermter and Lii}]{Beik-Mohammadi2020}
Beik-Mohammadi H, Kerzel M, Pleintinger B, Hulin T, Reisich P, Schmidt A,
  Pereira A, Wermter S and Lii N (2020) Model mediated teleoperation with a
  hand-arm exoskeleton in long time delays using reinforcement learning.
\newblock In: \emph{IEEE International Conference on Robot and Human
  Interactive Communication}. pp. 713--720.

\bibitem[{Ben~Amor et~al.(2014)Ben~Amor, Neumann, Kamthe, Kr{\"o}mer and
  Peters}]{BenAmor2014}
Ben~Amor H, Neumann G, Kamthe S, Kr{\"o}mer O and Peters J (2014) Interaction
  primitives for human-robot cooperation tasks.
\newblock In: \emph{IEEE International Conference on Robotics and Automation}.
  Hong Kong, China, pp. 2831--2837.

\bibitem[{Bian et~al.(2019)Bian, Ren, Li, Liang, Wang and
  Zhao}]{Bian2019extended}
Bian F, Ren D, Li R, Liang P, Wang K and Zhao L (2019) An extended dmp
  framework for robot learning and improving variable stiffness manipulation.
\newblock \emph{Assembly Automation} 40(1): 85--94.

\bibitem[{Billard et~al.(2016)Billard, Calinon and
  Dillmann}]{Billard16learning}
Billard A, Calinon S and Dillmann R (2016) Learning from humans.
\newblock In: Siciliano B and Khatib O (eds.) \emph{Handbook of Robotics},
  chapter~74. Secaucus, NJ, USA: Springer, pp. 1995--2014.
\newblock 2nd Edition.

\bibitem[{Bishop(2006)}]{bishop2006pattern}
Bishop CM (2006) \emph{Linear Models for Regression}.
\newblock Springer, pp. 172--173.

\bibitem[{Bitzer and Vijayakumar(2009)}]{bitzer2009latent}
Bitzer S and Vijayakumar S (2009) Latent spaces for dynamic movement
  primitives.
\newblock In: \emph{IEEE-RAS International Conference on Humanoid Robots}. pp.
  574--581.

\bibitem[{Blocher et~al.(2017)Blocher, Saveriano and Lee}]{Blocher2017learning}
Blocher C, Saveriano M and Lee D (2017) Learning stable dynamical systems using
  contraction theory.
\newblock In: \emph{International Conference on Ubiquitous Robots and Ambient
  Intelligence}. pp. 124--129.

\bibitem[{Bristow et~al.(2006)Bristow, Tharayil and
  Alleyne}]{bristow2006survey}
Bristow DA, Tharayil M and Alleyne AG (2006) A survey of iterative learning
  control.
\newblock \emph{Control Systems Magazine} 26(3): 96--114.

\bibitem[{Buchli et~al.(2011{\natexlab{a}})Buchli, Stulp, Theodorou and
  Schaal}]{buchli2011learning}
Buchli J, Stulp F, Theodorou E and Schaal S (2011{\natexlab{a}}) Learning
  variable impedance control.
\newblock \emph{The International Journal of Robotics Research} 30(7):
  820--833.

\bibitem[{Buchli et~al.(2011{\natexlab{b}})Buchli, Theodorou, Stulp and
  Schaal}]{buchli2010variable}
Buchli J, Theodorou E, Stulp F and Schaal S (2011{\natexlab{b}}) Variable
  impedance control: a reinforcement learning approach.
\newblock \emph{Robotics: Science and Systems VI} : 153.

\bibitem[{Burdet et~al.(2001)Burdet, Osu, Franklin, Milner and
  Kawato}]{burdet2001central}
Burdet E, Osu R, Franklin DW, Milner TE and Kawato M (2001) The central nervous
  system stabilizes unstable dynamics by learning optimal impedance.
\newblock \emph{Nature} 414(6862): 446--449.

\bibitem[{Caccavale et~al.(2019)Caccavale, Saveriano, Finzi and
  Lee}]{caccavale2019kinesthetic}
Caccavale R, Saveriano M, Finzi A and Lee D (2019) Kinesthetic teaching and
  attentional supervision of structured tasks in human–robot interaction.
\newblock \emph{Autonomous Robots} 43(6): 1291--1307.

\bibitem[{Caccavale et~al.(2018)Caccavale, Saveriano, Fontanelli, Ficuciello,
  Lee and Finzi}]{caccavale2018imitation}
Caccavale R, Saveriano M, Fontanelli G, Ficuciello F, Lee D and Finzi A (2018)
  Imitation learning and attentional supervision of dual-arm structured tasks.
\newblock In: \emph{Joint IEEE International Conference on Development and
  Learning and on Epigenetic Robotics}. Lisbon, Portugal, pp. 66--71.

\bibitem[{Calinon(2016)}]{Calinon2016}
Calinon S (2016) A tutorial on task-parameterized movement learning and
  retrieval.
\newblock \emph{Intelligent Service Robotics} 9(1): 1--29.

\bibitem[{{Calinon} et~al.(2009){Calinon}, {D'halluin}, {Caldwell} and
  {Billard}}]{calinon2009handling}
{Calinon} S, {D'halluin} F, {Caldwell} DG and {Billard} AG (2009) Handling of
  multiple constraints and motion alternatives in a robot programming by
  demonstration framework.
\newblock In: \emph{IEEE-RAS International Conference on Humanoid Robots}. pp.
  582--588.

\bibitem[{Calinon et~al.(2012)Calinon, Li, Alizadeh, Tsagarakis and
  Caldwell}]{calinon2012statistical}
Calinon S, Li Z, Alizadeh T, Tsagarakis N and Caldwell D (2012) Statistical
  dynamical systems for skills acquisition in humanoids.
\newblock In: \emph{IEEE-RAS International Conference on Humanoid Robots}. pp.
  323--329.

\bibitem[{Carrera et~al.(2015)Carrera, Palomeras, Hurtós, Kormushev and
  Carreras}]{Carrera2015Cognitive}
Carrera A, Palomeras N, Hurtós N, Kormushev P and Carreras M (2015) Cognitive
  system for autonomous underwater intervention.
\newblock \emph{Pattern Recognition Letters} 67: 91--99.

\bibitem[{Chang and Kuli{\'c}(2013)}]{chang2013motion}
Chang G and Kuli{\'c} D (2013) Motion learning from observation using affinity
  propagation clustering.
\newblock In: \emph{IEEE International Symposium on Robot and Human Interactive
  Communication}. pp. 662--667.

\bibitem[{Chen et~al.(2015)Chen, Bayer, Urban and Van
  Der~Smagt}]{chen2015efficient}
Chen N, Bayer J, Urban S and Van Der~Smagt P (2015) Efficient movement
  representation by embedding dynamic movement primitives in deep autoencoders.
\newblock In: \emph{IEEE-RAS International Conference on Humanoid Robots}.
  IEEE, pp. 434--440.

\bibitem[{Chen et~al.(2016)Chen, Karl and Van Der~Smagt}]{chen2016dynamic}
Chen N, Karl M and Van Der~Smagt P (2016) Dynamic movement primitives in latent
  space of time-dependent variational autoencoders.
\newblock In: \emph{IEEE-RAS International Conference on Humanoid Robots}. pp.
  629--636.

\bibitem[{Chen and Liu(2018)}]{chen2018lifelong}
Chen Z and Liu B (2018) Lifelong machine learning.
\newblock \emph{Synthesis Lectures on Artificial Intelligence and Machine
  Learning} 12(3): 1--207.

\bibitem[{Chi et~al.(2018)Chi, Liu, Abdelaziz, Dagnino, Riga, Bicknell and
  Yang}]{Chi2018}
Chi W, Liu J, Abdelaziz M, Dagnino G, Riga C, Bicknell C and Yang GZ (2018)
  Trajectory optimization of robot-assisted endovascular catheterization with
  reinforcement learning.
\newblock In: \emph{IEEE/RSJ International Conference on Intelligent Robots and
  Systems}. pp. 3875--3881.

\bibitem[{Chiaverini and Siciliano(1999)}]{chiaverini1999unit}
Chiaverini S and Siciliano B (1999) The unit quaternion: A useful tool for
  inverse kinematics of robot manipulators.
\newblock \emph{Systems Analysis Modelling Simulation} 35(1): 45--60.

\bibitem[{Cho et~al.(2019)Cho, Lee, Suh and Kim}]{Cho2019}
Cho N, Lee S, Suh I and Kim HS (2019) Relationship between the order for motor
  skill transfer and motion complexity in reinforcement learning.
\newblock \emph{IEEE Robotics and Automation Letters} 4(2): 293--300.

\bibitem[{Churchill and Fernando(2014)}]{churchill2014evolutionary}
Churchill A and Fernando C (2014) An evolutionary cognitive architecture made
  of a bag of networks.
\newblock \emph{Evolutionary Intelligence} 7(3): 169--182.

\bibitem[{Cohen and Berman(2014)}]{cohen2014integrating}
Cohen A and Berman S (2014) Integrating simulation with robotic learning from
  demonstration.
\newblock In: \emph{European Conference on Modelling and Simulation}. pp.
  421--427.

\bibitem[{Cohn et~al.(1996)Cohn, Ghahramani and Jordan}]{cohn1996active}
Cohn DA, Ghahramani Z and Jordan MI (1996) Active learning with statistical
  models.
\newblock \emph{Journal of artificial intelligence research} 4: 129--145.

\bibitem[{Colome et~al.(2015)Colome, Planells and Torras}]{Colome2015Afriction}
Colome A, Planells A and Torras C (2015) A friction-model-based framework for
  reinforcement learning of robotic tasks in non-rigid environments.
\newblock In: \emph{IEEE International Conference on Robotics and Automation}.
  pp. 5649--5654.

\bibitem[{Colom{\'e} and Torras(2014)}]{colome2014dimensionality}
Colom{\'e} A and Torras C (2014) Dimensionality reduction and motion
  coordination in learning trajectories with dynamic movement primitives.
\newblock In: \emph{IEEE/RSJ International Conference on Intelligent Robots and
  Systems}. pp. 1414--1420.

\bibitem[{Colom{\'e} and Torras(2018)}]{colome2018dimensionality}
Colom{\'e} A and Torras C (2018) Dimensionality reduction for dynamic movement
  primitives and application to bimanual manipulation of clothes.
\newblock \emph{IEEE Transactions on Robotics} 34(3): 602--615.

\bibitem[{Cui et~al.(2016)Cui, Poon, Matsubara, Miro, Sugimoto and
  Yamazaki}]{Cui2016Environment}
Cui Y, Poon J, Matsubara T, Miro J, Sugimoto K and Yamazaki K (2016)
  Environment-adaptive interaction primitives for human-robot motor skill
  learning.
\newblock In: \emph{IEEE-RAS International Conference on Humanoid Robots}. pp.
  711--717.

\bibitem[{Cui et~al.(2019)Cui, Poon, Miro, Yamazaki, Sugimoto and
  Matsubara}]{cui2019environment}
Cui Y, Poon J, Miro JV, Yamazaki K, Sugimoto K and Matsubara T (2019)
  Environment-adaptive interaction primitives through visual context for
  human--robot motor skill learning.
\newblock \emph{Autonomous Robots} 43(5): 1225--1240.

\bibitem[{Dahlin and Karayiannidis(2020)}]{Dahlin2020438}
Dahlin A and Karayiannidis Y (2020) Adaptive trajectory generation under
  velocity constraints using dynamical movement primitives.
\newblock \emph{IEEE Control Systems Letters} 4(2): 438--443.

\bibitem[{Deni{\v s}a and Ude(2013{\natexlab{a}})}]{denisa2013discovering}
Deni{\v s}a M and Ude A (2013{\natexlab{a}}) Discovering new motor primitives
  in transition graphs.
\newblock \emph{Advances in Intelligent Systems and Computing} 193(1):
  219--230.

\bibitem[{Deni{\v s}a and Ude(2013{\natexlab{b}})}]{denisa2013newmotor}
Deni{\v s}a M and Ude A (2013{\natexlab{b}}) New motor primitives through graph
  search, interpolation and generalization.
\newblock \emph{Studies in Computational Intelligence} 466: 137--148.

\bibitem[{Deni\v{s}a et~al.(2016{\natexlab{a}})Deni\v{s}a, Gams, Ude and
  Petri\v{c}}]{Denisa2016}
Deni\v{s}a M, Gams A, Ude A and Petri\v{c} T (2016{\natexlab{a}}) Learning
  compliant movement primitives through demonstration and statistical
  generalization.
\newblock \emph{IEEE/ASME Transactions on Mechatronics} 21(5): 2581--2594.

\bibitem[{Deni\v{s}a et~al.(2016{\natexlab{b}})Deni\v{s}a, Petri\v{c}, Gams and
  Ude}]{denivsa2016review}
Deni\v{s}a M, Petri\v{c} T, Gams A and Ude A (2016{\natexlab{b}}) A review of
  compliant movement primitives.
\newblock In: Hurtado EG (ed.) \emph{Robot Control}, chapter~1. Rijeka:
  IntechOpen, pp. 1--17.

\bibitem[{Deniša and Ude(2015)}]{Denisa2015Synthesis}
Deniša M and Ude A (2015) Synthesis of new dynamic movement primitives through
  search in a hierarchical database of example movements.
\newblock \emph{International Journal of Advanced Robotic Systems} 12(10).

\bibitem[{DeWolf et~al.(2016)DeWolf, Stewart, Slotine and
  Eliasmith}]{dewolf2016spiking}
DeWolf T, Stewart T, Slotine JJ and Eliasmith C (2016) A spiking neural model
  of adaptive arm control.
\newblock \emph{Proceedings of the Royal Society B: Biological Sciences}
  283(1843).

\bibitem[{Do et~al.(2014)Do, Schill, Ernesti and Asfour}]{do2014learn}
Do M, Schill J, Ernesti J and Asfour T (2014) Learn to wipe: A case study of
  structural bootstrapping from sensorimotor experience.
\newblock In: \emph{IEEE International Conference on Robotics and Automation}.
  pp. 1858--1864.

\bibitem[{Dometios et~al.(2018)Dometios, Zhou, Papageorgiou, Tzafestas and
  Asfour}]{Dometios2018Vision}
Dometios A, Zhou Y, Papageorgiou X, Tzafestas C and Asfour T (2018)
  Vision-based online adaptation of motion primitives to dynamic surfaces:
  Application to an interactive robotic wiping task.
\newblock \emph{IEEE Robotics and Automation Letters} 3(3): 1410--1417.

\bibitem[{Dragan et~al.(2015)Dragan, M{\"u}lling, Bagnell and
  Srinivasa}]{dragan2015movement}
Dragan AD, M{\"u}lling K, Bagnell JA and Srinivasa SS (2015) Movement
  primitives via optimization.
\newblock In: \emph{IEEE International Conference on Robotics and Automation}.
  Seattle, WA, USA, pp. 2339--2346.

\bibitem[{Duminy et~al.(2017)Duminy, Nguyen and Duhaut}]{Duminy2017Strategic}
Duminy N, Nguyen S and Duhaut D (2017) Strategic and interactive learning of a
  hierarchical set of tasks by the poppy humanoid robot.
\newblock In: \emph{IEEE International Conference on Development and Learning
  and Epigenetic Robotics}. pp. 204--209.

\bibitem[{Eiband et~al.(2019)Eiband, Saveriano and Lee}]{eiband2019learning}
Eiband T, Saveriano M and Lee D (2019) Learning haptic exploration schemes for
  adaptive task execution.
\newblock In: \emph{IEEE International Conference on Robotics and Automation}.
  Montreal, QC, Canada, pp. 7048--7054.

\bibitem[{End et~al.(2017)End, Akrour, Peters and Neumann}]{End2017Layered}
End F, Akrour R, Peters J and Neumann G (2017) Layered direct policy search for
  learning hierarchical skills.
\newblock In: \emph{IEEE International Conference on Robotics and Automation}.
  pp. 6442--6448.

\bibitem[{Ernesti et~al.(2012)Ernesti, Righetti, Do, Asfour and
  Schaal}]{ernesti2012encoding}
Ernesti J, Righetti L, Do M, Asfour T and Schaal S (2012) Encoding of periodic
  and their transient motions by a single dynamic movement primitive.
\newblock In: \emph{IEEE-RAS International Conference on Humanoid Robots}. pp.
  57--64.

\bibitem[{Fabisch and Metzen(2014)}]{fabisch2014active}
Fabisch A and Metzen J (2014) Active contextual policy search.
\newblock \emph{Journal of Machine Learning Research} 15: 3371--3399.

\bibitem[{Fang et~al.(2014)Fang, Wang, Li and Li}]{fang2014control}
Fang Z, Wang G, Li W and Li P (2014) Control-oriented modeling of flight
  demonstrations for quadrotors using higher-order statistics and dynamic
  movement primitives.
\newblock In: \emph{IEEE International Symposium on Industrial Electronics}.
  pp. 1518--1525.

\bibitem[{{Fanger} et~al.(2016){Fanger}, {Umlauft} and
  {Hirche}}]{Fanger2016Gaussian}
{Fanger} Y, {Umlauft} J and {Hirche} S (2016) Gaussian processes for dynamic
  movement primitives with application in knowledge-based cooperation.
\newblock In: \emph{IEEE/RSJ International Conference on Intelligent Robots and
  Systems}. pp. 3913--3919.

\bibitem[{Fei et~al.(2016)Fei, Wang and Liu}]{fei2016learning}
Fei G, Wang S and Liu B (2016) Learning cumulatively to become more
  knowledgeable.
\newblock In: \emph{Proceedings of the 22nd ACM SIGKDD International Conference
  on Knowledge Discovery and Data Mining}. pp. 1565--1574.

\bibitem[{Flash and Hochner(2005)}]{flash2005motor}
Flash T and Hochner B (2005) Motor primitives in vertebrates and invertebrates.
\newblock \emph{Current opinion in neurobiology} 15(6): 660--666.

\bibitem[{Fleischer and Hommel(2008)}]{fleischer2008human}
Fleischer C and Hommel G (2008) A human--exoskeleton interface utilizing
  electromyography.
\newblock \emph{IEEE Transactions on Robotics} 24(4): 872--882.

\bibitem[{Forte et~al.(2012)Forte, Gams, Morimoto and Ude}]{forte2012line}
Forte D, Gams A, Morimoto J and Ude A (2012) On-line motion synthesis and
  adaptation using a trajectory database.
\newblock \emph{Robotics and Autonomous Systems} 60(10): 1327--1339.

\bibitem[{Forte et~al.(2015)Forte, Nemec and Ude}]{Forte2015Exploration}
Forte D, Nemec B and Ude A (2015) Exploration in structured space of robot
  movements for autonomous augmentation of action knowledge.
\newblock In: \emph{The 17th International Conference on Advanced Robotics}.
  pp. 252--258.

\bibitem[{Forte et~al.(2011)Forte, Ude and Gams}]{forte2011realtime}
Forte D, Ude A and Gams A (2011) Real-time generalization and integration of
  different movement primitives.
\newblock In: \emph{IEEE-RAS International Conference on Humanoid Robots}.
  Bled, Slovenia, pp. 590--595.

\bibitem[{Gams et~al.(2015{\natexlab{a}})Gams, Deni\v{s}a and Ude}]{Gams15}
Gams A, Deni\v{s}a M and Ude A (2015{\natexlab{a}}) Learning of parametric
  coupling terms for robot-environment interaction.
\newblock In: \emph{IEEE-RAS International Conference on Humanoid Robots}.
  Seoul, South Korea, pp. 304--309.

\bibitem[{Gams et~al.(2010)Gams, Do, Ude, Asfour and Dillmann}]{gams2010line}
Gams A, Do M, Ude A, Asfour T and Dillmann R (2010) On-line periodic movement
  and force-profile learning for adaptation to new surfaces.
\newblock In: \emph{IEEE-RAS International Conference on Humanoid Robots}.
  Nashville, TN, USA, pp. 560--565.

\bibitem[{Gams et~al.(2009)Gams, Ijspeert, Schaal and
  Lenar\v{c}i\v{c}}]{Gams2009online}
Gams A, Ijspeert A, Schaal S and Lenar\v{c}i\v{c} J (2009) On-line learning and
  modulation of periodic movements with nonlinear dynamical systems.
\newblock \emph{Autonomous robots} 27(1): 3--23.

\bibitem[{Gams et~al.(2014)Gams, Nemec, Ijspeert and Ude}]{Gams2014}
Gams A, Nemec B, Ijspeert A and Ude A (2014) Coupling movement primitives:
  Interaction with the environment and bimanual tasks.
\newblock \emph{IEEE Transactions on Robotics} 30(4): 816--830.

\bibitem[{Gams et~al.(2016)Gams, Petri{\v{c}}, Do, Nemec, Morimoto, Asfour and
  Ude}]{gams2016adaptation}
Gams A, Petri{\v{c}} T, Do M, Nemec B, Morimoto J, Asfour T and Ude A (2016)
  Adaptation and coaching of periodic motion primitives through physical and
  visual interaction.
\newblock \emph{Robotics and Autonomous Systems} 75: 340--351.

\bibitem[{Gams and Ude(2009)}]{gams2009generalization}
Gams A and Ude A (2009) Generalization of example movements with dynamic
  systems.
\newblock In: \emph{IEEE-RAS International Conference on Humanoid Robots}.
  Paris, France, pp. 28--33.

\bibitem[{Gams et~al.(2015{\natexlab{b}})Gams, Ude and
  Morimoto}]{Gams2015Accelerating}
Gams A, Ude A and Morimoto J (2015{\natexlab{b}}) Accelerating synchronization
  of movement primitives: Dual-arm discrete-periodic motion of a humanoid
  robot.
\newblock In: \emph{IEEE/RSJ International Conference on Intelligent Robots and
  Systems}. pp. 2754--2760.

\bibitem[{Gao et~al.(2019)Gao, Zhou and Asfour}]{Gao2019}
Gao J, Zhou Y and Asfour T (2019) Projected force-admittance control for
  compliant bimanual tasks.
\newblock In: \emph{IEEE-RAS International Conference on Humanoid Robots}. pp.
  607--613.

\bibitem[{Ga\v{s}par et~al.(2020)Ga\v{s}par, Deni\v{s}a and
  Ude}]{Gaspar2020346}
Ga\v{s}par T, Deni\v{s}a M and Ude A (2020) Knowledge acquisition through human
  demonstration for industrial robotic assembly.
\newblock \emph{Advances in Intelligent Systems and Computing} 980: 346--353.

\bibitem[{Ga\v{s}par et~al.(2018)Ga\v{s}par, Nemec, Morimoto and
  Ude}]{Gaspar2018}
Ga\v{s}par T, Nemec B, Morimoto J and Ude A (2018) Skill learning and action
  recognition by arc-length dynamic movement primitives.
\newblock \emph{Robotics and Autonomous Systems} 100: 225--235.

\bibitem[{Ghalamzan~E. et~al.(2015)Ghalamzan~E., Paxton, Hager and
  Bascetta}]{Ghalamzan2015Anincremental}
Ghalamzan~E A, Paxton C, Hager G and Bascetta L (2015) An incremental approach
  to learning generalizable robot tasks from human demonstration.
\newblock In: \emph{IEEE International Conference on Robotics and Automation}.
  pp. 5616--5621.

\bibitem[{Ginesi et~al.(2019)Ginesi, Meli, Nakawala, Roberti and
  Fiorini}]{Ginesi201937}
Ginesi M, Meli D, Nakawala H, Roberti A and Fiorini P (2019) A knowledge-based
  framework for task automation in surgery.
\newblock In: \emph{International Conference on Advanced Robotics}. pp. 37--42.

\bibitem[{Guerin et~al.(2014)Guerin, Riedel, Bohren and
  Hager}]{guerin2014adjutant}
Guerin K, Riedel S, Bohren J and Hager G (2014) Adjutant: A framework for
  flexible human-machine collaborative systems.
\newblock In: \emph{IEEE/RSJ International Conference on Intelligent Robots and
  Systems}. pp. 1392--1399.

\bibitem[{Gutzeit et~al.(2018)Gutzeit, Fabisch, Otto, Metzen, Hansen, Kirchner
  and Kirchner}]{Gutzeit2018}
Gutzeit L, Fabisch A, Otto M, Metzen J, Hansen J, Kirchner F and Kirchner E
  (2018) The besman learning platform for automated robot skill learning.
\newblock \emph{Frontiers Robotics AI} 5.

\bibitem[{Haddadin et~al.(2016)Haddadin, Weitschat, Huber, Özparpucu, Mansfeld
  and Albu-Schäffer}]{Haddadin2016Optimal}
Haddadin S, Weitschat R, Huber F, Özparpucu M, Mansfeld N and Albu-Schäffer A
  (2016) Optimal control for viscoelastic robots and its generalization in
  real-time.
\newblock \emph{Springer Tracts in Advanced Robotics} 114: 131--148.

\bibitem[{Hangl et~al.(2015)Hangl, Ugur, Szedmak, Piater and
  Ude}]{Hangl2015Reactive}
Hangl S, Ugur E, Szedmak S, Piater J and Ude A (2015) Reactive, task-specific
  object manipulation by metric reinforcement learning.
\newblock In: \emph{International Conference on Advanced Robotics}. pp.
  557--564.

\bibitem[{Hazara and Kyrki(2016)}]{Hazara2016Reinforcement}
Hazara M and Kyrki V (2016) Reinforcement learning for improving imitated
  in-contact skills.
\newblock In: \emph{IEEE-RAS International Conference on Humanoid Robots}. pp.
  194--201.

\bibitem[{Hazara and Kyrki(2017)}]{hazara2017model}
Hazara M and Kyrki V (2017) Model selection for incremental learning of
  generalizable movement primitives.
\newblock In: \emph{International Conference on Advanced Robotics}. IEEE, pp.
  359--366.

\bibitem[{Hazara and Kyrki(2018)}]{hazara2018speeding}
Hazara M and Kyrki V (2018) Speeding up incremental learning using data
  efficient guided exploration.
\newblock In: \emph{IEEE International Conference on Robotics and Automation}.
  IEEE, pp. 1--8.

\bibitem[{Hazara and Kyrki(2019)}]{hazara2019transferring}
Hazara M and Kyrki V (2019) Transferring generalizable motor primitives from
  simulation to real world.
\newblock \emph{IEEE Robotics and Automation Letters} 4(2): 2172--2179.

\bibitem[{{Hazara} et~al.(2019){Hazara}, {Li} and {Kyrki}}]{Hazara2019Active}
{Hazara} M, {Li} X and {Kyrki} V (2019) Active incremental learning of a
  contextual skill model.
\newblock In: \emph{IEEE/RSJ International Conference on Intelligent Robots and
  Systems}. pp. 1834--1839.

\bibitem[{Herzog et~al.(2016)Herzog, Wörgötter and
  Kulvicius}]{Herzog2016Optimal}
Herzog S, Wörgötter F and Kulvicius T (2016) Optimal trajectory generation
  for generalization of discrete movements with boundary conditions.
\newblock In: \emph{IEEE/RSJ International Conference on Intelligent Robots and
  Systems}. pp. 3143--3149.

\bibitem[{Hoffmann et~al.(2009)Hoffmann, Pastor, Park and
  Schaal}]{hoffmann2009biologically}
Hoffmann H, Pastor P, Park DH and Schaal S (2009) Biologically-inspired
  dynamical systems for movement generation: automatic real-time goal
  adaptation and obstacle avoidance.
\newblock In: \emph{IEEE International Conference on Robotics and Automation}.
  Kobe, Japan, pp. 2587--2592.

\bibitem[{Hogan(1984)}]{hogan1984adaptive}
Hogan N (1984) Adaptive control of mechanical impedance by coactivation of
  antagonist muscles.
\newblock \emph{IEEE Transactions on automatic control} 29(8): 681--690.

\bibitem[{Hogan(1985)}]{hogan1985impedance}
Hogan N (1985) Impedance control: An approach to manipulation: Part i--theory,
  part ii--implementation, and part iii--applications.
\newblock \emph{Journal of Dynamic Systems, Measurement, and Control} 107(1):
  1--7, 8--16, 17--24.

\bibitem[{Hotson et~al.(2016)Hotson, Smith, Rouse, Schieber, Thakor and
  Wester}]{Hotson2016High}
Hotson G, Smith R, Rouse A, Schieber M, Thakor N and Wester B (2016) High
  precision neural decoding of complex movement trajectories using recursive
  bayesian estimation with dynamic movement primitives.
\newblock \emph{IEEE Robotics and Automation Letters} 1(2): 676--683.

\bibitem[{Hu et~al.(2018)Hu, Wu, Geng and Li}]{hu2018evolution}
Hu Y, Wu X, Geng P and Li Z (2018) Evolution strategies learning with variable
  impedance control for grasping under uncertainty.
\newblock \emph{IEEE Transactions on Industrial Electronics} 66(10):
  7788--7799.

\bibitem[{Huang et~al.(2016{\natexlab{a}})Huang, Cheng, Guo, Chen and
  Lin}]{huang2016hierarchical}
Huang R, Cheng H, Guo H, Chen Q and Lin X (2016{\natexlab{a}}) Hierarchical
  interactive learning for a human-powered augmentation lower exoskeleton.
\newblock In: \emph{IEEE international conference on robotics and automation}.
  IEEE, pp. 257--263.

\bibitem[{Huang et~al.(2016{\natexlab{b}})Huang, Cheng, Guo, Lin, Chen and
  Sun}]{huang2016learning}
Huang R, Cheng H, Guo H, Lin X, Chen Q and Sun F (2016{\natexlab{b}}) Learning
  cooperative primitives with physical human-robot interaction for a
  human-powered lower exoskeleton.
\newblock In: \emph{IEEE/RSJ International Conference on Intelligent Robots and
  Systems}. IEEE, pp. 5355--5360.

\bibitem[{{Huang} et~al.(2021){Huang}, {Abu-Dakka}, {Silv\'{e}rio} and
  {Caldwell}}]{Huang2020Toward}
{Huang} Y, {Abu-Dakka} FJ, {Silv\'{e}rio} J and {Caldwell} DG (2021) Toward
  orientation learning and adaptation in cartesian space.
\newblock \emph{IEEE Transactions on Robotics} 37(1): 82--98.

\bibitem[{Huang et~al.(2019)Huang, Rozo, Silv{\'e}rio and
  Caldwell}]{huang2019kernelized}
Huang Y, Rozo L, Silv{\'e}rio J and Caldwell DG (2019) Kernelized movement
  primitives.
\newblock \emph{The International Journal of Robotics Research} 38(7):
  833--852.

\bibitem[{Hwang et~al.(2019)Hwang, Lee, Shin, Baek, Kim, Sun, Kim, Hwang and
  Han}]{Hwang20191905}
Hwang S, Lee S, Shin D, Baek I, Kim M, Sun D, Kim B, Hwang S and Han C (2019)
  Intuitive gait pattern generation for an exoskeleton robot.
\newblock \emph{International Journal of Precision Engineering and
  Manufacturing} 20(11): 1905--1913.

\bibitem[{Ijspeert et~al.(2002{\natexlab{a}})Ijspeert, Nakanishi and
  Schaal}]{ijspeert2002learningAttractor}
Ijspeert A, Nakanishi J and Schaal S (2002{\natexlab{a}}) Learning attractor
  landscapes for learning motor primitives.
\newblock In: \emph{Advances in Neural Information Processing Systems 15}.
  Vancouver, BC, Canada: Cambridge, MA: MIT Press, pp. 1523--1530.

\bibitem[{Ijspeert et~al.(2013)Ijspeert, Nakanishi, Hoffmann, Pastor and
  Schaal}]{Ijspeert2013Dynamical}
Ijspeert AJ, Nakanishi J, Hoffmann H, Pastor P and Schaal S (2013) {Dynamical
  Movement Primitives: Learning Attractor Models for Motor Behaviors}.
\newblock \emph{Neural Computation} 25(2): 328--373.

\bibitem[{Ijspeert et~al.(2001)Ijspeert, Nakanishi and Schaal}]{Ijspeert2001a}
Ijspeert AJ, Nakanishi J and Schaal S (2001) Trajectory formation for imitation
  with nonlinear dynamical systems.
\newblock In: \emph{IEEE/RSJ International Conference on Intelligent Robots and
  Systems}. Maui, HI, USA, pp. 752--757.

\bibitem[{Ijspeert et~al.(2002{\natexlab{b}})Ijspeert, Nakanishi and
  Schaal}]{Ijspeert2002Learning}
Ijspeert AJ, Nakanishi J and Schaal S (2002{\natexlab{b}}) Learning rhythmic
  movements by demonstration using nonlinear oscillators.
\newblock In: \emph{IEEE/RSJ International Conference on Intelligent Robots and
  Systems}, volume~1. Lausanne, Switzerland, pp. 958--963.

\bibitem[{Ijspeert et~al.(2002{\natexlab{c}})Ijspeert, Nakanishi and
  Schaal}]{Ijspeert2002Movement}
Ijspeert AJ, Nakanishi J and Schaal S (2002{\natexlab{c}}) Movement imitation
  with nonlinear dynamical systems in humanoid robots.
\newblock \emph{IEEE International Conference on Robotics and Automation} 2:
  1398--1403.

\bibitem[{Joshi et~al.(2019)Joshi, Koganti and Shibata}]{Joshi20191156}
Joshi R, Koganti N and Shibata T (2019) A framework for robotic clothing
  assistance by imitation learning.
\newblock \emph{Advanced Robotics} 33(22): 1156--1174.

\bibitem[{Joshi et~al.(2017)Joshi, Koganti and Shibata}]{joshi2017robotic}
Joshi RP, Koganti N and Shibata T (2017) Robotic cloth manipulation for
  clothing assistance task using dynamic movement primitives.
\newblock In: \emph{Proceedings of the Advances in Robotics}. Association for
  Computing Machinery, pp. 1--6.

\bibitem[{Kamali et~al.(2016)Kamali, Akbari and
  Akbarzadeh}]{kamali2016trajectory}
Kamali K, Akbari AA and Akbarzadeh A (2016) Trajectory generation and control
  of a knee exoskeleton based on dynamic movement primitives for sit-to-stand
  assistance.
\newblock \emph{Advanced Robotics} 30(13): 846--860.

\bibitem[{Karlsson et~al.(2017)Karlsson, Robertsson and
  Johansson}]{Karlsson2017Autonomous}
Karlsson M, Robertsson A and Johansson R (2017) Autonomous interpretation of
  demonstrations for modification of dynamical movement primitives.
\newblock In: \emph{IEEE International Conference on Robotics and Automation}.
  pp. 316--321.

\bibitem[{Kastritsi et~al.(2018)Kastritsi, Dimeas and
  Doulgeri}]{kastritsi2018progressive}
Kastritsi T, Dimeas F and Doulgeri Z (2018) Progressive automation with dmp
  synchronization and variable stiffness control.
\newblock \emph{IEEE Robotics and Automation Letters} 3(4): 3789--3796.

\bibitem[{Khansari-Zadeh and Billard(2011)}]{Khansari2011learning}
Khansari-Zadeh SM and Billard A (2011) Learning stable non-linear dynamical
  systems with gaussian mixture models.
\newblock \emph{Transactions on Robotics} 27(5): 943--957.

\bibitem[{Khansari-Zadeh and Billard(2014)}]{Khansari2014learning}
Khansari-Zadeh SM and Billard A (2014) Learning control {L}yapunov function to
  ensure stability of dynamical system-based robot reaching motions.
\newblock \emph{Robotics and Autonomous Systems} 62(6): 752--765.

\bibitem[{Kim et~al.(2018{\natexlab{a}})Kim, Seo, Son, Lee, Kim and
  Jin~Kim}]{Kim2018Cooperation}
Kim H, Seo H, Son C, Lee H, Kim S and Jin~Kim H (2018{\natexlab{a}})
  Cooperation in the air: A learning-based approach for the efficient motion
  planning of aerial manipulators.
\newblock \emph{IEEE Robotics and Automation Magazine} 25(4): 76--85.

\bibitem[{Kim et~al.(2015)Kim, Park and Lee}]{Kim2015Adaptability}
Kim JJ, Park SY and Lee JJ (2015) Adaptability improvement of learning from
  demonstration with sequential quadratic programming for motion planning.
\newblock In: \emph{IEEE/ASME International Conference on Advanced Intelligent
  Mechatronics}. pp. 1032--1037.

\bibitem[{Kim et~al.(2018{\natexlab{b}})Kim, Lee and Kim}]{Kim2018Learning}
Kim W, Lee C and Kim H (2018{\natexlab{b}}) Learning and generalization of
  dynamic movement primitives by hierarchical deep reinforcement learning from
  demonstration.
\newblock In: \emph{IEEE/RSJ International Conference on Intelligent Robots and
  Systems}. pp. 3117--3123.

\bibitem[{Kober et~al.(2013)Kober, Bagnell and Peters}]{Kober2013reinforcement}
Kober J, Bagnell JA and Peters J (2013) Reinforcement learning in robotics: A
  survey.
\newblock \emph{The International Journal of Robotics Research} 32(11):
  1238--1274.

\bibitem[{Kober et~al.(2008)Kober, Mohler and Peters}]{kober2008learning}
Kober J, Mohler B and Peters J (2008) Learning perceptual coupling for motor
  primitives.
\newblock In: \emph{IEEE/RSJ International Conference on Intelligent Robots and
  Systems}. pp. 834--839.

\bibitem[{Kober et~al.(2010{\natexlab{a}})Kober, Mohler and
  Peters}]{kober2010imitation2}
Kober J, Mohler B and Peters J (2010{\natexlab{a}}) Imitation and reinforcement
  learning for motor primitives with perceptual coupling.
\newblock In: Sigaud O and Peters J (eds.) \emph{From Motor Learning to
  Interaction Learning in Robots}. Berlin, Heidelberg: Springer Berlin
  Heidelberg, pp. 209--225.

\bibitem[{Kober et~al.(2010{\natexlab{b}})Kober, M{\"u}lling, Kr{\"o}mer,
  Lampert, Sch{\"o}lkopf and Peters}]{kober2010movement}
Kober J, M{\"u}lling K, Kr{\"o}mer O, Lampert CH, Sch{\"o}lkopf B and Peters J
  (2010{\natexlab{b}}) Movement templates for learning of hitting and batting.
\newblock In: \emph{IEEE International Conference on Robotics and Automation}.
  Anchorage, AK, USA: IEEE, pp. 853--858.

\bibitem[{Kober et~al.(2011)Kober, Oztop and Peters}]{kober2011reinfrocement}
Kober J, Oztop E and Peters J (2011) Reinforcement learning to adjust robot
  movements to new situations.
\newblock In: \emph{International Joint Conference on Artificial Intelligence}.
  pp. 2650--2655.

\bibitem[{Kober and Peters(2010)}]{kober2010imitation}
Kober J and Peters J (2010) Imitation and reinforcement learning.
\newblock \emph{IEEE Robotics and Automation Magazine} 17(2): 55--62.

\bibitem[{Kober and Peters(2011)}]{kober2011policy}
Kober J and Peters J (2011) Policy search for motor primitives in robotics.
\newblock \emph{Machine Learning} 84(1-2): 171--203.

\bibitem[{Kober et~al.(2012)Kober, Wilhelm, Oztop and
  Peters}]{kober2012reinforcement}
Kober J, Wilhelm A, Oztop E and Peters J (2012) Reinforcement learning to
  adjust parametrized motor primitives to new situations.
\newblock \emph{Autonomous Robots} 33(4): 361--379.

\bibitem[{Koene et~al.(2014)Koene, Endo, Remazeilles, Prada and
  Wing}]{koene2014experimental}
Koene A, Endo S, Remazeilles A, Prada M and Wing A (2014) Experimental testing
  of the coglaboration prototype system for fluent human-robot object handover
  interactions.
\newblock In: \emph{IEEE International Symposium on Robot and Human Interactive
  Communication}. pp. 249--254.

\bibitem[{Kong and Jeon(2006)}]{kong2006design}
Kong K and Jeon D (2006) Design and control of an exoskeleton for the elderly
  and patients.
\newblock \emph{IEEE/ASME Transactions on mechatronics} 11(4): 428--432.

\bibitem[{{Kormushev} et~al.(2010){Kormushev}, {Calinon} and
  {Caldwell}}]{kormushev2010robot}
{Kormushev} P, {Calinon} S and {Caldwell} DG (2010) Robot motor skill
  coordination with em-based reinforcement learning.
\newblock In: \emph{IEEE/RSJ International Conference on Intelligent Robots and
  Systems}. pp. 3232--3237.

\bibitem[{Kormushev et~al.(2011)Kormushev, Calinon and
  Caldwell}]{kormushev2011imitation}
Kormushev P, Calinon S and Caldwell DG (2011) Imitation learning of positional
  and force skills demonstrated via kinesthetic teaching and haptic input.
\newblock \emph{Advanced Robotics} 25(5): 581--603.

\bibitem[{Koropouli et~al.(2015)Koropouli, Hirche and Lee}]{Koropouli2015}
Koropouli V, Hirche S and Lee D (2015) Generalization of force control policies
  from demonstrations for constrained robotic motion tasks.
\newblock \emph{Journal of Intelligent \& Robotic Systems} 80(1): 133--148.

\bibitem[{Koutras and Doulgeri(2020{\natexlab{a}})}]{koutras2020correct}
Koutras L and Doulgeri Z (2020{\natexlab{a}}) A correct formulation for the
  orientation dynamic movement primitives for robot control in the cartesian
  space.
\newblock In: \emph{Conference on Robot Learning}. PMLR, pp. 293--302.

\bibitem[{Koutras and Doulgeri(2020{\natexlab{b}})}]{koutras2020dynamic}
Koutras L and Doulgeri Z (2020{\natexlab{b}}) Dynamic movement primitives for
  moving goals with temporal scaling adaptation.
\newblock In: \emph{IEEE International Conference on Robotics and Automation}.
  IEEE, pp. 144--150.

\bibitem[{Kramberger et~al.(2017)Kramberger, Gams, Nemec, Chrysostomou, Madsen
  and Ude}]{kramberger2017generalization}
Kramberger A, Gams A, Nemec B, Chrysostomou D, Madsen O and Ude A (2017)
  Generalization of orientation trajectories and force-torque profiles for
  robotic assembly.
\newblock \emph{Robotics and Autonomous Systems} 98: 333--346.

\bibitem[{Kramberger et~al.(2016{\natexlab{a}})Kramberger, Gams, Nemec and
  Ude}]{kramberger2016generalization}
Kramberger A, Gams A, Nemec B and Ude A (2016{\natexlab{a}}) Generalization of
  orientational motion in unit quaternion space.
\newblock In: \emph{IEEE-RAS International Conference on Humanoid Robots}.
  Cancun, Mexico, pp. 808--813.

\bibitem[{Kramberger et~al.(2016{\natexlab{b}})Kramberger, Piltaver, Nemec,
  Gams, Ude et~al.}]{kramberger2016learning}
Kramberger A, Piltaver R, Nemec B, Gams M, Ude A et~al. (2016{\natexlab{b}})
  Learning of assembly constraints by demonstration and active exploration.
\newblock \emph{Industrial Robot: An International Journal} 43: 524–534.

\bibitem[{Kramberger et~al.(2018)Kramberger, Shahriari, Gams, Nemec, Ude and
  Haddadin}]{kramberger2018passivity}
Kramberger A, Shahriari E, Gams A, Nemec B, Ude A and Haddadin S (2018)
  Passivity based iterative learning of admittance-coupled dynamic movement
  primitives for interaction with changing environments.
\newblock In: \emph{IEEE/RSJ International Conference on Intelligent Robots and
  Systems}. Madrid, Spain: IEEE, pp. 6023--6028.

\bibitem[{Kr{\"o}mer et~al.(2010{\natexlab{a}})Kr{\"o}mer, Detry, Piater and
  Peters}]{kroemer2010grasping}
Kr{\"o}mer O, Detry R, Piater J and Peters J (2010{\natexlab{a}}) Grasping with
  vision descriptors and motor primitives.
\newblock In: \emph{International Conference on Informatics in Control,
  Automation and Robotics}, volume~2. pp. 47--54.

\bibitem[{Kr{\"o}mer et~al.(2010{\natexlab{b}})Kr{\"o}mer, Detry, Piater and
  Peters}]{kroemer2010combining}
Kr{\"o}mer OB, Detry R, Piater J and Peters J (2010{\natexlab{b}}) Combining
  active learning and reactive control for robot grasping.
\newblock \emph{Robotics and Autonomous Systems} 58(9): 1105--1116.

\bibitem[{Kronander and Billard(2015)}]{kronander2015passive}
Kronander K and Billard A (2015) Passive interaction control with dynamical
  systems.
\newblock \emph{IEEE Robotics and Automation Letters} 1(1): 106--113.

\bibitem[{Krug and Dimitrov(2015)}]{Krug2015}
Krug R and Dimitrov D (2015) Model predictive motion control based on
  generalized dynamical movement primitives.
\newblock \emph{Journal of Intelligent \& Robotic Systems} 77(1): 17--35.

\bibitem[{Krug and Dimitrovz(2013)}]{krug2013representing}
Krug R and Dimitrovz D (2013) Representing movement primitives as implicit
  dynamical systems learned from multiple demonstrations.
\newblock In: \emph{IEEE International Conference on Advanced Robotics}. pp.
  1--8.

\bibitem[{Kr{\"u}ger et~al.(2014)Kr{\"u}ger, Ude, Petersen, Nemec, Ellekilde,
  Savarimuthu, Rytz, Fischer, Buch, Kraft, Mustafa, Aksoy, Papon, Kramberger
  and W{\"o}rg{\"o}tter}]{kruger2014technologies}
Kr{\"u}ger N, Ude A, Petersen H, Nemec B, Ellekilde LP, Savarimuthu T, Rytz J,
  Fischer K, Buch A, Kraft D, Mustafa W, Aksoy E, Papon J, Kramberger A and
  W{\"o}rg{\"o}tter F (2014) Technologies for the fast set-up of automated
  assembly processes.
\newblock \emph{Kunstliche Intelligenz} 28(4): 305--313.

\bibitem[{Kulvicius et~al.(2013)Kulvicius, Biehl, Aein, Tamosiunaite and
  W{\"o}rg{\"o}tter}]{kulvicius2013interaction}
Kulvicius T, Biehl M, Aein MJ, Tamosiunaite M and W{\"o}rg{\"o}tter F (2013)
  Interaction learning for dynamic movement primitives used in cooperative
  robotic tasks.
\newblock \emph{Robotics and Autonomous Systems} 61(12): 1450--1459.

\bibitem[{Kulvicius et~al.(2011)Kulvicius, Ning, Tamosiunaite and
  W{\"o}rg{\"o}tter}]{kulvicius2011modified}
Kulvicius T, Ning K, Tamosiunaite M and W{\"o}rg{\"o}tter F (2011) Modified
  dynamic movement primitives for joining movement sequences.
\newblock In: \emph{IEEE International Conference on Robotics and Automation}.
  pp. 2275--2280.

\bibitem[{Kulvicius et~al.(2012)Kulvicius, Ning, Tamosiunaite and
  Worg{\"o}tter}]{kulvicius2012joining}
Kulvicius T, Ning K, Tamosiunaite M and Worg{\"o}tter F (2012) Joining movement
  sequences: Modified dynamic movement primitives for robotics applications
  exemplified on handwriting.
\newblock \emph{IEEE Transactions on Robotics} 28(1): 145--157.

\bibitem[{Kupcsik et~al.(2017)Kupcsik, Deisenroth, Peters, Loh, Vadakkepat and
  Neumann}]{Kupcsik2017Model}
Kupcsik A, Deisenroth M, Peters J, Loh A, Vadakkepat P and Neumann G (2017)
  Model-based contextual policy search for data-efficient generalization of
  robot skills.
\newblock \emph{Artificial Intelligence} 247: 415--439.

\bibitem[{Kuppuswamy and Alessandro(2011)}]{kuppuswamy2011impact}
Kuppuswamy N and Alessandro C (2011) Impact of body parameters on dynamic
  movement primitives for robot control.
\newblock \emph{Procedia Computer Science} 7: 166--168.
\newblock 2nd European Future Technologies Conference and Exhibition 2011 (FET
  11).

\bibitem[{Lafleche et~al.(2019)Lafleche, Saunderson and Nejat}]{Lafleche2019}
Lafleche JF, Saunderson S and Nejat G (2019) Robot cooperative behavior
  learning using single-shot learning from demonstration and parallel hidden
  markov models.
\newblock \emph{IEEE Robotics and Automation Letters} 4(2): 193--200.

\bibitem[{Lantz and Murray-Smith(2004)}]{lantz2004rhytmic}
Lantz V and Murray-Smith R (2004) Rhythmic interaction with a mobile device.
\newblock In: \emph{Nordic Conference on Human--Computer Interaction},
  volume~82. pp. 97--100.

\bibitem[{Lauretti et~al.(2018)Lauretti, Cordella, Ciancio, Trigili, Catalan,
  Badesa, Crea, Pagliara, Sterzi, Vitiello et~al.}]{lauretti2018learning}
Lauretti C, Cordella F, Ciancio AL, Trigili E, Catalan JM, Badesa FJ, Crea S,
  Pagliara SM, Sterzi S, Vitiello N et~al. (2018) Learning by demonstration for
  motion planning of upper-limb exoskeletons.
\newblock \emph{Frontiers in neurorobotics} 12: 5.

\bibitem[{Lauretti et~al.(2017)Lauretti, Cordella, Guglielmelli and
  Zollo}]{Lauretti2017Learning}
Lauretti C, Cordella F, Guglielmelli E and Zollo L (2017) Learning by
  demonstration for planning activities of daily living in rehabilitation and
  assistive robotics.
\newblock \emph{IEEE Robotics and Automation Letters} 2(3): 1375--1382.

\bibitem[{LeCun et~al.(2015)LeCun, Bengio and Hinton}]{lecun2015deep}
LeCun Y, Bengio Y and Hinton G (2015) Deep learning.
\newblock \emph{nature} 521(7553): 436--444.

\bibitem[{Lee et~al.(2018)Lee, Kim, Kim and Jin~Kim}]{Lee2018}
Lee H, Kim H, Kim W and Jin~Kim H (2018) An integrated framework for
  cooperative aerial manipulators in unknown environments.
\newblock \emph{IEEE Robotics and Automation Letters} 3(3): 2307--2314.

\bibitem[{Lee et~al.(2020)Lee, Seo and Kim}]{Lee2020135406}
Lee H, Seo H and Kim HG (2020) Trajectory optimization and replanning framework
  for a micro air vehicle in cluttered environments.
\newblock \emph{IEEE Access} 8: 135406--135415.

\bibitem[{Lee and Suh(2013)}]{lee2013skill}
Lee S and Suh I (2013) Skill learning and inference framework for skilligent
  robot.
\newblock In: \emph{IEEE/RSJ International Conference on Intelligent Robots and
  Systems}. pp. 108--115.

\bibitem[{Lemme et~al.(2014)Lemme, Reinhart and
  Steil}]{Lemme2014Selfsupervised}
Lemme A, Reinhart R and Steil J (2014) Self-supervised bootstrapping of a
  movement primitive library from complex trajectories.
\newblock In: \emph{IEEE-RAS International Conference on Humanoid Robots}. pp.
  726--732.

\bibitem[{Lentini et~al.(2020)Lentini, Grioli, Catalano and
  Bicchi}]{Lentini2020}
Lentini G, Grioli G, Catalano M and Bicchi A (2020) Robot programming without
  coding.
\newblock In: \emph{IEEE International Conference on Robotics and Automation}.
  pp. 7576--7582.

\bibitem[{Li et~al.(2020)Li, Fahmy, Li and Sienz}]{Li2020}
Li C, Fahmy A, Li S and Sienz J (2020) An enhanced robot massage system in
  smart homes using force sensing and a dynamic movement primitive.
\newblock \emph{Frontiers in Neurorobotics} 14.

\bibitem[{Li and Fritz(2015)}]{Li2015Teaching}
Li W and Fritz M (2015) Teaching robots the use of human tools from
  demonstration with non-dexterous end-effectors.
\newblock In: \emph{IEEE-RAS International Conference on Humanoid Robots}. pp.
  547--553.

\bibitem[{Li et~al.(2018)Li, Zhao, Chen, Hu, Su and Fukuda}]{Li2018}
Li Z, Zhao T, Chen F, Hu Y, Su CY and Fukuda T (2018) Reinforcement learning of
  manipulation and grasping using dynamical movement primitives for a
  humanoidlike mobile manipulator.
\newblock \emph{IEEE/ASME Transactions on Mechatronics} 23(1): 121--131.

\bibitem[{Lioutikov et~al.(2016)Lioutikov, Kroemer, Maeda and
  Peters}]{lioutikov2016learning}
Lioutikov R, Kroemer O, Maeda G and Peters J (2016) Learning manipulation by
  sequencing motor primitives with a two-armed robot.
\newblock In: Menegatti E, Michael N, Berns K and Yamaguchi H (eds.)
  \emph{Intelligent Autonomous Systems 13}. Cham: Springer International
  Publishing, pp. 1601--1611.

\bibitem[{Liu et~al.(2020)Liu, Geng, Liu and Chen}]{Liu202054652}
Liu C, Geng W, Liu M and Chen Q (2020) Workspace trajectory generation method
  for humanoid adaptive walking with dynamic motion primitives.
\newblock \emph{IEEE Access} 8: 54652--54662.

\bibitem[{Liu et~al.(2014)Liu, Hu, Luo and Wu}]{liu2014visual}
Liu Z, Hu F, Luo D and Wu X (2014) Visual gesture recognition for human robot
  interaction using dynamic movement primitives.
\newblock In: \emph{IEEE International Conference on Systems, Man and
  Cybernetics}. pp. 2094--2100.

\bibitem[{Lon\v{c}arevi\'{c} et~al.(2021)Lon\v{c}arevi\'{c}, Pahi\v{c}, Ude and
  Gams}]{loncarevi2021Generalization}
Lon\v{c}arevi\'{c} Z, Pahi\v{c} R, Ude A and Gams A (2021) Generalization-based
  acquisition of training data for motor primitive learning by neural networks.
\newblock \emph{Applied Sciences} 11(3).

\bibitem[{Lundell et~al.(2017)Lundell, Hazara and
  Kyrki}]{Lundell2017Generalizing}
Lundell J, Hazara M and Kyrki V (2017) Generalizing movement primitives to new
  situations.
\newblock In: Gao Y, Fallah S, Jin Y and Lekakou C (eds.) \emph{Towards
  Autonomous Robotic Systems}. Cham: Springer International Publishing, pp.
  16--31.

\bibitem[{Luo et~al.(2015)Luo, Han, Ding, Ma, Liu and Wu}]{Luo2015Learning}
Luo D, Han X, Ding Y, Ma Y, Liu Z and Wu X (2015) Learning push recovery for a
  bipedal humanoid robot with dynamical movement primitives.
\newblock In: \emph{IEEE-RAS International Conference on Humanoid Robots}. pp.
  1013--1019.

\bibitem[{M.~Wensing and Slotine(2017)}]{Wensing2017Sparse}
M~Wensing P and Slotine JJ (2017) Sparse control for dynamic movement
  primitives.
\newblock \emph{IFAC-PapersOnLine} 50(1): 10114--10121.

\bibitem[{Mandery et~al.(2016)Mandery, Borr{\'a}s, Jöchner and
  Asfour}]{mandery2016using}
Mandery C, Borr{\'a}s J, Jöchner M and Asfour T (2016) Using language models
  to generate whole-body multi-contact motions.
\newblock In: \emph{IEEE/RSJ International Conference on Intelligent Robots and
  Systems}. pp. 5411--5418.

\bibitem[{Mao et~al.(2015)Mao, Yang, Fermüller, Aloimonos and
  Baras}]{Mao2015Learning}
Mao R, Yang Y, Fermüller C, Aloimonos Y and Baras J (2015) Learning hand
  movements from markerless demonstrations for humanoid tasks.
\newblock In: \emph{IEEE-RAS International Conference on Humanoid Robots}. pp.
  938--943.

\bibitem[{Matsubara et~al.(2010)Matsubara, Hyon and
  Morimoto}]{matsubara2010learning}
Matsubara T, Hyon SH and Morimoto J (2010) Learning stylistic dynamic movement
  primitives from multiple demonstrations.
\newblock In: \emph{IEEE/RSJ International Conference on Intelligent Robots and
  Systems}. pp. 1277--1283.

\bibitem[{Matsubara et~al.(2011)Matsubara, Hyon and Morimoto}]{Matsubara2011}
Matsubara T, Hyon SH and Morimoto J (2011) Learning parametric dynamic movement
  primitives from multiple demonstrations.
\newblock \emph{Neural Networks} 24(5): 493--500.

\bibitem[{Meier et~al.(2011)Meier, Theodorou, Stulp and
  Schaal}]{Meier2011movement}
Meier F, Theodorou E, Stulp F and Schaal S (2011) Movement segmentation using a
  primitive library.
\newblock In: \emph{IEEE/RSJ International Conference on Intelligent Robots and
  Systems}. pp. 3407--3412.

\bibitem[{Montebelli et~al.(2015)Montebelli, Steinmetz and
  Kyrki}]{Montebelli2015Onhanding}
Montebelli A, Steinmetz F and Kyrki V (2015) On handing down our tools to
  robots: Single-phase kinesthetic teaching for dynamic in-contact tasks.
\newblock In: \emph{IEEE International Conference on Robotics and Automation}.
  pp. 5628--5634.

\bibitem[{M{\"u}lling et~al.(2013)M{\"u}lling, Kober, Kr{\"o}mer and
  Peters}]{mulling2013learning}
M{\"u}lling K, Kober J, Kr{\"o}mer O and Peters J (2013) Learning to select and
  generalize striking movements in robot table tennis.
\newblock \emph{International Journal of Robotics Research} 32(3): 263--279.

\bibitem[{M{\"u}lling et~al.(2010)M{\"u}lling, Kober and Peters}]{Muelling10}
M{\"u}lling K, Kober J and Peters J (2010) Learning table tennis with a mixture
  of motor primitives.
\newblock In: \emph{IEEE-RAS International Conference on Humanoid Robots}.
  Nashville, TN, USA, pp. 411--416.

\bibitem[{Mussa-Ivaldi(1999)}]{mussa1999modular}
Mussa-Ivaldi FA (1999) Modular features of motor control and learning.
\newblock \emph{Current opinion in neurobiology} 9(6): 713--717.

\bibitem[{Nah et~al.(2020)Nah, Krotov, Russo, Sternad and Hogan}]{Nah2020}
Nah M, Krotov A, Russo M, Sternad D and Hogan N (2020) Dynamic primitives
  facilitate manipulating a whip.
\newblock In: \emph{IEEE RAS and EMBS International Conference on Biomedical
  Robotics and Biomechatronics}. pp. 685--691.

\bibitem[{Nakanishi et~al.(2011)Nakanishi, Rawlik and
  Vijayakumar}]{nakanishi2011stiffness}
Nakanishi J, Rawlik K and Vijayakumar S (2011) Stiffness and temporal
  optimization in periodic movements: An optimal control approach.
\newblock In: \emph{IEEE/RSJ International Conference on Intelligent Robots and
  Systems}. IEEE, pp. 718--724.

\bibitem[{Nemec et~al.(2012)Nemec, Forte, Vuga, Tamosiunaite, W{\"o}rg{\"o}tter
  and Ude}]{nemec2012applying}
Nemec B, Forte D, Vuga R, Tamosiunaite M, W{\"o}rg{\"o}tter F and Ude A (2012)
  Applying statistical generalization to determine search direction for
  reinforcement learning of movement primitives.
\newblock In: \emph{IEEE-RAS International Conference on Humanoid Robots}. pp.
  65--70.

\bibitem[{Nemec et~al.(2013{\natexlab{a}})Nemec, Gams and
  Ude}]{nemec2013velocity}
Nemec B, Gams A and Ude A (2013{\natexlab{a}}) Velocity adaptation for
  self-improvement of skills learned from user demonstrations.
\newblock In: \emph{IEEE-RAS International Conference on Humanoid Robots}.
  Atlanta, GA, USA, pp. 423--428.

\bibitem[{Nemec et~al.(2016)Nemec, Likar, Gams and Ude}]{Nemec2016Bimanual}
Nemec B, Likar N, Gams A and Ude A (2016) Bimanual human robot cooperation with
  adaptive stiffness control.
\newblock In: \emph{IEEE-RAS International Conference on Humanoid Robots}. pp.
  607--613.

\bibitem[{Nemec et~al.(2020)Nemec, Simonic and Ude}]{Nemec20206521}
Nemec B, Simonic M and Ude A (2020) Learning of exception strategies in
  assembly tasks.
\newblock In: \emph{IEEE International Conference on Robotics and Automation}.
  pp. 6521--6527.

\bibitem[{Nemec and Ude(2012)}]{nemec2012action}
Nemec B and Ude A (2012) Action sequencing using dynamic movement primitives.
\newblock \emph{Robotica} 30(5): 837.

\bibitem[{Nemec et~al.(2011)Nemec, Vuga and Ude}]{nemec2011exploiting}
Nemec B, Vuga R and Ude A (2011) Exploiting previous experience to constrain
  robot sensorimotor learning.
\newblock In: \emph{IEEE-RAS International Conference on Humanoid Robots}.
  Bled, Slovenia, pp. 727--732.

\bibitem[{Nemec et~al.(2013{\natexlab{b}})Nemec, Vuga and
  Ude}]{nemec2013efficient}
Nemec B, Vuga R and Ude A (2013{\natexlab{b}}) Efficient sensorimotor learning
  from multiple demonstrations.
\newblock \emph{Advanced Robotics} 27(13): 1023--1031.

\bibitem[{Nemec et~al.(2018)Nemec, \v{Z}lajpah, \v{S}lajpa, Pi\v{s}kur and
  Ude}]{nemec2018anefficient}
Nemec B, \v{Z}lajpah L, \v{S}lajpa S, Pi\v{s}kur J and Ude A (2018) An
  efficient pbd framework for fast deployment of bi-manual assembly tasks.
\newblock In: \emph{IEEE-RAS International Conference on Humanoid Robots}.
  Beijing, China, pp. 166--173.

\bibitem[{Neumann and Steil(2015)}]{Neumann2015learning}
Neumann K and Steil JJ (2015) Learning robot motions with stable dynamical
  systems under diffeomorphic transformations.
\newblock \emph{Robotics and Autonomous Systems} 70: 1--15.

\bibitem[{Niekum et~al.(2012)Niekum, Osentoski, Konidaris and
  Barto}]{niekum2012learning}
Niekum S, Osentoski S, Konidaris G and Barto A (2012) Learning and
  generalization of complex tasks from unstructured demonstrations.
\newblock In: \emph{IEEE/RSJ International Conference on Intelligent Robots and
  Systems}. pp. 5239--5246.

\bibitem[{Niekum et~al.(2015)Niekum, Osentoski, Konidaris, Chitta, Marthi and
  Barto}]{niekum2015learning}
Niekum S, Osentoski S, Konidaris G, Chitta S, Marthi B and Barto A (2015)
  Learning grounded finite-state representations from unstructured
  demonstrations.
\newblock \emph{International Journal of Robotics Research} 34(2): 131--157.

\bibitem[{Ning et~al.(2011)Ning, Kulvicius, Tamosiunaite and
  W{\"o}rg{\"o}tter}]{ning2011accurate}
Ning K, Kulvicius T, Tamosiunaite M and W{\"o}rg{\"o}tter F (2011) Accurate
  position and velocity control for trajectories based on dynamic movement
  primitives.
\newblock In: \emph{IEEE International Conference on Robotics and Automation}.
  pp. 5006--5011.

\bibitem[{Ning et~al.(2012)Ning, Kulvicius, Tamosiunaite and
  W{\"o}rg{\"o}tter}]{ning2012novel}
Ning K, Kulvicius T, Tamosiunaite M and W{\"o}rg{\"o}tter F (2012) A novel
  trajectory generation method for robot control.
\newblock \emph{Journal of Intelligent and Robotic Systems} 68(2): 165--184.

\bibitem[{Norrl{\"o}f(1991)}]{norrlof2002adaptive}
Norrl{\"o}f M (1991) An adaptive iterative learning control algorithm with
  experiments on an industrial robot.
\newblock \emph{IEEE Transaction on Robotics and Automation} 18(2): 188--197.

\bibitem[{Oikonomidis et~al.(2011)Oikonomidis, Kyriazis and
  Argyros}]{Oikonomidis}
Oikonomidis I, Kyriazis N and Argyros A (2011) Efficient model-based 3d
  tracking of hand articulations using kinect.
\newblock In: \emph{British Machine Vision Conference}. pp. 101.1--101.11.

\bibitem[{Ojer De~Andres et~al.(2018)Ojer De~Andres, Mahdi Ghazaei~Ardakani and
  Robertsson}]{OjerDeAndres2018}
Ojer De~Andres M, Mahdi Ghazaei~Ardakani M and Robertsson A (2018)
  Reinforcement learning for 4-finger-gripper manipulation.
\newblock In: \emph{IEEE International Conference on Robotics and Automation}.
  pp. 4257--4262.

\bibitem[{Pahic et~al.(2018)Pahic, Gams, Ude and Morimoto}]{Pahic2018}
Pahic R, Gams A, Ude A and Morimoto J (2018) Deep encoder-decoder networks for
  mapping raw images to dynamic movement primitives.
\newblock In: \emph{IEEE International Conference on Robotics and Automation}.
  pp. 5863--5868.

\bibitem[{Pahi\v{c} et~al.(2020)Pahi\v{c}, Ridge, Gams, Morimoto and
  Ude}]{Pahic2020}
Pahi\v{c} R, Ridge B, Gams A, Morimoto J and Ude A (2020) Training of deep
  neural networks for the generation of dynamic movement primitives.
\newblock \emph{Neural Networks} 127: 121--131.

\bibitem[{Pan and Manocha(2018)}]{Pan2018}
Pan Z and Manocha D (2018) Realtime planning for high-dof deformable bodies
  using two-stage learning.
\newblock In: \emph{IEEE International Conference on Robotics and Automation}.
  pp. 5582--5589.

\bibitem[{Papageorgiou et~al.(2020{\natexlab{a}})Papageorgiou, Dimeas,
  Kastritsi and Doulgeri}]{Papageorgiou2020}
Papageorgiou D, Dimeas F, Kastritsi T and Doulgeri Z (2020{\natexlab{a}})
  Kinesthetic guidance utilizing dmp synchronization and assistive virtual
  fixtures for progressive automation.
\newblock \emph{Robotica} 38(10): 1824--1841.

\bibitem[{Papageorgiou et~al.(2020{\natexlab{b}})Papageorgiou, Kastritsi and
  Doulgeri}]{Papageorgiou2020Apassive}
Papageorgiou D, Kastritsi T and Doulgeri Z (2020{\natexlab{b}}) A passive robot
  controller aiding human coaching for kinematic behavior modifications.
\newblock \emph{Robotics and Computer-Integrated Manufacturing} 61: 101824.

\bibitem[{Paraschos et~al.(2013)Paraschos, Daniel, Peters and
  Neumann}]{Paraschos2013}
Paraschos A, Daniel C, Peters J and Neumann G (2013) Probabilistic movement
  primitives.
\newblock In: Burges C, Bottou L, Welling M, Ghahramani Z and Weinberger K
  (eds.) \emph{Advances in Neural Information Processing Systems 26}. Lake
  Tahoe, Nevada, US: Curran Associates, Inc., pp. 2616--2624.

\bibitem[{Park et~al.(2008)Park, Hoffmann, Pastor and
  Schaal}]{park2008movement}
Park DH, Hoffmann H, Pastor P and Schaal S (2008) Movement reproduction and
  obstacle avoidance with dynamic movement primitives and potential fields.
\newblock In: \emph{IEEE-RAS International Conference on Humanoid Robots}.
  Daejeon, South Korea, pp. 91--98.

\bibitem[{Pastor et~al.(2009)Pastor, Hoffmann, Asfour and Schaal}]{Pastor2009}
Pastor P, Hoffmann H, Asfour T and Schaal S (2009) Learning and generalization
  of motor skills by learning from demonstration.
\newblock In: \emph{IEEE International Conference on Robotics and Automation}.
  Kobe, Japan, pp. 763--768.

\bibitem[{Pastor et~al.(2011)Pastor, Kalakrishnan, Chitta, Theodorou and
  Schaal}]{Pastor2011skill}
Pastor P, Kalakrishnan M, Chitta S, Theodorou E and Schaal S (2011) Skill
  learning and task outcome prediction for manipulation.
\newblock In: \emph{IEEE International Conference on Robotics and Automation}.
  Shanghai, China: IEEE, pp. 3828--3834.

\bibitem[{Pastor et~al.(2013)Pastor, Kalakrishnan, Meier, Stulp, Buchli,
  Theodorou and Schaal}]{pastor2013dynamic}
Pastor P, Kalakrishnan M, Meier F, Stulp F, Buchli J, Theodorou E and Schaal S
  (2013) From dynamic movement primitives to associative skill memories.
\newblock \emph{Robotics and Autonomous Systems} 61(4): 351--361.

\bibitem[{Pastor et~al.(2012)Pastor, Kalakrishnan, Righetti and
  Schaal}]{pastor2012towards}
Pastor P, Kalakrishnan M, Righetti L and Schaal S (2012) Towards associative
  skill memories.
\newblock In: \emph{IEEE-RAS International Conference on Humanoid Robots}.
  Osaka, Japan, pp. 309--315.

\bibitem[{{Pastor} et~al.(2011){Pastor}, {Righetti}, {Kalakrishnan} and
  {Schaal}}]{Pastor2011Online}
{Pastor} P, {Righetti} L, {Kalakrishnan} M and {Schaal} S (2011) Online
  movement adaptation based on previous sensor experiences.
\newblock In: \emph{IEEE/RSJ International Conference on Intelligent Robots and
  Systems}. San Francisco, CA, USA, pp. 365--371.

\bibitem[{Paxton et~al.(2016)Paxton, Jonathan, Kobilarov and
  Hager}]{paxton2016dowhat}
Paxton C, Jonathan F, Kobilarov M and Hager G (2016) Do what i want, not what i
  did: Imitation of skills by planning sequences of actions.
\newblock In: \emph{IEEE/RSJ International Conference on Intelligent Robots and
  Systems}. pp. 3778--3785.

\bibitem[{Perk and Slotine(2006)}]{perk2006motion}
Perk BE and Slotine JJE (2006) Motion primitives for robotic flight control.
\newblock \emph{arXiv preprint cs/0609140} .

\bibitem[{Perrin and Schlehuber-Caissier(2016)}]{Perrin16fast}
Perrin N and Schlehuber-Caissier P (2016) Fast diffeomorphic matching to learn
  globally asymptotically stable nonlinear dynamical systems.
\newblock \emph{Systems \& Control Letters} 96: 51--59.

\bibitem[{Pervez et~al.(2017{\natexlab{a}})Pervez, Ali, Ryu and
  Lee}]{Pervez2017Novel}
Pervez A, Ali A, Ryu JH and Lee D (2017{\natexlab{a}}) Novel learning from
  demonstration approach for repetitive teleoperation tasks.
\newblock In: \emph{IEEE World Haptics Conference}. pp. 60--65.

\bibitem[{Pervez et~al.(2019)Pervez, Latifee, Ryu and Lee}]{Pervez20192055}
Pervez A, Latifee H, Ryu JH and Lee D (2019) Motion encoding with asynchronous
  trajectories of repetitive teleoperation tasks and its extension to
  human-agent shared teleoperation.
\newblock \emph{Autonomous Robots} 43(8): 2055--2069.

\bibitem[{Pervez and Lee(2018)}]{pervez2018learning}
Pervez A and Lee D (2018) Learning task-parameterized dynamic movement
  primitives using mixture of gmms.
\newblock \emph{Intelligent Service Robotics} 11(1): 61--78.

\bibitem[{Pervez et~al.(2017{\natexlab{b}})Pervez, Mao and Lee}]{Pervez2017}
Pervez A, Mao Y and Lee D (2017{\natexlab{b}}) Learning deep movement
  primitives using convolutional neural networks.
\newblock In: \emph{IEEE-RAS International Conference on Humanoid Robots}. pp.
  191--197.

\bibitem[{Peternel and Ajoudani(2017)}]{peternel2017robots}
Peternel L and Ajoudani A (2017) Robots learning from robots: A proof of
  concept study for co-manipulation tasks.
\newblock In: \emph{IEEE-RAS International Conference on Humanoid Robots}.
  Birmingham, UK: IEEE, pp. 484--490.

\bibitem[{Peternel et~al.(2016{\natexlab{a}})Peternel, Noda, Petri{\v{c}}, Ude,
  Morimoto and Babi{\v{c}}}]{peternel2016adaptive}
Peternel L, Noda T, Petri{\v{c}} T, Ude A, Morimoto J and Babi{\v{c}} J
  (2016{\natexlab{a}}) Adaptive control of exoskeleton robots for periodic
  assistive behaviours based on emg feedback minimisation.
\newblock \emph{PLOS ONE} 11(2): e0148942.

\bibitem[{Peternel et~al.(2015)Peternel, Petri{\v{c}} and
  Babi{\v{c}}}]{peternel2015human}
Peternel L, Petri{\v{c}} T and Babi{\v{c}} J (2015) Human-in-the-loop approach
  for teaching robot assembly tasks using impedance control interface.
\newblock In: \emph{IEEE international conference on robotics and automation}.
  Seattle, WA, USA: IEEE, pp. 1497--1502.

\bibitem[{Peternel et~al.(2018{\natexlab{a}})Peternel, Petri{\v{c}} and
  Babi{\v{c}}}]{peternel2018robotic}
Peternel L, Petri{\v{c}} T and Babi{\v{c}} J (2018{\natexlab{a}}) Robotic
  assembly solution by human-in-the-loop teaching method based on real-time
  stiffness modulation.
\newblock \emph{Autonomous Robots} 42(1): 1--17.

\bibitem[{Peternel et~al.(2014)Peternel, Petri{\v{c}}, Oztop and
  Babi{\v{c}}}]{peternel2014teaching}
Peternel L, Petri{\v{c}} T, Oztop E and Babi{\v{c}} J (2014) Teaching robots to
  cooperate with humans in dynamic manipulation tasks based on multi-modal
  human-in-the-loop approach.
\newblock \emph{Autonomous robots} 36(1-2): 123--136.

\bibitem[{Peternel et~al.(2017{\natexlab{a}})Peternel, Rozo, Caldwell and
  Ajoudani}]{peternel2017method}
Peternel L, Rozo L, Caldwell D and Ajoudani A (2017{\natexlab{a}}) A method for
  derivation of robot task-frame control authority from repeated sensory
  observations.
\newblock \emph{IEEE Robotics and Automation Letters} 2(2): 719--726.

\bibitem[{Peternel et~al.(2017{\natexlab{b}})Peternel, Tsagarakis and
  Ajoudani}]{peternel2017human}
Peternel L, Tsagarakis N and Ajoudani A (2017{\natexlab{b}}) A human--robot
  co-manipulation approach based on human sensorimotor information.
\newblock \emph{IEEE Transactions on Neural Systems and Rehabilitation
  Engineering} 25(7): 811--822.

\bibitem[{Peternel et~al.(2016{\natexlab{b}})Peternel, Tsagarakis, Caldwell and
  Ajoudani}]{peternel2016adaptation}
Peternel L, Tsagarakis N, Caldwell D and Ajoudani A (2016{\natexlab{b}})
  Adaptation of robot physical behaviour to human fatigue in human-robot
  co-manipulation.
\newblock In: \emph{IEEE-RAS International Conference on Humanoid Robots}.
  Cancun, Mexico: IEEE, pp. 489--494.

\bibitem[{Peternel et~al.(2018{\natexlab{b}})Peternel, Tsagarakis, Caldwell and
  Ajoudani}]{peternel2018robot}
Peternel L, Tsagarakis N, Caldwell D and Ajoudani A (2018{\natexlab{b}}) Robot
  adaptation to human physical fatigue in human--robot co-manipulation.
\newblock \emph{Autonomous Robots} 42(5): 1011--1021.

\bibitem[{Peters and Schaal(2008{\natexlab{a}})}]{peters2008policy}
Peters J and Schaal S (2008{\natexlab{a}}) Policy learning for motor skills.
\newblock In: Ishikawa M, Doya K, Miyamoto H and Yamakawa T (eds.) \emph{Neural
  Information Processing}. Springer Berlin Heidelberg, pp. 233--242.

\bibitem[{Peters and Schaal(2008{\natexlab{b}})}]{peters2008reinforcement}
Peters J and Schaal S (2008{\natexlab{b}}) Reinforcement learning of motor
  skills with policy gradients.
\newblock \emph{Neural Networks} 21(4): 682--697.

\bibitem[{Petric et~al.(2015)Petric, Colasanto, Gams, Ude and
  Ijspeert}]{Petric2015Bioinspired}
Petric T, Colasanto L, Gams A, Ude A and Ijspeert A (2015) Bio-inspired
  learning and database expansion of compliant movement primitives.
\newblock In: \emph{IEEE-RAS International Conference on Humanoid Robots}. pp.
  346--351.

\bibitem[{Petric et~al.(2018)Petric, Gams, Colasanto, Ijspeert and
  Ude}]{Petric2018}
Petric T, Gams A, Colasanto L, Ijspeert A and Ude A (2018) Accelerated
  sensorimotor learning of compliant movement primitives.
\newblock \emph{IEEE Transactions on Robotics} 34(6): 1636--1642.

\bibitem[{Petri\v{c} et~al.(2011)Petri\v{c}, Gams, Ijspeert and
  \v{Z}lajpah}]{Petric2011online}
Petri\v{c} T, Gams A, Ijspeert AJ and \v{Z}lajpah L (2011) On-line frequency
  adaptation and movement imitation for rhythmic robotic tasks.
\newblock \emph{The International Journal of Robotics Research} 30(14):
  1775--1788.

\bibitem[{Petri\v{c} et~al.(2014{\natexlab{a}})Petri\v{c}, Gams, \v{Z}lajpah
  and Ude}]{petric2014onlinelearning}
Petri\v{c} T, Gams A, \v{Z}lajpah L and Ude A (2014{\natexlab{a}}) Online
  learning of task-specific dynamics for periodic tasks.
\newblock In: \emph{IEEE/RSJ International Conference on Intelligent Robots and
  Systems}. Chicago, IL, USA, pp. 1790--1795.

\bibitem[{Petri\v{c} et~al.(2014{\natexlab{b}})Petri\v{c}, Gams, \v{Z}lajpah,
  Ude and Morimoto}]{Petric2014Onlineapproach}
Petri\v{c} T, Gams A, \v{Z}lajpah L, Ude A and Morimoto J (2014{\natexlab{b}})
  Online approach for altering robot behaviors based on human in the loop
  coaching gestures.
\newblock In: \emph{IEEE International Conference on Robotics and Automation}.
  Hong Kong, China, pp. 4770--4776.

\bibitem[{Petri\v{c} et~al.(2016)Petri\v{c}, Goljat and
  Babi\v{c}}]{Petric2016Cooperative}
Petri\v{c} T, Goljat R and Babi\v{c} J (2016) Cooperative human-robot control
  based on fitts' law.
\newblock In: \emph{IEEE-RAS International Conference on Humanoid Robots}. pp.
  345--350.

\bibitem[{Pfeiffer and Angulo(2015)}]{pfeiffer2015gesture}
Pfeiffer S and Angulo C (2015) Gesture learning and execution in a humanoid
  robot via dynamic movement primitives.
\newblock \emph{Pattern Recognition Letters} 67: 100--107.

\bibitem[{Prada et~al.(2013)Prada, Remazeilles, Koene and
  Endo}]{prada2013dynamic}
Prada M, Remazeilles A, Koene A and Endo S (2013) Dynamic movement primitives
  for human-robot interaction: comparison with human behavioral observation.
\newblock In: \emph{IEEE/RSJ International Conference on Intelligent Robots and
  Systems}. Tokyo, Japan, pp. 1168--1175.

\bibitem[{Prada et~al.(2014)Prada, Remazeilles, Koene and
  Endo}]{prada2014implementation}
Prada M, Remazeilles A, Koene A and Endo S (2014) Implementation and
  experimental validation of dynamic movement primitives for object handover.
\newblock In: \emph{IEEE/RSJ International Conference on Intelligent Robots and
  Systems}. pp. 2146--2153.

\bibitem[{Prakash et~al.(2020)Prakash, Behera, Mohan and
  Jagannathan}]{Prakash2020}
Prakash R, Behera L, Mohan S and Jagannathan S (2020) Dynamic trajectory
  generation and a robust controller to intercept a moving ball in a game
  setting.
\newblock \emph{IEEE Transactions on Control Systems Technology} 28(4):
  1418--1432.

\bibitem[{Quei{\ss}er and Steil(2018)}]{Queiber2018bootstrapping}
Quei{\ss}er J and Steil J (2018) Bootstrapping of parameterized skills through
  hybrid optimization in task and policy spaces.
\newblock \emph{Frontiers Robotics AI} 5(JUN).

\bibitem[{Queißer et~al.(2016)Queißer, Reinhart and
  Steil}]{QueiBer2016Incremental}
Queißer J, Reinhart R and Steil J (2016) Incremental bootstrapping of
  parameterized motor skills.
\newblock In: \emph{IEEE-RAS International Conference on Humanoid Robots}. pp.
  223--229.

\bibitem[{Rai et~al.(2017)Rai, Sutanto, Schaal and Meier}]{Rai2017Learning}
Rai A, Sutanto G, Schaal S and Meier F (2017) Learning feedback terms for
  reactive planning and control.
\newblock In: \emph{IEEE International Conference on Robotics and Automation}.
  pp. 2184--2191.

\bibitem[{Ramirez-Amaro et~al.(2015)Ramirez-Amaro, Beetz and
  Cheng}]{Ramirez-Amaro2015understanding}
Ramirez-Amaro K, Beetz M and Cheng G (2015) Understanding the intention of
  human activities through semantic perception: Observation, understanding and
  execution on a humanoid robot.
\newblock \emph{Advanced Robotics} 29(5): 345--362.

\bibitem[{Rasmussen and Williams(2006)}]{Rasmussen2006}
Rasmussen CE and Williams CKI (2006) \emph{{Gaussian Processes for Machine
  Learning}}.
\newblock Cambridge, Massachusetts: The MIT Press.

\bibitem[{Ravichandar and Dani(2015)}]{ravichandar2015learning}
Ravichandar H and Dani A (2015) Learning contracting nonlinear dynamics from
  human demonstration for robot motion planning.
\newblock In: \emph{ASME Dynamic Systems and Control Conference}.

\bibitem[{Reinhart and Steil(2014)}]{reinhart2014efficient}
Reinhart R and Steil J (2014) Efficient policy search with a parameterized
  skill memory.
\newblock In: \emph{IEEE/RSJ International Conference on Intelligent Robots and
  Systems}. pp. 1400--1407.

\bibitem[{Reinhart and Steil(2015)}]{reinhart2015efficient}
Reinhart R and Steil J (2015) Efficient policy search in low-dimensional
  embedding spaces by generalizing motion primitives with a parameterized skill
  memory.
\newblock \emph{Autonomous Robots} 38(4): 331--348.

\bibitem[{R{\"u}ckert and d'Avella(2013)}]{ruckert2013learned}
R{\"u}ckert E and d'Avella A (2013) Learned parametrized dynamic movement
  primitives with shared synergies for controlling robotic and musculoskeletal
  systems.
\newblock \emph{Frontiers in Computational Neuroscience} .

\bibitem[{Salvador and Chan(2007)}]{salvador2007toward}
Salvador S and Chan P (2007) Toward accurate dynamic time warping in linear
  time and space.
\newblock \emph{Intelligent Data Analysis} 11(5): 561--580.

\bibitem[{Samant et~al.(2016)Samant, Behera and Pandey}]{Samant2016Adaptive}
Samant R, Behera L and Pandey G (2016) Adaptive learning of dynamic movement
  primitives through demonstration.
\newblock In: \emph{International Joint Conference on Neural Networks}. pp.
  1068--1075.

\bibitem[{San~Juan et~al.(2019)San~Juan, Sloth, Kramberger, Petersen,
  {\O}sterg{\aa}rd and Savarimuthu}]{san2019towards}
San~Juan II, Sloth C, Kramberger A, Petersen HG, {\O}sterg{\aa}rd EH and
  Savarimuthu TR (2019) Towards reversible dynamic movement primitives.
\newblock In: \emph{IEEE/RSJ International Conference on Intelligent Robots and
  Systems}. Macau, China: IEEE, pp. 5063--5070.

\bibitem[{Saveriano et~al.(2019)Saveriano, Franzel and
  Lee}]{saveriano2019merging}
Saveriano M, Franzel F and Lee D (2019) Merging position and orientation motion
  primitives.
\newblock In: \emph{IEEE International Conference on Robotics and Automation}.
  Montreal, QC, Canada, pp. 7041--7047.

\bibitem[{Schaal(1999)}]{schaal1999imitation}
Schaal S (1999) Is imitation learning the route to humanoid robots?
\newblock \emph{Trends in cognitive sciences} 3(6): 233--242.

\bibitem[{Schaal(2006{\natexlab{a}})}]{Schaal2006}
Schaal S (2006{\natexlab{a}}) Dynamic movement primitives -a framework for
  motor control in humans and humanoid robotics.
\newblock In: Kimura H, Tsuchiya K, Ishiguro A and Witte H (eds.)
  \emph{Adaptive Motion of Animals and Machines}. Tokyo: Springer Tokyo, pp.
  261--280.

\bibitem[{Schaal(2006{\natexlab{b}})}]{schaal2006dynamic}
Schaal S (2006{\natexlab{b}}) \emph{Dynamic systems: brain, body, and
  imitation}, chapter~1.
\newblock Cambridge University Press, pp. 177--214.

\bibitem[{Schaal and Atkeson(1998)}]{Schaal1998}
Schaal S and Atkeson CG (1998) Constructive incremental learning from only
  local information.
\newblock \emph{Neural Comput.} 10(8): 2047--2084.

\bibitem[{Schaal et~al.(2007)Schaal, Mohajerian and
  Ijspeert}]{schaal2007dynamics}
Schaal S, Mohajerian P and Ijspeert A (2007) Dynamics systems vs. optimal
  control--a unifying view.
\newblock \emph{Progress in brain research} 165: 425--445.

\bibitem[{Schindlbeck and Haddadin(2015)}]{schindlbeck2015unified}
Schindlbeck C and Haddadin S (2015) Unified passivity-based cartesian
  force/impedance control for rigid and flexible joint robots via task-energy
  tanks.
\newblock In: \emph{IEEE International Conference on Robotics and Automation}.
  Seattle, WA, USA, pp. 440--447.

\bibitem[{Schroecker et~al.(2016)Schroecker, Amor and
  Thomaz}]{Schroecker2016Directing}
Schroecker Y, Amor H and Thomaz A (2016) Directing policy search with
  interactively taught via-points.
\newblock In: \emph{Proceedings of the International Joint Conference on
  Autonomous Agents and Multiagent Systems, AAMAS}. pp. 1052--1059.

\bibitem[{Shahriari et~al.(2017)Shahriari, Kramberger, Gams, Ude and
  Haddadin}]{Shahriari2017}
Shahriari E, Kramberger A, Gams A, Ude A and Haddadin S (2017) Adapting to
  contacts: Energy tanks and task energy for passivity-based dynamic movement
  primitives.
\newblock In: \emph{IEEE-RAS International Conference on Humanoid Robots}.
  Birmingham, UK, pp. 136--142.

\bibitem[{Sidiropoulos et~al.(2019)Sidiropoulos, Karayiannidis and
  Doulgeri}]{sidiropoulos2019human}
Sidiropoulos A, Karayiannidis Y and Doulgeri Z (2019) Human-robot collaborative
  object transfer using human motion prediction based on dynamic movement
  primitives.
\newblock In: \emph{European Control Conference}. Naples, Italy, pp.
  2583--2588.

\bibitem[{Silv{\'e}rio et~al.(2019)Silv{\'e}rio, Huang, Abu-Dakka, Rozo and
  Caldwell}]{silverio2019uncertainty}
Silv{\'e}rio J, Huang Y, Abu-Dakka FJ, Rozo L and Caldwell DG (2019)
  Uncertainty-aware imitation learning using kernelized movement primitives.
\newblock In: \emph{IEEE/RSJ International Conference on Intelligent Robots and
  Systems}. IEEE, pp. 90--97.

\bibitem[{Sloth et~al.(2020)Sloth, Kramberger and Iturrate}]{Sloth2020499}
Sloth C, Kramberger A and Iturrate I (2020) Towards easy setup of robotic
  assembly tasks.
\newblock \emph{Advanced Robotics} 34(7-8): 499--513.

\bibitem[{Solak and Jamone(2019)}]{Solak20198246}
Solak G and Jamone L (2019) Learning by demonstration and robust control of
  dexterous in-hand robotic manipulation skills.
\newblock In: \emph{IEEE/RSJ International Conference on Intelligent Robots and
  Systems}. pp. 8246--8251.

\bibitem[{Song et~al.(2020)Song, Liu, Zhang, Zang, Xu and Zhao}]{Song2020325}
Song C, Liu G, Zhang X, Zang X, Xu C and Zhao J (2020) Robot complex motion
  learning based on unsupervised trajectory segmentation and movement
  primitives.
\newblock \emph{ISA Transactions} 97: 325--335.

\bibitem[{Stein et~al.(2014)Stein, W{\"o}rg{\"o}tter, Schoeler, Papon and
  Kulvicius}]{stein2014convexity}
Stein S, W{\"o}rg{\"o}tter F, Schoeler M, Papon J and Kulvicius T (2014)
  Convexity based object partitioning for robot applications.
\newblock In: \emph{IEEE International Conference on Robotics and Automation}.
  pp. 3213--3220.

\bibitem[{Steinmetz et~al.(2015)Steinmetz, Montebelli and
  Kyrki}]{Steinmetz2015Simultaneous}
Steinmetz F, Montebelli A and Kyrki V (2015) Simultaneous kinesthetic teaching
  of positional and force requirements for sequential in-contact tasks.
\newblock In: \emph{IEEE-RAS International Conference on Humanoid Robots}. pp.
  202--209.

\bibitem[{Strachan et~al.(2004)Strachan, Murray-Smith, Oakley and
  {\"A}ngeslev{\"a}}]{strachan2004dynamic}
Strachan S, Murray-Smith R, Oakley I and {\"A}ngeslev{\"a} J (2004) Dynamic
  primitives for gestural interaction.
\newblock In: Brewster S and Dunlop M (eds.) \emph{Mobile Human--Computer
  Interaction}. Springer Berlin Heidelberg, pp. 325--330.

\bibitem[{Straizys et~al.(2020)Straizys, Burke and Ramamoorthy}]{Straizys2020}
Straizys A, Burke M and Ramamoorthy S (2020) Surfing on an uncertain edge:
  Precision cutting of soft tissue using torque-based medium classification.
\newblock In: \emph{IEEE International Conference on Robotics and Automation}.
  pp. 4623--4629.

\bibitem[{Stramigioli(2001)}]{stramigioli2001}
Stramigioli S (2001) \emph{Modeling and IPC Control of Interactive Mechanical
  Systems -- A Coordinate-Free Approach}, \emph{Lecture Notes in Control and
  Information Sciences}, volume 266.
\newblock Springer-Verlag London.

\bibitem[{Stulp et~al.(2012{\natexlab{a}})Stulp, Buchli, Ellmer, Mistry,
  Theodorou and Schaal}]{stulp2012model}
Stulp F, Buchli J, Ellmer A, Mistry M, Theodorou EA and Schaal S
  (2012{\natexlab{a}}) Model-free reinforcement learning of impedance control
  in stochastic environments.
\newblock \emph{IEEE Transactions on Autonomous Mental Development} 4(4):
  330--341.

\bibitem[{Stulp et~al.(2009)Stulp, Oztop, Pastor, Beetz and
  Schaaz}]{stulp2009compact2}
Stulp F, Oztop E, Pastor P, Beetz M and Schaaz S (2009) Compact models of motor
  primitive variations for predictable reaching and obstacle avoidance.
\newblock In: \emph{IEEE-RAS International Conference on Humanoid Robots}. pp.
  589--595.

\bibitem[{Stulp et~al.(2013)Stulp, Raiola, Hoarau, Ivaldi and
  Sigaud}]{Stulp13_hum}
Stulp F, Raiola G, Hoarau A, Ivaldi S and Sigaud O (2013) Learning compact
  parameterized skills with a single regression.
\newblock In: \emph{IEEE-RAS International Conference on Humanoid Robots}.
  Atlanta, GA, USA, pp. 417--422.

\bibitem[{Stulp and Schaal(2011)}]{stulp2011hierarchical}
Stulp F and Schaal S (2011) Hierarchical reinforcement learning with movement
  primitives.
\newblock In: \emph{IEEE-RAS International Conference on Humanoid Robots}.
  Bled, Slovenia, pp. 231--238.

\bibitem[{Stulp et~al.(2011)Stulp, Theodorou, Buchli and
  Schaal}]{stulp2011learning}
Stulp F, Theodorou E, Buchli J and Schaal S (2011) Learning to grasp under
  uncertainty.
\newblock In: \emph{IEEE International Conference on Robotics and Automation}.
  pp. 5703--5708.

\bibitem[{{Stulp} et~al.(2011){Stulp}, {Theodorou}, {Kalakrishnan}, {Pastor},
  {Righetti} and {Schaal}}]{stulp2011learning2}
{Stulp} F, {Theodorou} E, {Kalakrishnan} M, {Pastor} P, {Righetti} L and
  {Schaal} S (2011) Learning motion primitive goals for robust manipulation.
\newblock In: \emph{IEEE/RSJ International Conference on Intelligent Robots and
  Systems}. pp. 325--331.

\bibitem[{Stulp et~al.(2012{\natexlab{b}})Stulp, Theodorou and
  Schaal}]{stulp2012reinforcement}
Stulp F, Theodorou E and Schaal S (2012{\natexlab{b}}) Reinforcement learning
  with sequences of motion primitives for robust manipulation.
\newblock \emph{IEEE Transactions on Robotics} 28(6): 1360--1370.

\bibitem[{Su et~al.(2020)Su, Hu, Li, Knoll, Ferrigno and De~Momi}]{Su20202203}
Su H, Hu Y, Li Z, Knoll A, Ferrigno G and De~Momi E (2020) Reinforcement
  learning based manipulation skill transferring for robot-assisted minimally
  invasive surgery.
\newblock In: \emph{IEEE International Conference on Robotics and Automation}.
  pp. 2203--2208.

\bibitem[{Su et~al.(2021)Su, Mariani, Ovur, Menciassi, Ferrigno and
  De~Momi}]{su2021}
Su H, Mariani A, Ovur SE, Menciassi A, Ferrigno G and De~Momi E (2021) Toward
  teaching by demonstration for robot-assisted minimally invasive surgery.
\newblock \emph{IEEE Transactions on Automation Science and Engineering} .

\bibitem[{Sutanto et~al.(2018)Sutanto, Su, Schaal and Meier}]{Sutanto2018}
Sutanto G, Su Z, Schaal S and Meier F (2018) Learning sensor feedback models
  from demonstrations via phase-modulated neural networks.
\newblock In: \emph{IEEE International Conference on Robotics and Automation}.
  pp. 1142--1149.

\bibitem[{Tamosiunaite et~al.(2011)Tamosiunaite, Nemec, Ude and
  W{\"{o}}rg{\"{o}}tter}]{tamosiunaite2011learning}
Tamosiunaite M, Nemec B, Ude A and W{\"{o}}rg{\"{o}}tter F (2011) Learning to
  pour with a robot arm combining goal and shape learning for dynamic movement
  primitives.
\newblock \emph{Robotics and Autonomous Systems} 59(11): 910--922.

\bibitem[{Tan et~al.(2011)Tan, Erdemir, Kawamura and Du}]{tan2011potential}
Tan H, Erdemir E, Kawamura K and Du Q (2011) A potential field method-based
  extension of the dynamic movement primitive algorithm for imitation learning
  with obstacle avoidance.
\newblock In: \emph{IEEE International Conference on Mechatronics and
  Automation}. pp. 525--530.

\bibitem[{Tan and Kawamura(2011)}]{tan2011computational}
Tan H and Kawamura K (2011) A computational framework for integrating robotic
  exploration and human demonstration in imitation learning.
\newblock In: \emph{IEEE International Conference on Systems, Man and
  Cybernetics}. pp. 2501--2506.

\bibitem[{Tan et~al.(2016)Tan, Zhao and Kannan}]{Tan2016Applying}
Tan H, Zhao Y and Kannan B (2016) Applying adaptive control in modeling human
  motion behaviors in reinforcement robotic learning from demonstrations.
\newblock In: \emph{AAAI Fall Symposium - Technical Report}, volume FS-16-01 -
  FS-16-05. pp. 79--85.

\bibitem[{Tayebi(2004)}]{tayebi2004adaptive}
Tayebi A (2004) Adaptive iterative learning control for robot manipulators.
\newblock \emph{Automatica} 40(7): 1195--1203.

\bibitem[{Theodorou et~al.(2010)Theodorou, Buchli and
  Schaal}]{theodorou2010generalized}
Theodorou E, Buchli J and Schaal S (2010) A generalized path integral control
  approach to reinforcement learning.
\newblock \emph{The Journal of Machine Learning Research} 11: 3137--3181.

\bibitem[{Thota et~al.(2016)Thota, Ravichandar and Dani}]{Thota2016Learning}
Thota P, Ravichandar H and Dani A (2016) Learning and synchronization of
  movement primitives for bimanual manipulation tasks.
\newblock In: \emph{IEEE 55th Conference on Decision and Control}. pp.
  945--950.

\bibitem[{Thrun(1996)}]{thrun1996learning}
Thrun S (1996) Is learning the n-th thing any easier than learning the first?
\newblock In: \emph{Advances in neural information processing systems}. pp.
  640--646.

\bibitem[{Tomi{\'c} et~al.(2014)Tomi{\'c}, Maier and
  Haddadin}]{tomic2014learning}
Tomi{\'c} T, Maier M and Haddadin S (2014) Learning quadrotor maneuvers from
  optimal control and generalizing in real-time.
\newblock In: \emph{IEEE International Conference on Robotics and Automation}.
  pp. 1747--1754.

\bibitem[{Travers et~al.(2018)Travers, Whitman and Choset}]{Travers2018Shape}
Travers M, Whitman J and Choset H (2018) Shape-based coordination in locomotion
  control.
\newblock \emph{The International Journal of Robotics Research} 37(10):
  1253--1268.

\bibitem[{Travers et~al.(2016)Travers, Whitman, Schiebel, Goldman and
  Choset}]{Travers2016Shape}
Travers M, Whitman J, Schiebel P, Goldman D and Choset H (2016) Shape-based
  compliance in locomotion.
\newblock In: \emph{Robotics: Science and Systems}, volume~12.

\bibitem[{Tsagarakis et~al.(2017)Tsagarakis, Caldwell, Negrello, Choi,
  Baccelliere, Loc, Noorden, Muratore, Margan, Cardellino
  et~al.}]{tsagarakis2017walk}
Tsagarakis NG, Caldwell DG, Negrello F, Choi W, Baccelliere L, Loc VG, Noorden
  J, Muratore L, Margan A, Cardellino A et~al. (2017) Walk-man: A
  high-performance humanoid platform for realistic environments.
\newblock \emph{Journal of Field Robotics} 34(7): 1225--1259.

\bibitem[{Ude et~al.(2010)Ude, Gams, Asfour and Morimoto}]{ude2010task}
Ude A, Gams A, Asfour T and Morimoto J (2010) Task-specific generalization of
  discrete and periodic dynamic movement primitives.
\newblock \emph{IEEE Transactions on Robotics} 26(5): 800--815.

\bibitem[{Ude et~al.(2014)Ude, Nemec, Petric and Morimoto}]{Ude2014Orientation}
Ude A, Nemec B, Petric T and Morimoto J (2014) Orientation in cartesian space
  dynamic movement primitives.
\newblock In: \emph{IEEE International Conference on Robotics and Automation},
  3. Hong Kong, China: IEEE, pp. 2997--3004.

\bibitem[{Ugur and Girgin(2020)}]{ugur2020compliant}
Ugur E and Girgin H (2020) Compliant parametric dynamic movement primitives.
\newblock \emph{Robotica} 38(3): 457--474.

\bibitem[{Umlauft et~al.(2017)Umlauft, Fanger and Hirche}]{Umlauft2017Bayesian}
Umlauft J, Fanger Y and Hirche S (2017) Bayesian uncertainty modeling for
  programming by demonstration.
\newblock In: \emph{IEEE International Conference on Robotics and Automation}.
  pp. 6428--6434.

\bibitem[{Umlauft et~al.(2014)Umlauft, Sieber and Hirche}]{umlauft2014dynamic}
Umlauft J, Sieber D and Hirche S (2014) Dynamic movement primitives for
  cooperative manipulation and synchronized motions.
\newblock In: \emph{IEEE International Conference on Robotics and Automation}.
  Hong Kong, China, pp. 766--771.

\bibitem[{Villani and De~Schutter(2008)}]{villani2008force}
Villani L and De~Schutter J (2008) Force control.
\newblock In: Siciliano B and Khatib O (eds.) \emph{Springer Handbook of
  Robotics}. Berlin, Heidelberg: Springer Berlin Heidelberg, pp. 161--185.

\bibitem[{Vuga et~al.(2015{\natexlab{a}})Vuga, Nemec and
  Ude}]{Vuga2015Enhanced}
Vuga R, Nemec B and Ude A (2015{\natexlab{a}}) Enhanced policy adaptation
  through directed explorative learning.
\newblock \emph{International Journal of Humanoid Robotics} 12(3).

\bibitem[{Vuga et~al.(2015{\natexlab{b}})Vuga, Nemec and Ude}]{Vuga2015Speed}
Vuga R, Nemec B and Ude A (2015{\natexlab{b}}) Speed profile optimization
  through directed explorative learning.
\newblock In: \emph{IEEE-RAS International Conference on Humanoid Robots}. pp.
  547--553.

\bibitem[{Vuga et~al.(2016)Vuga, Nemec and Ude}]{Vuga2016Speed}
Vuga R, Nemec B and Ude A (2016) Speed adaptation for self-improvement of
  skills learned from user demonstrations.
\newblock \emph{Robotica} 34(12): 2806--2822.

\bibitem[{Wang et~al.(2019)Wang, Gong and Chen}]{wang2019motion}
Wang B, Gong J and Chen H (2019) Motion primitives representation, extraction
  and connection for automated vehicle motion planning applications.
\newblock \emph{IEEE Transactions on Intelligent Transportation Systems} .

\bibitem[{Wang et~al.(2018)Wang, Gong, Zhang and Chen}]{Wang2018}
Wang B, Gong J, Zhang R and Chen H (2018) Learning to segment and represent
  motion primitives from driving data for motion planning applications.
\newblock In: \emph{IEEE Conference on Intelligent Transportation Systems}. pp.
  1408--1414.

\bibitem[{Wang and Payandeh(2015)}]{Wang2015Astudy}
Wang J and Payandeh S (2015) A study of hand motion/posture recognition in
  two-camera views.
\newblock \emph{Lecture Notes in Computer Science (including subseries Lecture
  Notes in Artificial Intelligence and Lecture Notes in Bioinformatics)} 9475:
  314--323.

\bibitem[{Wang et~al.(2020)Wang, Chen and Yang}]{WANG2020Arobot}
Wang N, Chen C and Yang C (2020) A robot learning framework based on adaptive
  admittance control and generalizable motion modeling with neural network
  controller.
\newblock \emph{Neurocomputing} 390: 260--267.

\bibitem[{Wang et~al.(2016)Wang, Wu, Chan and Tee}]{Wang2016Dynamic}
Wang R, Wu Y, Chan W and Tee K (2016) Dynamic movement primitives plus: For
  enhanced reproduction quality and efficient trajectory modification using
  truncated kernels and local biases.
\newblock In: \emph{IEEE/RSJ International Conference on Intelligent Robots and
  Systems}. pp. 3765--3771.

\bibitem[{Weitschat and Aschemann(2018)}]{weitschat2018safe}
Weitschat R and Aschemann H (2018) Safe and efficient human--robot
  collaboration part ii: Optimal generalized human-in-the-loop real-time motion
  generation.
\newblock \emph{IEEE Robotics and Automation Letters} 3(4): 3781--3788.

\bibitem[{Weitschat et~al.(2013)Weitschat, Haddadin, Huber and
  Albu-Sch{\"a}ffer}]{weitschat2013dynamic}
Weitschat R, Haddadin S, Huber F and Albu-Sch{\"a}ffer A (2013) Dynamic
  optimality in real-time: A learning framework for near-optimal robot motions.
\newblock In: \emph{IEEE/RSJ International Conference on Intelligent Robots and
  Systems}. Tokyo, Japan: IEEE, pp. 5636--5643.

\bibitem[{W{\"o}rg{\"o}tter et~al.(2015)W{\"o}rg{\"o}tter, Geib, Tamosiunaite,
  Aksoy, Piater, Xiong, Ude, Nemec, Kraft, Kruger, Wachter and
  Asfour}]{Worgotter2015Structural}
W{\"o}rg{\"o}tter F, Geib C, Tamosiunaite M, Aksoy E, Piater J, Xiong H, Ude A,
  Nemec B, Kraft D, Kruger N, Wachter M and Asfour T (2015) Structural
  bootstrapping-a novel, generative mechanism for faster and more efficient
  acquisition of action-knowledge.
\newblock \emph{IEEE Transactions on Autonomous Mental Development} 7(2):
  140--154.

\bibitem[{Wu et~al.(2018)Wu, Wang, D'Haro, Banchs and Tee}]{Wu2018}
Wu Y, Wang R, D'Haro L, Banchs R and Tee K (2018) Multi-modal robot
  apprenticeship: Imitation learning using linearly decayed dmp+ in a
  human-robot dialogue system.
\newblock In: \emph{IEEE/RSJ International Conference on Intelligent Robots and
  Systems}. pp. 8582--8588.

\bibitem[{Xu et~al.(2020)Xu, Huang, Cheng, Qiu, Xiang, Shi and Ma}]{Xu2020}
Xu F, Huang R, Cheng H, Qiu J, Xiang S, Shi C and Ma W (2020) Stair-ascent
  strategies and performance evaluation for a lower limb exoskeleton.
\newblock \emph{International Journal of Intelligent Robotics and Applications}
  4(3): 278--293.

\bibitem[{Xu and Wang(2004)}]{xu2004multiple}
Xu JX and Wang W (2004) A multiple internal model approach to movement
  planning.
\newblock In: \emph{IEEE International Symposium on Intelligent Control}. pp.
  186--191.

\bibitem[{Xu et~al.(2005)Xu, Wang, Goh and Lee}]{xu2005internal}
Xu JX, Wang W, Goh J and Lee G (2005) Internal model approach for gait modeling
  and classification.
\newblock In: \emph{IEEE International Conference of the IEEE Engineering in
  Medicine and Biology}. pp. 7688--7691.

\bibitem[{Yang et~al.(2011)Yang, Ganesh, Haddadin, Parusel, Albu-Sch{\"a}ffer
  and Burdet}]{Yang2011}
Yang C, Ganesh G, Haddadin S, Parusel S, Albu-Sch{\"a}ffer A and Burdet E
  (2011) Human-like adaptation of force and impedance in stable and unstable
  interactions.
\newblock \emph{Robotics, IEEE Transactions on} 27(5): 918--930.

\bibitem[{Yang et~al.(2019)Yang, {Zeng}, {Cong}, {Wang} and
  {Wang}}]{yang2019learning}
Yang C, {Zeng} C, {Cong} Y, {Wang} N and {Wang} M (2019) A learning framework
  of adaptive manipulative skills from human to robot.
\newblock \emph{IEEE Transactions on Industrial Informatics} 15(2): 1153--1161.
\newblock \doi{10.1109/TII.2018.2826064}.

\bibitem[{Yang et~al.(2018)Yang, Zeng, Fang, He and Li}]{yang2018dmps}
Yang C, Zeng C, Fang C, He W and Li Z (2018) A dmps-based framework for robot
  learning and generalization of humanlike variable impedance skills.
\newblock \emph{IEEE/ASME Transactions on Mechatronics} 23(3): 1193--1203.

\bibitem[{Yang et~al.(2016)Yang, Sasaki, Suzuki, Kase, Sugano and
  Ogata}]{yang2016repeatable}
Yang PC, Sasaki K, Suzuki K, Kase K, Sugano S and Ogata T (2016) Repeatable
  folding task by humanoid robot worker using deep learning.
\newblock \emph{IEEE Robotics and Automation Letters} 2(2): 397--403.

\bibitem[{Yin and Chen(2014)}]{yin2014learning}
Yin X and Chen Q (2014) Learning nonlinear dynamical system for movement
  primitives.
\newblock In: \emph{IEEE International Conference on Systems, Man and
  Cybernetics}. pp. 3761--3766.

\bibitem[{Yuan et~al.(2020)Yuan, Li, Zhao and Gan}]{Yuan20203830}
Yuan Y, Li Z, Zhao T and Gan D (2020) Dmp-based motion generation for a walking
  exoskeleton robot using reinforcement learning.
\newblock \emph{IEEE Transactions on Industrial Electronics} 67(5): 3830--3839.

\bibitem[{Zeng et~al.(2021)Zeng, Chen, Wang and Yang}]{ZENG2021Learning}
Zeng C, Chen X, Wang N and Yang C (2021) Learning compliant robotic movements
  based on biomimetic motor adaptation.
\newblock \emph{Robotics and Autonomous Systems} 135: 103668.

\bibitem[{Zhang et~al.(2017)Zhang, Fu, Luo and Zhou}]{Zhang2017Robust}
Zhang H, Fu M, Luo H and Zhou W (2017) Robust human action recognition using
  dynamic movement features.
\newblock In: \emph{Intelligent Robotics and Applications}. Springer
  International Publishing, pp. 474--484.

\bibitem[{Zhao et~al.(2020)Zhao, Deng, Li and Hu}]{Zhao202018}
Zhao T, Deng M, Li Z and Hu Y (2020) Cooperative manipulation for a mobile
  dual-arm robot using sequences of dynamic movement primitives.
\newblock \emph{IEEE Transactions on Cognitive and Developmental Systems}
  12(1): 18--29.

\bibitem[{Zhao et~al.(2014)Zhao, Xiong, Fang and Dai}]{zhao2014generating}
Zhao Y, Xiong R, Fang L and Dai X (2014) Generating a style-adaptive trajectory
  from multiple demonstrations.
\newblock \emph{International Journal of Advanced Robotic Systems} 11(1).

\bibitem[{Zhou and Asfour(2017)}]{Zhou2017}
Zhou Y and Asfour T (2017) Task-oriented generalization of dynamic movement
  primitive.
\newblock In: \emph{IEEE/RSJ International Conference on Intelligent Robots and
  Systems}. pp. 3202--3209.

\bibitem[{Zhou et~al.(2016{\natexlab{a}})Zhou, Do and
  Asfour}]{zhou2016coordinate}
Zhou Y, Do M and Asfour T (2016{\natexlab{a}}) Coordinate change dynamic
  movement primitives—a leader-follower approach.
\newblock In: \emph{IEEE/RSJ International Conference on Intelligent Robots and
  Systems}. Daejeon, South Korea, pp. 5481--5488.

\bibitem[{Zhou et~al.(2016{\natexlab{b}})Zhou, Do and
  Asfour}]{Zhou2016Learning}
Zhou Y, Do M and Asfour T (2016{\natexlab{b}}) Learning and force adaptation
  for interactive actions.
\newblock In: \emph{IEEE-RAS International Conference on Humanoid Robots}. pp.
  1129--1134.

\bibitem[{Zhou et~al.(2019)Zhou, Gao and Asfour}]{Zhou2019Learning}
Zhou Y, Gao J and Asfour T (2019) Learning via-point movement primitives with
  inter- and extrapolation capabilities.
\newblock In: \emph{IEEE/RSJ International Conference on Intelligent Robots and
  Systems}. Macau, China, pp. 4301--4308.

\end{thebibliography}
